\documentclass{article}

\usepackage[eandd, nonanonymous, final]{neurips_2026}

\usepackage[utf8]{inputenc}
\usepackage[T1]{fontenc}
\usepackage{hyperref}
\usepackage{url}
\usepackage{amsmath}
\usepackage{amssymb}
\usepackage{amsfonts}
\usepackage{booktabs}
\usepackage{multirow}
\usepackage{graphicx}
\usepackage{float}
\usepackage{adjustbox}
\usepackage{colortbl}
\usepackage[breakable]{tcolorbox}
\tcbuselibrary{listings}
\usepackage{microtype}
\usepackage{xcolor}
\usepackage{nicefrac}
\definecolor{darkblue}{rgb}{0, 0, 0.5}
\hypersetup{colorlinks=true, citecolor=darkblue, linkcolor=darkblue, urlcolor=darkblue}

\newcommand{\apptocentry}[2]{%
  \noindent\hyperref[#1]{#2}\hfill\hyperref[#1]{\pageref*{#1}}\par
}
\newcommand{\apptocsubentry}[2]{%
  \hspace*{1.5em}\hyperref[#1]{#2}\hfill\hyperref[#1]{\pageref*{#1}}\par
}

\title{RGFC: \underline{R}egistry \underline{G}rounded \underline{F}unction \underline{C}alling with stage-decomposed evaluation}

\author{
  \textbf{Ritvik Garimella$^{1}$, Vedant Khandelwal$^{1}$, Anvi Kohli$^{1}$, Amit Sheth$^{1,2}$} \\
  $^1$Artificial Intelligence Institute, University of South Carolina, USA\\
  $^2$Indian AI Research Organization, India \\
}

\begin{document}

\maketitle

\begin{abstract}
Tool-calling evaluations often report whether the final call matches a reference answer, but this single outcome merges different failures: choosing a function outside the available registry, violating the selected schema, or filling arguments with values not supported by the user request. These failures matter more when models must choose from larger tool registries. We introduce \textbf{RFCD}, a registry-scaled function-calling benchmark built from BFCL, Glaive Function Calling v2, and Tool-Call-Data, with standardized JSON-schema registries ranging from 5 to 3{,}000 functions. We also introduce \textbf{TAAG}, a deterministic stage-decomposed evaluator for registry conformance, structural completeness, and argument grounding. Across six locally served sub-4B models, three data sources, and four feedback modes, registry scale is the dominant source of degradation: single-pass TAAG pass rate drops sharply from small to large registries on all sources. Repeated feedback improves some grounding metrics but adds substantial token cost and does not consistently improve end-to-end F1. Requirement-level TAAG-R feedback gives modest gains over stage-only TAAG-S feedback, which indicates that TAAG signals are useful for analysis and correction probes but do not solve registry scaling by themselves. RFCD and TAAG provide an evaluation resource for measuring where tool-calling systems fail as registries grow.
\end{abstract}

\noindent \textbf{Dataset: \url{https://huggingface.co/datasets/functioncallingdataset/fcd}} \\

\begin{figure}[H]
    \centering
    \includegraphics[width=0.88\textwidth]{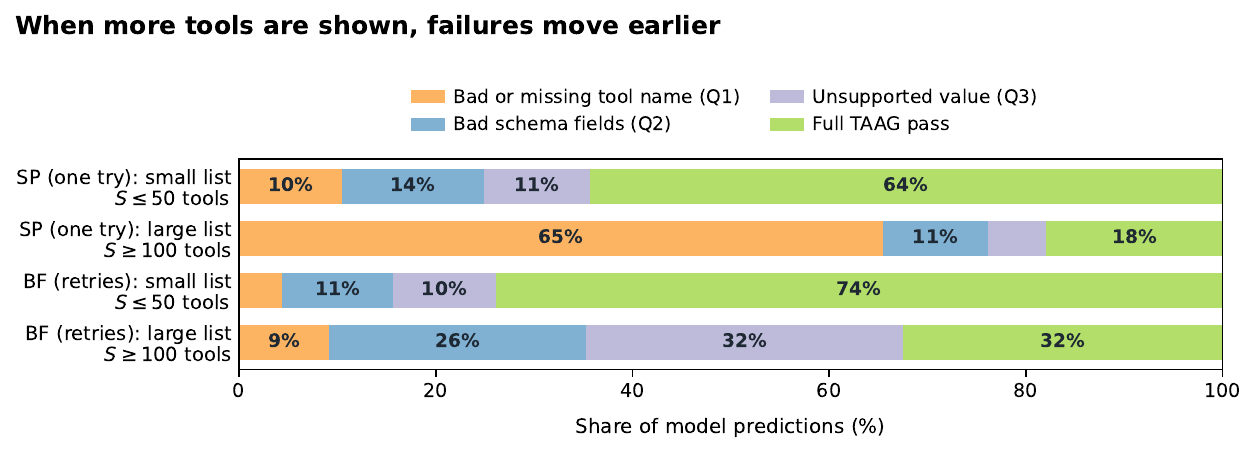}
    \caption{\textbf{Key result: larger registries move mass into earlier TAAG failures.} A registry is the tool list shown to the model, and $S$ is the number of tools in that list. Each bar decomposes retained row-level predictions by the first failed TAAG stage: wrong or missing tool name (Q1), schema-field error (Q2), unsupported argument value (Q3), or full TAAG pass.}
    \label{fig:key_failure_cascade}
\end{figure}

\section{Introduction}

Large language models increasingly act through external tools: APIs, search systems, databases, code interpreters, and domain services \citep{bubeck2023sparks,naveed2025comprehensive,schick2023toolformer,qin2024tool,mialon2023augmented}. This changes the evaluation problem. A generated response is not only text; it is a structured action that must name an available function, satisfy that function's schema, and use argument values supported by the user request. This setting is closely related to agentic orchestration and tool-sequence selection, where the model acts as a controller over external capabilities \citep{plaat2025agentic,xi2025rise,wang2024survey,yao2022react,gao2023retrieval,wei2025autotir,zhang2025landscape}. If any one of these obligations fails, the system can issue the wrong external action even when the output is syntactically close to a valid call.

Current tool-calling benchmarks have made this problem measurable \citep{li2023apibank,qin2024toolllm,patil2024gorilla,patil2025bfcl}, but two gaps remain. First, many evaluations expose a small task-specific tool list, while deployed systems increasingly present larger and changing registries of callable functions, as in emerging protocol-based tool ecosystems such as MCP \citep{anthropic2024mcp}. Prior work has shown that long contexts, tool-selection bias, and tool-use failure modes can affect model behavior in these settings \citep{levy2024same,modarressi2025nolima,paramanayakam2025less,blankenstein2026biasbusters,xu2025reducing,zhang2024toolbehonest}. Retrieval-based tool selection can mitigate some scale pressure \citep{gan2025rag,faghih2025tool,shi2025retrieval}, but it does not replace a benchmark that measures how tool selection and argument grounding change as the registry grows. In such settings, the model must select the right capability from many irrelevant candidates before it can produce arguments. Second, common exact-match or execution-style scores collapse multiple obligations into one number. A model that emits an unavailable function, a model that selects the right function but omits required fields, and a model that invents unsupported argument values may all be counted as wrong, although these failures point to different fixes.

We address these gaps with two connected works (\textit{parts of RGFC}). The first is \textbf{RFCD}, a registry-based function-calling benchmark that standardizes data from BFCL, Glaive, and Tool-Call-Data into a common JSON-schema format. RFCD pairs each query with registries of increasing size, from 5 to 3{,}000 functions where the source pool permits. This keeps the task format fixed while increasing the number of candidate tools shown to the model. The goal is not to simulate every production deployment, but to isolate how registry scale changes tool selection and argument generation.

The second contribution is \textbf{TAAG}, a deterministic evaluator for stage-decomposed tool-call analysis (based on \citep{garimella2026saag}). TAAG evaluates a predicted call through three ordered stages: \emph{registry conformance}, \emph{structural completeness}, and \emph{argument grounding}. Later measurements are computed only when earlier obligations pass, so the reported diagnostics state which part of the tool-use pipeline has been satisfied. This separates failures that a single final score would merge. It also enables feedback probes: when a call fails, the first failed TAAG stage can be converted into a short correction signal without revealing the gold answer.

Our experiments evaluate six locally served sub-4B models across the RFCD sources, registry scales, and four feedback modes: single pass (SP), binary feedback (BF), stage-only TAAG-S feedback, and requirement-level TAAG-R feedback. Registry scale is the strongest source of degradation in this setting. SP TAAG pass rate drops from $0.626\pm0.184$ to $0.091\pm0.080$ on BFCL, from $0.654\pm0.110$ to $0.244\pm0.231$ on Glaive, and from $0.649\pm0.069$ to $0.133\pm0.122$ on Tool-Call-Data when moving from small to large registries. Repeated feedback improves some intrinsic grounding metrics, but it increases token cost substantially and produces smaller, less consistent gains in end-to-end F1. TAAG-R is usually better than TAAG-S, but the improvement is modest compared with the registry-scale drop.
Figure~\ref{fig:key_failure_cascade} summarizes the main result visually: larger registries reduce full TAAG passes and show where failures first enter the registry, schema, and grounding cascade.

\textbf{Scope.}
This paper studies fixed-registry, single-action function calling. It does not evaluate real tool execution, dynamic tool discovery, multi-tool plans, or multi-agent workflows. Distractors are sampled uniformly from source-specific tool pools, so the registry stress test measures scale under controlled sampling rather than adversarially chosen confusable tools. TAAG's grounding metrics are lexical and type-aware; they are not semantic equivalence checks. The model suite is limited to small locally served models with deterministic decoding. These choices make the evaluation reproducible and diagnostic, but they bound claims about frontier systems and deployed tool platforms.

\textbf{Contributions.}
\begin{enumerate}
    \item We introduce RFCD, a registry-scaled function-calling benchmark built from three sources and standardized into a common JSON-schema format with registries from 5 to 3{,}000 functions.
    \item We introduce TAAG, a deterministic stage-decomposed evaluator that separates registry conformance, structural completeness, and argument grounding.
    \item We provide a controlled evaluation across six small local models and four feedback modes, showing that registry scale sharply degrades tool-calling reliability and that diagnostic feedback provides useful but limited correction signal.
\end{enumerate}

\section{RFCD: Registry-based Function Calling Dataset}
\label{sec:dataset}

\begin{figure*}[h]
\centering
\includegraphics[width=.85\textwidth]{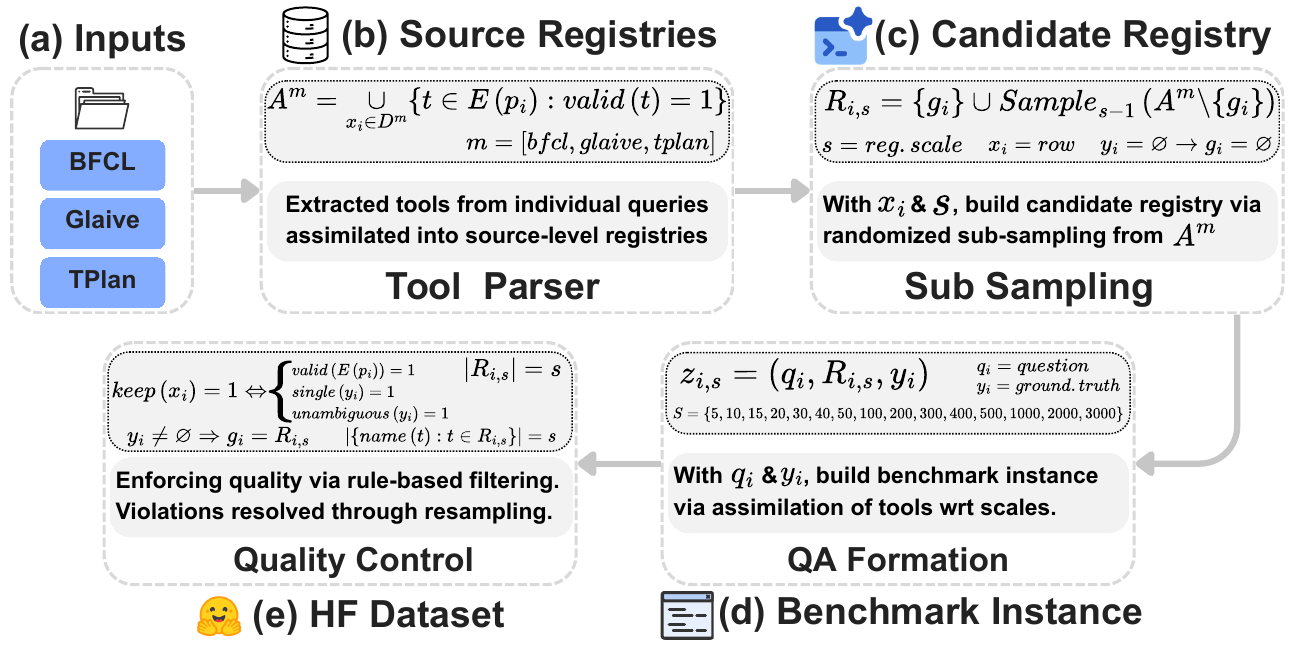}
\caption{\textbf{The RFCD data-curation and construction pipeline.} (a) Single-function datasets are input to (b) a Tool Parser that extracts and aggregates tools into source registries. (c) Candidate registries are built via randomized sub-sampling at different scales. (d) Benchmark instances are formed by combining sampled tools with QA pairs, and (e) rule-based filtering ensures quality, producing the final dataset.}
\label{fig:rfcd}
\end{figure*}

RFCD is a registry-based benchmark for evaluating tool-calling behavior at realistic registry sizes. Each instance consists of a user query, a registry of candidate function schemas, and a ground-truth target indicating either the correct function call or abstention. By varying registry size while keeping the task fixed, RFCD isolates the effect of scale on tool selection and argument generation.

RFCD is constructed from three complementary datasets: Berkeley Function Calling Leaderboard (BFCL) \citep{patil2025bfcl} , Glaive Function Calling v2 \citep{glaiveai2023functioncalling}, and Tool Plannings v0.2 (also referred to as Tool-Call-Data) \cite{vikhrmodels_tool_plannings_v0_2} . BFCL provides structured benchmark-style APIs, Glaive contributes broad instruction-driven examples including no-call cases, and Tool Plannings adds planning-oriented synthetic functions. All examples are normalized into a unified format with JSON-schema tool definitions. Malformed entries, ambiguous targets, and multi-call outputs were filtered out whilst preserving abstention cases to evaluate both tool selection and no-call behavior. Appendix \ref{app:dataset} provides full data creation pipeline, formulas, benchmark design principles, and registry snapshots.

\begin{table}[h]
\small
\centering
\begin{minipage}[t]{0.34\textwidth}
\centering
\begin{tabular}{lr}
\toprule
\textbf{Total Stats} & \textbf{Value} \\
\midrule
Main rows    & 1{,}192{,}662 \\
Sample rows  & 975 \\
Unique tasks & 79{,}798 \\
\bottomrule \\
\end{tabular}
\end{minipage}%
\hfill
\begin{minipage}[t]{0.62\textwidth}
\centering
\begin{tabular}{lrrc}
\toprule
\textbf{Data Source} & \textbf{Total rows} & \textbf{Rows/scale} & \textbf{Scales} \\
\midrule
\texttt{glaive}          & 1{,}175{,}430 & 78{,}362 & 15 \\
\texttt{bfcl}            & 6{,}600       & 550      & 12 \\
\texttt{tool\_plannings} & 10{,}632      & 886      & 12 \\
\bottomrule \\
\end{tabular}
\end{minipage}
\caption{Dataset statistics (left) and per-source breakdown (right). RFCD contains 1{,}192{,}662 main rows across three source configurations and 79{,}798 unique tasks before scale expansion.}
\label{tab:dataset-stats}
\end{table}

\textbf{Single-action setting. }
RFCD focuses on single-action function calling to isolate the registry-selection problem. Multi-step tool use introduces additional confounds such as model reasoning, decomposition and execution order. By restricting each instance to a single gold action, RFCD directly measures whether a model can map a query to the correct function and arguments in the presence of many irrelevant tools.

\textbf{Notations.}
Source datasets $\mathcal{D}^{m}$ are indexed by $m$, with rows within each source indexed by $i$. Each row is a pair $x_i = (q_i, y_i)$, where $q_i$ is a user query and $y_i$ its ground-truth answer: either a single tool call or the no-call symbol $\varnothing$. For positive examples ($y_i \neq \varnothing$), $g_i$ denotes the \emph{gold tool} invoked by $y_i$. The scalar $s$ denotes registry size, i.e., the number of tool definitions presented to the model at evaluation time.

\textbf{Registry Construction.}
\label{sec:registry_construction}
For each $\mathcal{D}^{m}$, tools are extracted via $\mathcal{E}(p_i)$ into a source-level pool $\mathcal{A}^{m} = \bigcup_{x_i \in \mathcal{D}^{m}} \{t \in \mathcal{E}(p_i) : \operatorname{valid}(t)=1\}$. For a given $s$, we construct $\mathcal{R}_{i,s}$ by inserting $g_i$ when $y_i \neq \varnothing$ and sampling $s$ distractors from $\mathcal{A}^{m}$ otherwise, yielding $z_{i,s} = (q_i, \mathcal{R}_{i,s}, y_i)$. We sweep $s$ up to a source-determined maximum (Glaive: 3000; others: 500), keeping queries and targets fixed so that only distractor count grows.

\textbf{Quality Control and Benchmark Design.}
\label{sec:quality_control}
Instances satisfy $\operatorname{keep}(x_i)=1 \iff \operatorname{valid}(\mathcal{E}(p_i)) \land \operatorname{single}(y_i) \land \operatorname{unambiguous}(y_i)$, and registry integrity $|\mathcal{R}_{i,s}| = |\{\operatorname{name}(t) : t \in \mathcal{R}_{i,s}\}| = s \land (y_i \neq \varnothing \Rightarrow g_i \in \mathcal{R}_{i,s})$ is enforced by resampling. Organized by source and scale, RFCD enables evaluation of four core tool-calling capabilities: tool necessity, function selection, schema-valid argument generation, and query grounding. This is operationalized via a stage-decomposed evaluation framework, TAAG (Section \ref{sec:methodology}).

\section{Tool-Augmentation Assessment and Grounding}
\label{sec:methodology}

\begin{figure*}[h]
\centering
\includegraphics[width=0.93\textwidth]{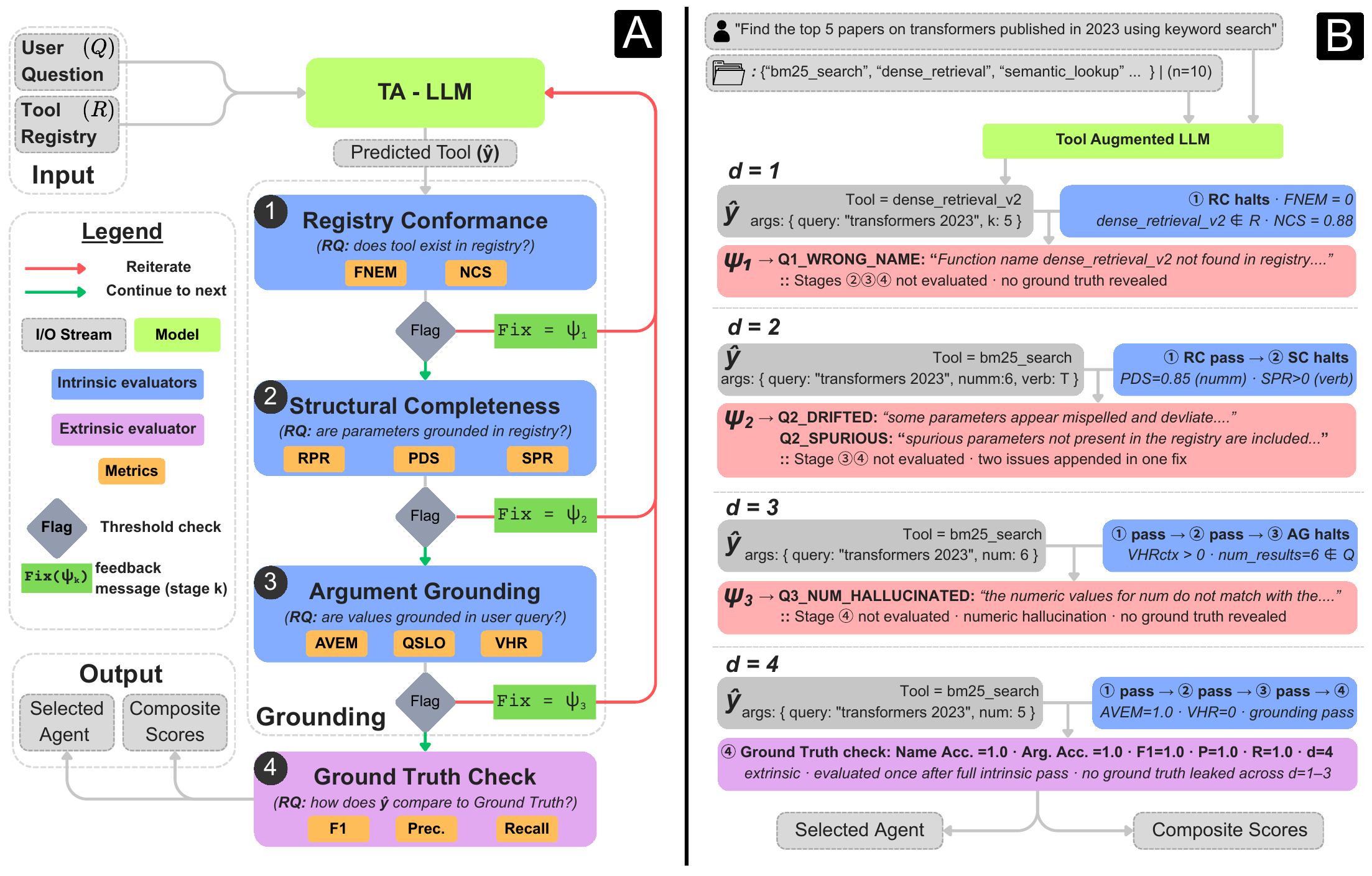}
\caption{\textbf{TAAG as a diagnostic cascade.} (A) A prediction is checked in order for registry conformance, schema completeness, and argument grounding. Each intrinsic stage uses only the prompt registry, selected schema, and user request; ground truth is reserved for extrinsic scoring after the intrinsic checks. (B) The same stage outputs can be converted into deterministic correction signals $\psi_k$, which identify the first failed obligation without revealing the gold call.}
\label{fig:main}
\end{figure*}

\textsc{TAAG} evaluates whether an emitted tool call is usable with respect to the information given to the model. A valid call must name a function in the shown registry, satisfy that function's JSON schema, and fill arguments with values supported by the user request. These obligations are related but separable: a model can select a registered function and still omit required fields, or produce a schema-valid call whose values are invented. A single exact-match score counts all of these as wrong; TAAG records where the failure first enters the tool-use pipeline.

TAAG is a strict cascade. Evaluation halts at the first failed stage, so later diagnostics are \emph{stage-conditional}: they describe only calls that reached that stage, not unconditional population rates. The same deterministic check can emit a correction signal $\psi_k$ for feedback experiments, but the correction loop is used only as an actionability probe for the evaluation. Appendix~\ref{app:fixes} gives the formulas, thresholds, and correction messages.

\textbf{Registry Conformance. }
\label{sec:q1}Stage~1 asks whether the predicted function name $\hat{y}_\text{name}$ belongs to the registry $\mathcal{R}$ shown in the prompt. We report Function Name Exact Match (FNEM), with separate outcomes for parse/no-name failure, registry mismatch, and registry conformance. When the name is not in $\mathcal{R}$, TAAG also computes a Nearest Candidate Score (NCS) against valid registry names. High NCS indicates recoverable drift toward a real tool name; low NCS indicates a more fabricated name. This stage checks availability, not whether the selected available tool matches the source reference.

\textbf{Structural Completeness. }
\label{sec:q2}Stage~2 runs only after the name is registry-conformant. It checks whether the predicted argument keys satisfy the selected tool schema. Required Parameter Recall (RPR) must be $1$, the Spurious Parameter Rate (SPR) must be $0$, and any recoverable misspellings detected by the Parameter Drift Score (PDS) must be corrected. This separates missing required fields, unsupported extra fields, and drifted field names before value grounding is considered.

\textbf{Argument Value Grounding. }
\label{sec:q3}Stage~3 checks whether argument values are supported by the user request $Q$. Argument Value Exact Match (AVEM) requires exact type-aware support: strings are matched case-insensitively at word boundaries, numeric values must match numeric tokens in $Q$, and list values require every element to be supported. Query-Supported Lexical Overlap (QSLO) gives partial lexical evidence for non-exact string values, while the contextual Value Hallucination Rate $\text{VHR}_\text{ctx}$ flags unsupported contextual values. Numeric mismatches receive no partial credit; values without grounding context, such as symbols legitimately inferred from names, are excluded rather than counted as hallucinations.

\textbf{Correction and Extrinsic Scoring. }
\label{sec:loop}For BF, TAAG-S, and TAAG-R feedback conditions, the model can revise up to $D_\text{max}=15$ times and stops early on a full TAAG pass. BF supplies only a fixed negative response. TAAG-S reports the first failed stage, and TAAG-R adds the violated requirement, such as a missing field or unsupported value. None of these messages reveal the gold function call. Extrinsic precision, recall, and F1 are computed separately against the source ground truth and are reported as end-to-end validity checks rather than intrinsic TAAG diagnostics.

\section{Experimental Setup}
\label{sec:experiments}

Our experiments characterize RFCD and TAAG as an evaluation resource rather than a model leaderboard. The central question is whether a stage-decomposed evaluation exposes tool-calling failures that are hidden by a single end-to-end correctness number. We therefore vary three factors under a fixed inference protocol: the source dataset and registry scale, the model class, and the amount of feedback returned after an unsuccessful tool call. This design compares direct tool calling, repeated attempts with uninformative feedback, and repeated attempts with grounded diagnostic feedback under the same registry conditions.

\textbf{Datasets and registry scales.}
We evaluate on three function-calling sources: BFCL~\citep{patil2025bfcl}, Glaive Function Calling v2~\citep{glaiveai2023functioncalling}, and Tool-Call-Data. Each instance is paired with a registry that contains the target function and additional distractor functions from the same source family. The registry size $S$ is the number of available functions shown to the model for that query. We use the same definition of $S$ across all sources so that changes in performance can be interpreted as registry-scale effects rather than changes in the evaluation protocol.

The registry grid covers small, medium, and large tool pools. For BFCL and Tool-Call-Data, we evaluate $S \in \{5,10,15,20,30,40,50,100,200,300,400,500\}$. For Glaive, whose processed registry supports larger expansions, we additionally evaluate $S \in \{1000,2000,3000\}$. Dataset construction, schema normalization, and source-specific filtering details are reported in Appendix~\ref{app:dataset}; the main experiments use the resulting standardized registries without changing the task format across models or feedback modes.

\textbf{Models.}
We evaluate six locally served small models through Ollama: Llama~3.2:3B~\citep{meta2024llama32modelcard}, Qwen~2.5:3B~\citep{qwen2024qwen25}, Gemma~4:e4B~\citep{google2026gemma4}, Qwen~3:4B-Thinking~\citep{qwen2025qwen3technicalreport}, Phi-4-mini:3.8B~\citep{microsoft2025phi4mini}, and Granite~4:3B~\citep{ibm2025granite4}. For analysis, we group them as compact instruction-following models, reasoning/thinking models, and tool- or structured-output-aware models. The grouping is analytic rather than exclusive: it lets us compare broad behavior classes under the same registry and feedback conditions. We use small local models to make exhaustive source--scale--mode sweeps reproducible and to expose failures that would be harder to study if the experiment mainly measured frontier-model capacity. All runs use deterministic decoding, no fine-tuning, and a fixed prompt format within each feedback mode. Appendix~\ref{app:models} gives the model rationale, exact tags, and inference settings.

\textbf{Feedback modes.}
We evaluate three experiment families, which together define four paper-facing modes. \textbf{SP} is single pass: the model receives the query and registry once and must emit a tool call without correction. \textbf{BF} is binary feedback: the model may revise its answer for up to $D_\text{max}=15$ iterations, but every failed attempt receives only a fixed negative signal. This condition separates the effect of having another attempt from the effect of receiving useful diagnostic information. \textbf{TAAG feedback} is diagnostic feedback produced from the TAAG cascade at the first failed stage, again with a maximum of $D_\text{max}=15$ iterations.

We use two TAAG-feedback variants in the analysis. \textbf{TAAG-S} returns a concise stage-level diagnosis, identifying whether the failure occurred at registry conformance, structural completeness, or argument grounding. \textbf{TAAG-R} adds the specific violated requirement, such as the missing field, schema mismatch, or unsupported argument value. Neither variant provides the gold tool call directly; both expose only diagnostic information that TAAG can deterministically justify from the prediction, registry, and request. Table~\ref{tab:conditions} summarizes the four modes.

\begin{table}[h]
\centering
\small
\renewcommand{\arraystretch}{1.25}
\begin{tabular}{llcc}
\toprule
\textbf{Mode} & \textbf{Feedback signal} & \textbf{Loop} & \textbf{$D_\text{max}$} \\
\midrule
SP & None & No & 1 \\
BF & Fixed negative response & Yes & 15 \\
TAAG-S & Failed TAAG stage & Yes & 15 \\
TAAG-R & Failed stage and violated requirement & Yes & 15 \\
\bottomrule \\
\end{tabular}
\caption{Experimental feedback modes. SP is one-shot inference, BF gives repeated attempts with only a fixed failure signal, TAAG-S gives the failed TAAG stage, and TAAG-R gives the failed stage plus the violated requirement. The four modes are evaluated across the same model suite, dataset sources, and registry scales.}
\label{tab:conditions}
\end{table}

\textbf{Metrics.}
We report both intrinsic and extrinsic metrics because they answer different evaluation questions. Intrinsic metrics are computed from the prediction, prompt registry, schema, and user request, without consulting the source ground-truth call. These include Q1 registry conformance, Q2 structural completeness, Q3 argument grounding, the overall TAAG pass rate, and the grounding score. Since TAAG is a strict cascade, later-stage metrics are stage-conditional and should be read as diagnostics over the subset of predictions that reached that stage. These metrics are the main reason to use TAAG: they identify whether an incorrect tool call failed because the model selected an unavailable function, emitted an incomplete or malformed call, or supplied arguments that were not grounded in the request.

Extrinsic metrics compare the final emitted call against the source ground truth and include precision, recall, and F1. We use these end-to-end metrics as validity checks on the stage-wise diagnostics: a useful diagnostic should explain changes in final task success rather than only producing interpretable intermediate labels. The main result table labels each reported value as intrinsic, extrinsic, or cost so that TAAG-stage behavior is not confused with ground-truth-call matching.

\section{Results}
\label{sec:results}

We report means with standard deviations over configuration-level aggregates from completed non-error instances. The main text emphasizes four findings; Appendix~\ref{app:res} gives additional source, scale, model-family, and cost views.

\begin{figure*}[t]
    \centering
    \includegraphics[width=\textwidth]{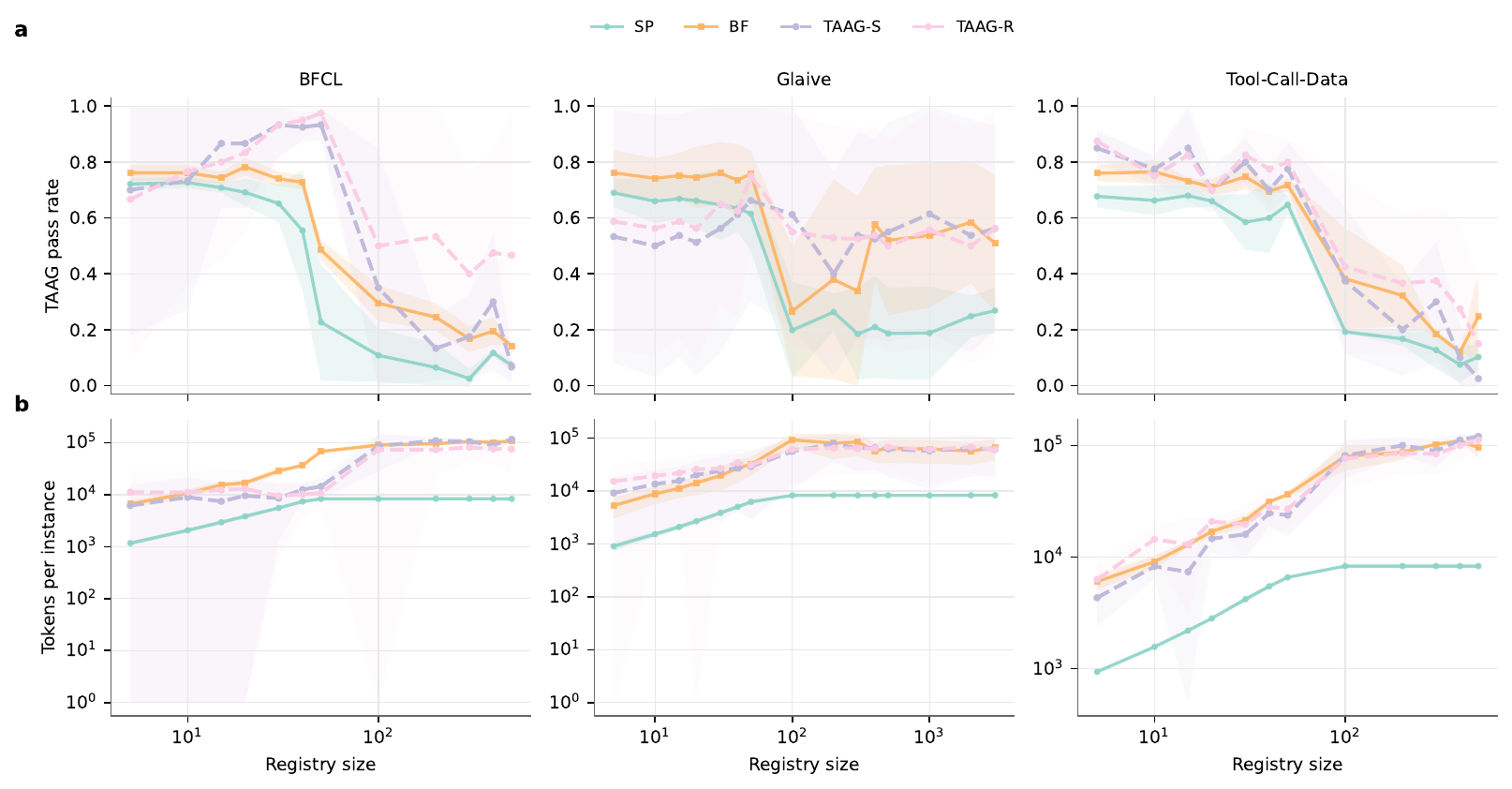}
    \caption{Registry scaling and resource cost across BFCL, Glaive, and Tool-Call-Data. Top row: TAAG pass rate across registry sizes. Bottom row: tokens per instance as a line plot on a log scale, including measured TAAG-S and TAAG-R token curves from the retained structured-feedback summary logs. SP and BF curves are computed from row-level outputs. Lines show means and shaded bands show standard deviations over configuration-level aggregates.}
    \label{fig:scale_cost}
\end{figure*}

\textbf{Registry scale is the dominant degradation.}
Figure~\ref{fig:scale_cost} shows the same pattern on BFCL, Glaive, and Tool-Call-Data: increasing the number of shown tools reduces TAAG pass rate even though the query format and evaluator are fixed. We use small for $S\leq50$, medium for $100\leq S\leq300$, and large for $S\geq400$ in bucketed views; the small-to-large summary tables compare retained small scales against retained large scales $S\geq100$. In SP, pass rate drops from $0.626\pm0.184$ to $0.091\pm0.080$ on BFCL, from $0.654\pm0.110$ to $0.244\pm0.231$ on Glaive, and from $0.649\pm0.069$ to $0.133\pm0.122$ on Tool-Call-Data. BF improves absolute rates but leaves large-registry pass rates low: $0.199\pm0.064$, $0.427\pm0.240$, and $0.242\pm0.121$. RFCD is therefore not only a long-context test; the added context is a larger set of executable candidates whose names, schemas, and requirements compete in the same prompt.

\textbf{Retries repair some intrinsic failures but are expensive.}
BF raises average TAAG pass rate from $0.438\pm0.281$ to $0.586\pm0.247$ and grounding score from $0.574\pm0.332$ to $0.759\pm0.225$, but extrinsic F1 changes only from $0.511\pm0.301$ to $0.540\pm0.296$. This is the diagnostic gap TAAG is meant to expose: a model can repair registry conformance or value grounding without matching the source reference proportionally better. The repair is also costly. BF uses roughly an order of magnitude more tokens than SP, and large-registry BF runs use $4.4\times$ more tokens than small-registry BF runs on BFCL, $4.4\times$ more on Glaive, and $5.0\times$ more on Tool-Call-Data. Raising the BF budget from lower depths to $D_\text{max}=15$ changes average pass rate only from $0.617\pm0.243$ to $0.629\pm0.247$ while token use grows from $15165\pm12165$ to $39286\pm35963$ tokens (Table~\ref{tab:appendix_binary_depth}). More attempts therefore probe recoverability, but they are not a scalable substitute for better registry organization or retrieval.

\begin{table*}[t]
\centering
\footnotesize
\setlength{\tabcolsep}{2.5pt}
\renewcommand{\arraystretch}{1.10}
\begin{adjustbox}{max width=\textwidth}
\begin{tabular}{lllcccc}
\toprule
\multicolumn{7}{l}{\textbf{A. Source-level summary across all feedback modes}} \\
\textbf{Source} & \textbf{Type} & \textbf{Metric} & \textbf{SP} & \textbf{BF} & \textbf{TAAG-S} & \textbf{TAAG-R} \\
\midrule
BFCL & Intrinsic & TAAG pass rate & \cellcolor[HTML]{F2FAEF}0.429 $\pm$ 0.302 & \cellcolor[HTML]{E5F5E0}0.557 $\pm$ 0.265 & \cellcolor[HTML]{E5F5E0}0.579 $\pm$ 0.404 & \cellcolor[HTML]{E5F5E0}0.684 $\pm$ 0.382 \\
 & Extrinsic & Extrinsic F1 & \cellcolor[HTML]{E5F5E0}0.605 $\pm$ 0.353 & \cellcolor[HTML]{E5F5E0}0.628 $\pm$ 0.358 & \cellcolor[HTML]{D9F0D3}0.967 $\pm$ 0.060 & \cellcolor[HTML]{D9F0D3}0.949 $\pm$ 0.067 \\
 & Intrinsic & Grounding score & \cellcolor[HTML]{E5F5E0}0.562 $\pm$ 0.348 & \cellcolor[HTML]{E5F5E0}0.717 $\pm$ 0.251 & \cellcolor[HTML]{F2FAEF}0.490 $\pm$ 0.071 & \cellcolor[HTML]{E5F5E0}0.555 $\pm$ 0.089 \\
 & Cost & Tokens & \cellcolor[HTML]{D9EAF7}5879 $\pm$ 2749 & \cellcolor[HTML]{F3F8FD}48550 $\pm$ 40107 & \cellcolor[HTML]{F3F8FD}48629 $\pm$ 49146 & \cellcolor[HTML]{F3F8FD}39421 $\pm$ 42726 \\
\midrule
Glaive & Intrinsic & TAAG pass rate & \cellcolor[HTML]{F2FAEF}0.446 $\pm$ 0.274 & \cellcolor[HTML]{E5F5E0}0.616 $\pm$ 0.234 & \cellcolor[HTML]{E5F5E0}0.553 $\pm$ 0.390 & \cellcolor[HTML]{E5F5E0}0.573 $\pm$ 0.388 \\
 & Extrinsic & Extrinsic F1 & \cellcolor[HTML]{F2FAEF}0.440 $\pm$ 0.253 & \cellcolor[HTML]{F2FAEF}0.489 $\pm$ 0.228 & \cellcolor[HTML]{E5F5E0}0.650 $\pm$ 0.281 & \cellcolor[HTML]{E5F5E0}0.644 $\pm$ 0.269 \\
 & Intrinsic & Grounding score & \cellcolor[HTML]{E5F5E0}0.550 $\pm$ 0.311 & \cellcolor[HTML]{E5F5E0}0.740 $\pm$ 0.235 & \cellcolor[HTML]{E5F5E0}0.555 $\pm$ 0.071 & \cellcolor[HTML]{E5F5E0}0.620 $\pm$ 0.089 \\
 & Cost & Tokens & \cellcolor[HTML]{D9EAF7}5785 $\pm$ 2877 & \cellcolor[HTML]{F3F8FD}41225 $\pm$ 34903 & \cellcolor[HTML]{F3F8FD}42385 $\pm$ 39026 & \cellcolor[HTML]{F3F8FD}45635 $\pm$ 39938 \\
\midrule
Tool-Call-Data & Intrinsic & TAAG pass rate & \cellcolor[HTML]{F2FAEF}0.434 $\pm$ 0.274 & \cellcolor[HTML]{E5F5E0}0.566 $\pm$ 0.249 & \cellcolor[HTML]{E5F5E0}0.537 $\pm$ 0.327 & \cellcolor[HTML]{E5F5E0}0.586 $\pm$ 0.299 \\
 & Extrinsic & Extrinsic F1 & \cellcolor[HTML]{E5F5E0}0.527 $\pm$ 0.284 & \cellcolor[HTML]{E5F5E0}0.526 $\pm$ 0.307 & \cellcolor[HTML]{D9F0D3}0.819 $\pm$ 0.242 & \cellcolor[HTML]{D9F0D3}0.802 $\pm$ 0.208 \\
 & Intrinsic & Grounding score & \cellcolor[HTML]{E5F5E0}0.630 $\pm$ 0.349 & \cellcolor[HTML]{D9F0D3}0.846 $\pm$ 0.138 & \cellcolor[HTML]{F2FAEF}0.425 $\pm$ 0.089 & \cellcolor[HTML]{F2FAEF}0.490 $\pm$ 0.107 \\
 & Cost & Tokens & \cellcolor[HTML]{D9EAF7}5423 $\pm$ 2873 & \cellcolor[HTML]{F3F8FD}45574 $\pm$ 39017 & \cellcolor[HTML]{F3F8FD}50403 $\pm$ 46326 & \cellcolor[HTML]{F3F8FD}50556 $\pm$ 41564 \\
\midrule
\multicolumn{7}{l}{\textbf{B. Model-family summary across all feedback modes}} \\
\textbf{Model family} & \textbf{Type} & \textbf{Metric} & \textbf{SP} & \textbf{BF} & \textbf{TAAG-S} & \textbf{TAAG-R} \\
\midrule
Instruction-following & Intrinsic & TAAG pass rate & \cellcolor[HTML]{F2FAEF}0.435 $\pm$ 0.259 & \cellcolor[HTML]{E5F5E0}0.522 $\pm$ 0.256 & \cellcolor[HTML]{D9F0D3}0.760 $\pm$ 0.336 & \cellcolor[HTML]{D9F0D3}0.860 $\pm$ 0.313 \\
 & Extrinsic & Extrinsic F1 & \cellcolor[HTML]{E5F5E0}0.562 $\pm$ 0.242 & \cellcolor[HTML]{F2FAEF}0.489 $\pm$ 0.318 & \cellcolor[HTML]{D9F0D3}0.882 $\pm$ 0.176 & \cellcolor[HTML]{D9F0D3}0.965 $\pm$ 0.063 \\
 & Intrinsic & Grounding score & \cellcolor[HTML]{E5F5E0}0.613 $\pm$ 0.289 & \cellcolor[HTML]{D9F0D3}0.763 $\pm$ 0.204 & \cellcolor[HTML]{E5F5E0}0.620 $\pm$ 0.089 & \cellcolor[HTML]{E5F5E0}0.685 $\pm$ 0.107 \\
 & Cost & Tokens & \cellcolor[HTML]{D9EAF7}5721 $\pm$ 2822 & \cellcolor[HTML]{F3F8FD}53863 $\pm$ 39537 & \cellcolor[HTML]{F3F8FD}35482 $\pm$ 46448 & \cellcolor[HTML]{E8F2FB}19573 $\pm$ 39169 \\
\midrule
Reasoning/thinking & Intrinsic & TAAG pass rate & \cellcolor[HTML]{F2FAEF}0.371 $\pm$ 0.280 & \cellcolor[HTML]{E5F5E0}0.560 $\pm$ 0.248 & \cellcolor[HTML]{F2FAEF}0.425 $\pm$ 0.107 & \cellcolor[HTML]{F2FAEF}0.490 $\pm$ 0.035 \\
 & Extrinsic & Extrinsic F1 & \cellcolor[HTML]{F2FAEF}0.387 $\pm$ 0.317 & \cellcolor[HTML]{E5F5E0}0.563 $\pm$ 0.286 & \cellcolor[HTML]{F2FAEF}0.425 $\pm$ 0.107 & \cellcolor[HTML]{F2FAEF}0.490 $\pm$ 0.035 \\
 & Intrinsic & Grounding score & \cellcolor[HTML]{F2FAEF}0.427 $\pm$ 0.331 & \cellcolor[HTML]{E5F5E0}0.649 $\pm$ 0.302 & \cellcolor[HTML]{F2FAEF}0.425 $\pm$ 0.053 & \cellcolor[HTML]{F2FAEF}0.490 $\pm$ 0.071 \\
 & Cost & Tokens & \cellcolor[HTML]{D9EAF7}5679 $\pm$ 2673 & \cellcolor[HTML]{F3F8FD}39728 $\pm$ 38451 & \cellcolor[HTML]{FFF7EC}124784 $\pm$ 1800 & \cellcolor[HTML]{FFF7EC}123496 $\pm$ 2525 \\
\midrule
Tool/structured-output & Intrinsic & TAAG pass rate & \cellcolor[HTML]{F2FAEF}0.476 $\pm$ 0.301 & \cellcolor[HTML]{E5F5E0}0.689 $\pm$ 0.199 & \cellcolor[HTML]{E5F5E0}0.700 $\pm$ 0.071 & \cellcolor[HTML]{E5F5E0}0.550 $\pm$ 0.071 \\
 & Extrinsic & Extrinsic F1 & \cellcolor[HTML]{E5F5E0}0.522 $\pm$ 0.332 & \cellcolor[HTML]{E5F5E0}0.603 $\pm$ 0.257 & \cellcolor[HTML]{E5F5E0}0.600 $\pm$ 0.071 & \cellcolor[HTML]{E5F5E0}0.537 $\pm$ 0.053 \\
 & Intrinsic & Grounding score & \cellcolor[HTML]{E5F5E0}0.608 $\pm$ 0.356 & \cellcolor[HTML]{D9F0D3}0.799 $\pm$ 0.206 & \cellcolor[HTML]{F2FAEF}0.490 $\pm$ 0.107 & \cellcolor[HTML]{E5F5E0}0.555 $\pm$ 0.035 \\
 & Cost & Tokens & \cellcolor[HTML]{D9EAF7}5745 $\pm$ 2948 & \cellcolor[HTML]{F3F8FD}32880 $\pm$ 30181 & \cellcolor[HTML]{E8F2FB}19834 $\pm$ 2525 & \cellcolor[HTML]{F3F8FD}30349 $\pm$ 1226 \\
\bottomrule
\end{tabular}
\end{adjustbox}
\caption{Main feedback summary with metric type made explicit. Intrinsic values are TAAG-derived diagnostics computed without the ground-truth call, extrinsic values compare the final call with the source ground truth, and cost values measure resource use. SP denotes single pass, BF denotes binary feedback, TAAG-S denotes stage-only TAAG feedback, and TAAG-R denotes requirement-level TAAG feedback. Panel A reports source-level SP/BF row-level aggregates together with retained TAAG-S/TAAG-R summary-log aggregates. Panel B reports model-family SP/BF row-level aggregates together with retained TAAG-S/TAAG-R row-level aggregates. Values are means with standard deviations; token values are per instance. Green cells mark higher-is-better score metrics, blue cells mark lower-is-better cost metrics, and grey cells mark unavailable retained metrics.}
\label{tab:main_feedback_summary}
\end{table*}

\textbf{The stage view identifies the bottleneck.}
Table~\ref{tab:stage_pass_summary} shows that low final pass rate should not be read as a single generic grounding failure. BFCL is most clearly grounding-limited because Q3 is lowest in both SP and BF. Glaive is mixed: SP is slightly more schema-limited, while BF leaves Q3 as the lowest remaining stage. Tool-Call-Data is schema-limited because Q2 is lower than Q3 in both SP and BF. Thus, the same end-to-end failure rate can imply different interventions: registry retrieval for name failures, schema constraints for Q2 failures, or better value grounding for Q3 failures.

\begin{table*}[h]
\centering
\small
\setlength{\tabcolsep}{3.5pt}
\renewcommand{\arraystretch}{1.12}
\begin{tabular}{llcccc}
\toprule
\textbf{Source} & \textbf{Mode} & \textbf{Q1 Registry} & \textbf{Q2 Schema} & \textbf{Q3 Grounding} & \textbf{Lowest stage} \\
\midrule
BFCL & SP & \cellcolor[HTML]{E5F5E0}0.659 $\pm$ 0.382 & \cellcolor[HTML]{E5F5E0}0.558 $\pm$ 0.362 & \cellcolor[HTML]{E5F5E0}0.501 $\pm$ 0.311 & Q3 Grounding \\
BFCL & BF & \cellcolor[HTML]{D9F0D3}0.990 $\pm$ 0.014 & \cellcolor[HTML]{D9F0D3}0.858 $\pm$ 0.121 & \cellcolor[HTML]{E5F5E0}0.655 $\pm$ 0.242 & Q3 Grounding \\
Glaive & SP & \cellcolor[HTML]{E5F5E0}0.638 $\pm$ 0.363 & \cellcolor[HTML]{E5F5E0}0.526 $\pm$ 0.312 & \cellcolor[HTML]{E5F5E0}0.532 $\pm$ 0.305 & Q2 Schema \\
Glaive & BF & \cellcolor[HTML]{D9F0D3}0.884 $\pm$ 0.248 & \cellcolor[HTML]{E5F5E0}0.747 $\pm$ 0.230 & \cellcolor[HTML]{E5F5E0}0.725 $\pm$ 0.237 & Q3 Grounding \\
Tool-Call-Data & SP & \cellcolor[HTML]{E5F5E0}0.669 $\pm$ 0.365 & \cellcolor[HTML]{F2FAEF}0.481 $\pm$ 0.293 & \cellcolor[HTML]{E5F5E0}0.608 $\pm$ 0.338 & Q2 Schema \\
Tool-Call-Data & BF & \cellcolor[HTML]{D9F0D3}0.975 $\pm$ 0.025 & \cellcolor[HTML]{E5F5E0}0.706 $\pm$ 0.174 & \cellcolor[HTML]{D9F0D3}0.817 $\pm$ 0.149 & Q2 Schema \\
\bottomrule
\end{tabular}
\caption{Intrinsic TAAG stage-pass rates for the retained single-pass and binary-feedback row outputs. Q1 checks whether the emitted function name appears in the registry, Q2 checks whether the selected schema is structurally satisfied, and Q3 checks whether argument values are grounded in the user request. Values are means with standard deviations over configuration-level aggregates. The lowest-stage column marks the smallest average pass rate in that source--mode row, which identifies the main observed bottleneck without implying that downstream failures caused upstream failures.}
\label{tab:stage_pass_summary}
\end{table*}

\textbf{Model family and structured feedback effects are secondary to scale.}
Table~\ref{tab:main_feedback_summary} shows model-family differences, with the tool/structured-output group obtaining the strongest BF pass rate ($0.689\pm0.199$), followed by reasoning/thinking models ($0.560\pm0.248$) and instruction-following models ($0.522\pm0.256$). None removes the scale drop in Figure~\ref{fig:scale_cost}, so the result is not merely a model ranking: the registry interface itself becomes a bottleneck when many plausible tools are shown together.

Structured feedback is useful as an actionability probe, not as a complete fix. TAAG-R improves TAAG pass rate over TAAG-S on all three sources: BFCL rises from $0.579\pm0.404$ to $0.684\pm0.382$, Glaive from $0.553\pm0.390$ to $0.573\pm0.388$, and Tool-Call-Data from $0.537\pm0.327$ to $0.586\pm0.299$. These gains indicate that requirement-level diagnostics can help models act on a failed obligation, but they are modest compared with the registry-scale degradation and do not consistently reduce token cost. The evaluation contribution is therefore the benchmark and stage signal that reveal where tool-calling systems fail, not a claim that prompting alone solves registry scaling.


\section{Related Work}
\label{sec:related_work}

\textbf{TA-LLMs and Existing Benchmarks.}
Early work in TA-LLMs demonstrated that LLMs can invoke tools and symbolic modules through prompting~\citep{schick2023toolformer,nakano2021webgpt,karpas2022mrkl}, with subsequent approaches enabling tool use via fine-tuning~\citep{patil2024gorilla,qin2024toolllm}, reinforcement learning~\citep{chen2023fireact}, and in-context reasoning strategies such as ReAct~\citep{yao2023react,shen2023hugginggpt}. More recent work explores multi-tool chaining~\citep{lu2023chameleon,gao2023pal,paranjape2023art} and hierarchical planning over tool graphs~\citep{wu2023taskmatrix,liang2023taskbench}. These approaches implicitly assume that the correct tool has already been selected, focusing primarily on argument generation and execution. This assumption becomes increasingly limiting as real-world systems shift toward Model Context Protocol (MCP)-style deployments~\citep{anthropic2024mcp}, where a single LLM must orchestrate across large, heterogeneous tool registries, poising \emph{tool selection} as a primary challenge preceding downstream execution.

Existing benchmarks for tool-augmented LLMs, including API-Bank~\citep{li2023apibank}, ToolLLM~\citep{qin2024toolllm}, Gorilla~\citep{patil2024gorilla}, ToolACE~\citep{liu2024toolace}, APIGenT~\citep{prabhakar2025apigen}, and BFCL~\citep{patil2025bfcl} among others, have established standardized protocols for benchmarking tool use. However, these datasets predominantly operate in constrained settings with closed tool sets, typically single-shot execution, reducing the problem to mapping user intent into a known schema. While useful for measuring execution success, they do not capture the challenges of large-scale registries where thousands of candidate functions may be present. As modern deployments scale toward hundreds or thousands of tools, this gap becomes critical. RFCD directly targets this limitation by constructing large-scale training data that explicitly models tool selection under expanding registries.

\textbf{Tool Hallucinations and Implicit Evaluations.}
Tool hallucination has emerged as a fundamental challenge in reliable tool use. Prior work distinguishes between \emph{tool selection hallucination} and \emph{tool usage hallucination}~\citep{xu2025reducing}, reflecting failures at different stages of the tool-calling pipeline. Existing approaches mitigate these issues through process-level reward models~\citep{qian2025toolrlrewardtoollearning} and self-reflection mechanisms~\citep{masterman2024landscapeemergingaiagent}. However, evaluation remains predominantly \emph{extrinsic}, relying on aggregate metrics such as exact-match accuracy~\citep{patil2024gorilla}, AST similarity~\citep{yan2024bfcl}, and LLM-as-judge pass rates~\citep{qin2024toolllm}. These metrics collapse multiple failure modes into a single scalar outcome, providing little insight into \emph{where} or \emph{why} a model fails. In large tool registries, an incorrect prediction may arise from failure to retrieve the correct tool, misinterpretation of the schema, incorrect argument grounding, or hallucination of a non-existent tool, yet all are treated equivalently.

Iterative reasoning frameworks such as ReAct~\citep{yao2022react} partially address this by enabling models to refine outputs through multi-step interaction, but the feedback remains implicit and undifferentiated, relying on the model's own reflection rather than structured signals about specific failure modes. In contrast, TAAG introduces an \emph{implicit evaluation} framework that decomposes tool-calling performance into three interpretable dimensions: registry conformance, structural completeness, and argument grounding. By evaluating these aspects independently, TAAG enables fine-grained identification of failure sources, offering a more transparent and actionable alternative to purely extrinsic metrics and enabling targeted correction strategies grounded in specific model deficiencies.

\section{Conclusion}
We presented RGFC's two components: RFCD, a registry-scaled function-calling benchmark, and TAAG, a stage-decomposed evaluator for registry conformance, structural completeness, and argument grounding. Across three dataset sources and six locally served small models, performance declines as registry size increases, while feedback loops add substantial token and time cost. BF improves TAAG pass rate and grounding on average, but the gains are smaller than the registry-scale degradation and do not always translate into extrinsic F1. TAAG-S and TAAG-R show that TAAG diagnostics can be used as actionability probes, with requirement-level feedback providing modest additional improvement over stage-only feedback. These results suggest that tool-calling reliability cannot be reduced to larger prompts or more retries alone; scalable systems also need better registry organization, retrieval, and grounded interfaces.

\bibliographystyle{unsrt}
\bibliography{custom}

@article{bubeck2023sparks,
  title={Sparks of Artificial General Intelligence: Early experiments with GPT-4},
  author={Bubeck, S{\'e}bastien and Chandrasekaran, Varun and Eldan, Ronen and Gehrke, Johannes and Horvitz, Eric and Kamar, Ece and Lee, Peter and Lee, Yin Tat and Li, Yuanzhi and Lundberg, Scott and others},
  journal={arXiv preprint arXiv:2303.12712},
  year={2023}
}

@article{naveed2025comprehensive,
  title={A comprehensive overview of large language models},
  author={Naveed, Humza and Khan, Asad Ullah and Qiu, Shi and Saqib, Muhammad and Anwar, Saeed and Usman, Muhammad and Akhtar, Naveed and Barnes, Nick and Mian, Ajmal},
  journal={ACM Transactions on Intelligent Systems and Technology},
  volume={16},
  number={5},
  pages={1--72},
  year={2025},
  publisher={ACM New York, NY}
}

@article{schick2023toolformer,
  title={Toolformer: Language models can teach themselves to use tools},
  author={Schick, Timo and Dwivedi-Yu, Jane and Dess{\`\i}, Roberto and Raileanu, Roberta and Lomeli, Maria and Hambro, Eric and Zettlemoyer, Luke and Cancedda, Nicola and Scialom, Thomas},
  journal={Advances in neural information processing systems},
  volume={36},
  pages={68539--68551},
  year={2023}
}

@article{qin2024tool,
  title={Tool learning with foundation models},
  author={Qin, Yujia and Hu, Shengding and Lin, Yankai and Chen, Weize and Ding, Ning and Cui, Ganqu and Zeng, Zheni and Zhou, Xuanhe and Huang, Yufei and Xiao, Chaojun and others},
  journal={ACM Computing Surveys},
  volume={57},
  number={4},
  pages={1--40},
  year={2024},
  publisher={ACM New York, NY}
}

@article{mialon2023augmented,
  title={Augmented language models: a survey},
  author={Mialon, Gr{\'e}goire and Dess{\`\i}, Roberto and Lomeli, Maria and Nalmpantis, Christoforos and Pasunuru, Ram and Raileanu, Roberta and Rozi{\`e}re, Baptiste and Schick, Timo and Dwivedi-Yu, Jane and Celikyilmaz, Asli and others},
  journal={arXiv preprint arXiv:2302.07842},
  year={2023}
}

@article{plaat2025agentic,
  title={Agentic large language models, a survey},
  author={Plaat, Aske and van Duijn, Max and Van Stein, Niki and Preuss, Mike and van der Putten, Peter and Batenburg, Kees Joost},
  journal={Journal of Artificial Intelligence Research},
  volume={84},
  year={2025}
}

@article{xi2025rise,
  title={The rise and potential of large language model based agents: A survey},
  author={Xi, Zhiheng and Chen, Wenxiang and Guo, Xin and He, Wei and Ding, Yiwen and Hong, Boyang and Zhang, Ming and Wang, Junzhe and Jin, Senjie and Zhou, Enyu and others},
  journal={Science China Information Sciences},
  volume={68},
  number={2},
  pages={121101},
  year={2025},
  publisher={Springer}
}

@article{wang2024survey,
  title={A survey on large language model based autonomous agents},
  author={Wang, Lei and Ma, Chen and Feng, Xueyang and Zhang, Zeyu and Yang, Hao and Zhang, Jingsen and Chen, Zhiyuan and Tang, Jiakai and Chen, Xu and Lin, Yankai and others},
  journal={Frontiers of Computer Science},
  volume={18},
  number={6},
  pages={186345},
  year={2024},
  publisher={Springer}
}

@article{gao2023retrieval,
  title={Retrieval-Augmented Generation for Large Language Models: A Survey},
  author={Gao, Yunfan and Xiong, Yun and Gao, Xinyu and Jia, Kangxiang and Pan, Jinliu and Bi, Yuxi and Dai, Yi and Sun, Jiawei and Wang, Meng and Wang, Haofen},
  journal={arXiv preprint arXiv:2312.10997},
  year={2023}
}

@article{wei2025autotir,
  title={Autotir: Autonomous tools integrated reasoning via reinforcement learning},
  author={Wei, Yifan and Yu, Xiaoyan and Weng, Yixuan and Pan, Tengfei and Li, Angsheng and Du, Li},
  journal={arXiv preprint arXiv:2507.21836},
  year={2025}
}

@article{zhang2025landscape,
  title={The landscape of agentic reinforcement learning for llms: A survey},
  author={Zhang, Guibin and Geng, Hejia and Yu, Xiaohang and Yin, Zhenfei and Zhang, Zaibin and Tan, Zelin and Zhou, Heng and Li, Zhongzhi and Xue, Xiangyuan and Li, Yijiang and others},
  journal={arXiv preprint arXiv:2509.02547},
  year={2025}
}

@inproceedings{shi2025retrieval,
    title = "Retrieval Models Aren{'}t Tool-Savvy: Benchmarking Tool Retrieval for Large Language Models",
    author = "Shi, Zhengliang  and
      Wang, Yuhan  and
      Yan, Lingyong  and
      Ren, Pengjie  and
      Wang, Shuaiqiang  and
      Yin, Dawei  and
      Ren, Zhaochun",
    editor = "Che, Wanxiang  and
      Nabende, Joyce  and
      Shutova, Ekaterina  and
      Pilehvar, Mohammad Taher",
    booktitle = "Findings of the Association for Computational Linguistics: ACL 2025",
    month = jul,
    year = "2025",
    address = "Vienna, Austria",
    publisher = "Association for Computational Linguistics",
    url = "https://aclanthology.org/2025.findings-acl.1258/",
    doi = "10.18653/v1/2025.findings-acl.1258",
    pages = "24497--24524",
    ISBN = "979-8-89176-256-5"
}

@inproceedings{levy2024same,
  title={Same task, more tokens: the impact of input length on the reasoning performance of large language models},
  author={Levy, Mosh and Jacoby, Alon and Goldberg, Yoav},
  booktitle={Proceedings of the 62nd Annual Meeting of the Association for Computational Linguistics (Volume 1: Long Papers)},
  pages={15339--15353},
  year={2024}
}

@inproceedings{modarressi2025nolima,
  title={NoLiMa: Long-Context Evaluation Beyond Literal Matching},
  author={Modarressi, Ali and Deilamsalehy, Hanieh and Dernoncourt, Franck and Bui, Trung and Rossi, Ryan A and Yoon, Seunghyun and Schuetze, Hinrich},
  booktitle={International Conference on Machine Learning},
  pages={44554--44570},
  year={2025},
  organization={PMLR}
}

@inproceedings{paramanayakam2025less,
  title={Less is more: Optimizing function calling for llm execution on edge devices},
  author={Paramanayakam, Varatheepan and Karatzas, Andreas and Anagnostopoulos, Iraklis and Stamoulis, Dimitrios},
  booktitle={2025 Design, Automation \& Test in Europe Conference (DATE)},
  pages={1--7},
  year={2025},
  organization={IEEE}
}

@article{blankenstein2026biasbusters,
  title={BiasBusters: Uncovering and Mitigating Tool Selection Bias in Large Language Models},
  author={Blankenstein, Thierry and Yu, Jialin and Li, Zixuan and Plachouras, Vassilis and Sengupta, Sunando and Torr, Philip and Gal, Yarin and Paren, Alasdair and Bibi, Adel},
  journal={arXiv preprint arXiv:2510.00307},
  year={2025}
}

@inproceedings{faghih2025tool,
    title = "Tool Preferences in Agentic {LLM}s are Unreliable",
    author = "Faghih, Kazem  and
      Wang, Wenxiao  and
      Cheng, Yize  and
      Bharti, Siddhant  and
      Sriramanan, Gaurang  and
      Balasubramanian, Sriram  and
      Hosseini, Parsa  and
      Feizi, Soheil",
    editor = "Christodoulopoulos, Christos  and
      Chakraborty, Tanmoy  and
      Rose, Carolyn  and
      Peng, Violet",
    booktitle = "Proceedings of the 2025 Conference on Empirical Methods in Natural Language Processing",
    month = nov,
    year = "2025",
    address = "Suzhou, China",
    publisher = "Association for Computational Linguistics",
    url = "https://aclanthology.org/2025.emnlp-main.1060/",
    doi = "10.18653/v1/2025.emnlp-main.1060",
    pages = "20954--20969",
    ISBN = "979-8-89176-332-6"
}

@inproceedings{li2023apibank,
    title = "{API}-Bank: A Comprehensive Benchmark for Tool-Augmented {LLM}s",
    author = "Li, Minghao  and
      Zhao, Yingxiu  and
      Yu, Bowen  and
      Song, Feifan  and
      Li, Hangyu  and
      Yu, Haiyang  and
      Li, Zhoujun  and
      Huang, Fei  and
      Li, Yongbin",
    editor = "Bouamor, Houda  and
      Pino, Juan  and
      Bali, Kalika",
    booktitle = "Proceedings of the 2023 Conference on Empirical Methods in Natural Language Processing",
    month = dec,
    year = "2023",
    address = "Singapore",
    publisher = "Association for Computational Linguistics",
    url = "https://aclanthology.org/2023.emnlp-main.187/",
    doi = "10.18653/v1/2023.emnlp-main.187",
    pages = "3102--3116"
}

@inproceedings{qin2024toolllm,
  author       = {Yujia Qin and
                  Shihao Liang and
                  Yining Ye and
                  Kunlun Zhu and
                  Lan Yan and
                  Yaxi Lu and
                  Yankai Lin and
                  Xin Cong and
                  Xiangru Tang and
                  Bill Qian and
                  Sihan Zhao and
                  Lauren Hong and
                  Runchu Tian and
                  Ruobing Xie and
                  Jie Zhou and
                  Mark Gerstein and
                  Dahai Li and
                  Zhiyuan Liu and
                  Maosong Sun},
  title        = {ToolLLM: Facilitating Large Language Models to Master 16000+ Real-world
                  APIs},
  booktitle    = {The Twelfth International Conference on Learning Representations,
                  {ICLR} 2024, Vienna, Austria, May 7-11, 2024},
  publisher    = {OpenReview.net},
  year         = {2024},
  url          = {https://openreview.net/forum?id=dHng2O0Jjr},
  bibsource    = {dblp computer science bibliography, https://dblp.org}
}

@inproceedings{patil2024gorilla,
 author = {Patil, Shishir G. and Zhang, Tianjun and Wang, Xin and Gonzalez, Joseph E.},
 booktitle = {Advances in Neural Information Processing Systems},
 doi = {10.52202/079017-4020},
 editor = {A. Globerson and L. Mackey and D. Belgrave and A. Fan and U. Paquet and J. Tomczak and C. Zhang},
 pages = {126544--126565},
 publisher = {Curran Associates, Inc.},
 title = {Gorilla: Large Language Model Connected with Massive APIs},
 url = {https://proceedings.neurips.cc/paper_files/paper/2024/file/e4c61f578ff07830f5c37378dd3ecb0d-Paper-Conference.pdf},
 volume = {37},
 year = {2024}
}

@InProceedings{patil2025bfcl,
  title = 	 {The Berkeley Function Calling Leaderboard ({BFCL}): From Tool Use to Agentic Evaluation of Large Language Models},
  author =       {Patil, Shishir G and Mao, Huanzhi and Yan, Fanjia and Ji, Charlie Cheng-Jie and Suresh, Vishnu and Stoica, Ion and Gonzalez, Joseph E.},
  booktitle = 	 {Proceedings of the 42nd International Conference on Machine Learning},
  pages = 	 {48371--48392},
  year = 	 {2025},
  editor = 	 {Singh, Aarti and Fazel, Maryam and Hsu, Daniel and Lacoste-Julien, Simon and Berkenkamp, Felix and Maharaj, Tegan and Wagstaff, Kiri and Zhu, Jerry},
  volume = 	 {267},
  series = 	 {Proceedings of Machine Learning Research},
  month = 	 {13--19 Jul},
  publisher =    {PMLR},
  url = 	 {https://proceedings.mlr.press/v267/patil25a.html}
}

@inproceedings{xu2025reducing,
  title={Reducing Tool Hallucination via Reliability Alignment},
  author={Xu, Hongshen and Zhu, Zichen and Pan, Lei and Wang, Zihan and Zhu, Su and Ma, Da and Cao, Ruisheng and Chen, Lu and Yu, Kai},
  booktitle={International Conference on Machine Learning},
  pages={69992--70006},
  year={2025},
  organization={PMLR}
}

@inproceedings{zhang2024toolbehonest,
    title = "{T}ool{B}e{H}onest: A Multi-level Hallucination Diagnostic Benchmark for Tool-Augmented Large Language Models",
    author = "Zhang, Yuxiang  and
      Chen, Jing  and
      Wang, Junjie  and
      Liu, Yaxin  and
      Yang, Cheng  and
      Shi, Chufan  and
      Zhu, Xinyu  and
      Lin, Zihao  and
      Wan, Hanwen  and
      Yang, Yujiu  and
      Sakai, Tetsuya  and
      Feng, Tian  and
      Yamana, Hayato",
    editor = "Al-Onaizan, Yaser  and
      Bansal, Mohit  and
      Chen, Yun-Nung",
    booktitle = "Proceedings of the 2024 Conference on Empirical Methods in Natural Language Processing",
    month = nov,
    year = "2024",
    address = "Miami, Florida, USA",
    publisher = "Association for Computational Linguistics",
    url = "https://aclanthology.org/2024.emnlp-main.637/",
    doi = "10.18653/v1/2024.emnlp-main.637",
    pages = "11388--11422"
}

@inproceedings{yao2023react,
  title     = {ReAct: Synergizing Reasoning and Acting in Language Models},
  author    = {Yao, Shunyu and Zhao, Jeffrey and Yu, Dian and Du, Nan and Shafran, Izhak and Narasimhan, Karthik R. and Cao, Yuan},
  booktitle = {The Eleventh International Conference on Learning Representations},
  year      = {2023},
  url       = {https://openreview.net/forum?id=WE_vluYUL-X}
}

@inproceedings{yao2022react,
  author = {Yao, Shunyu and Zhao, Jeffrey and Yu, Dian and Du, Nan and Shafran, Izhak and Narasimhan, Karthik R. and Cao, Yuan},
  booktitle = {ICLR},
  ee = {https://openreview.net/forum?id=WE_vluYUL-X},
  publisher = {OpenReview.net},
  title = {ReAct: Synergizing Reasoning and Acting in Language Models.},
  url = {http://dblp.uni-trier.de/db/conf/iclr/iclr2023.html#YaoZYDSN023},
  year = 2023
}

@misc{glaiveai2023functioncalling,
  author       = {{Glaive AI}},
  title        = {glaive-function-calling-v2},
  year         = {2023},
  publisher    = {Hugging Face},
  howpublished = {\url{https://huggingface.co/datasets/glaiveai/glaive-function-calling-v2}},
}

@misc{qian2025toolrlrewardtoollearning,
      title={ToolRL: Reward is All Tool Learning Needs}, 
      author={Cheng Qian and Emre Can Acikgoz and Qi He and Hongru Wang and Xiusi Chen and Dilek Hakkani-Tür and Gokhan Tur and Heng Ji},
      year={2025},
      eprint={2504.13958},
      archivePrefix={arXiv},
      primaryClass={cs.LG},
      url={https://arxiv.org/abs/2504.13958}, 
}

@misc{masterman2024landscapeemergingaiagent,
      title={The Landscape of Emerging AI Agent Architectures for Reasoning, Planning, and Tool Calling: A Survey}, 
      author={Tula Masterman and Sandi Besen and Mason Sawtell and Alex Chao},
      year={2024},
      eprint={2404.11584},
      archivePrefix={arXiv},
      primaryClass={cs.AI},
      url={https://arxiv.org/abs/2404.11584}, 
}

@article{gan2025rag,
  title={Rag-mcp: Mitigating prompt bloat in llm tool selection via retrieval-augmented generation},
  author={Gan, Tiantian and Sun, Qiyao},
  journal={arXiv preprint arXiv:2505.03275},
  year={2025}
}

@misc{anthropic2024mcp,
  title     = {Introducing the Model Context Protocol},
  author    = {{Anthropic}},
  year      = {2024},
  howpublished = {\url{https://www.anthropic.com/news/model-context-protocol}}
}

@misc{rapidfuzz,
  author       = {Max Bachmann},
  title        = {rapidfuzz/RapidFuzz: Release 3.13.0},
  month        = apr,
  year         = 2025,
  publisher    = {Zenodo},
  version      = {v3.13.0},
  doi          = {10.5281/zenodo.15133267},
  url          = {https://doi.org/10.5281/zenodo.15133267},
}

@misc{meta2024llama32modelcard,
  title = {Llama 3.2 3B Instruct Model Card},
  author = {{Meta}},
  year = {2024},
  howpublished = {\url{https://huggingface.co/meta-llama/Llama-3.2-3B-Instruct}},
}

@misc{qwen2024qwen25,
  title = {Qwen2.5: A Party of Foundation Models},
  author = {{Qwen Team}},
  year = {2024},
  month = {September},
  howpublished = {\url{https://qwenlm.github.io/blog/qwen2.5/}},
}

@misc{google2026gemma4,
  title = {Gemma 4: Byte for byte, the most capable open models},
  author = {Farabet, Clement and Lacombe, Olivier},
  year = {2026},
  howpublished = {\url{https://blog.google/innovation-and-ai/technology/developers-tools/gemma-4/}},
}

@article{qwen2025qwen3technicalreport,
  title={Qwen3 technical report},
  author={Yang, An and Li, Anfeng and Yang, Baosong and Zhang, Beichen and Hui, Binyuan and Zheng, Bo and Yu, Bowen and Gao, Chang and Huang, Chengen and Lv, Chenxu and others},
  journal={arXiv preprint arXiv:2505.09388},
  year={2025}
}

@article{microsoft2025phi4mini,
  title={Phi-4-mini technical report: Compact yet powerful multimodal language models via mixture-of-loras},
  author={Abouelenin, Abdelrahman and Ashfaq, Atabak and Atkinson, Adam and Awadalla, Hany and Bach, Nguyen and Bao, Jianmin and Benhaim, Alon and Cai, Martin and Chaudhary, Vishrav and Chen, Congcong and others},
  journal={arXiv preprint arXiv:2503.01743},
  year={2025}
}

@misc{ibm2025granite4,
  title = {Granite 4.0 Models},
  author = {{IBM}},
  year = {2025},
  howpublished = {\url{https://www.ibm.com/granite/docs/models/granite}},
}

@article{nakano2021webgpt,
  title={Webgpt: Browser-assisted question-answering with human feedback},
  author={Nakano, Reiichiro and Hilton, Jacob and Balaji, Suchir and Wu, Jeff and Ouyang, Long and Kim, Christina and Hesse, Christopher and Jain, Shantanu and Kosaraju, Vineet and Saunders, William and others},
  journal={arXiv preprint arXiv:2112.09332},
  year={2021}
}

@article{karpas2022mrkl,
  title={MRKL Systems: A modular, neuro-symbolic architecture that combines large language models, external knowledge sources and discrete reasoning},
  author={Karpas, Ehud and Abend, Omri and Berant, Jonathan and Goldberg, Yoav and Haviv, Adi and Herzig, Jonathan and Nachshon, Amir and Wolfson, Tomer and Lappin, Shalom},
  journal={arXiv preprint arXiv:2205.00445},
  year={2022}
}

@article{chen2023fireact,
  title={FireAct: Toward Language Agent Fine-tuning},
  author={Chen, Baian and Shu, Chang and Shareghi, Ehsan and Collier, Nigel and Narasimhan, Karthik and Yao, Shunyu},
  journal={arXiv preprint arXiv:2310.05915},
  year={2023}
}

@inproceedings{
shen2023hugginggpt,
title={Hugging{GPT}: Solving {AI} Tasks with Chat{GPT} and its Friends in Hugging Face},
author={Yongliang Shen and Kaitao Song and Xu Tan and Dongsheng Li and Weiming Lu and Yueting Zhuang},
booktitle={Thirty-seventh Conference on Neural Information Processing Systems},
year={2023},
url={https://openreview.net/forum?id=yHdTscY6Ci}
}

@inproceedings{lu2023chameleon,
 author = {Lu, Pan and Peng, Baolin and Cheng, Hao and Galley, Michel and Chang, Kai-Wei and Wu, Ying Nian and Zhu, Song-Chun and Gao, Jianfeng},
 booktitle = {Advances in Neural Information Processing Systems},
 editor = {A. Oh and T. Naumann and A. Globerson and K. Saenko and M. Hardt and S. Levine},
 pages = {43447--43478},
 publisher = {Curran Associates, Inc.},
 title = {Chameleon: Plug-and-Play Compositional Reasoning with Large Language Models},
 url = {https://proceedings.neurips.cc/paper_files/paper/2023/file/871ed095b734818cfba48db6aeb25a62-Paper-Conference.pdf},
 volume = {36},
 year = {2023}
}

@inproceedings{gao2023pal,
author = {Gao, Luyu and Madaan, Aman and Zhou, Shuyan and Alon, Uri and Liu, Pengfei and Yang, Yiming and Callan, Jamie and Neubig, Graham},
title = {PAL: program-aided language models},
year = {2023},
publisher = {JMLR.org},
booktitle = {Proceedings of the 40th International Conference on Machine Learning},
articleno = {435},
numpages = {36},
location = {Honolulu, Hawaii, USA},
series = {ICML'23}
}

@misc{vikhrmodels_tool_plannings_v0_2,
  author       = {Dmitry Tishencko and Sergei Bratchikov and Aleksandr Nikolich and Vikhr Team},
  title        = {tool-plannings-v0.2},
  year         = {2025},
  publisher    = {Hugging Face},
  howpublished = {\url{https://huggingface.co/datasets/Vikhrmodels/tool-plannings-v0.2}},
  note         = {English-Russian synthetic dataset for function calling and tool planning}
}

@article{paranjape2023art,
  title={Art: Automatic multi-step reasoning and tool-use for large language models},
  author={Paranjape, Bhargavi and Lundberg, Scott and Singh, Sameer and Hajishirzi, Hannaneh and Zettlemoyer, Luke and Ribeiro, Marco Tulio},
  journal={arXiv preprint arXiv:2303.09014},
  year={2023}
}

@article{wu2023taskmatrix,
  title={Taskmatrix. ai: Completing tasks by connecting foundation models with millions of apis},
  author={Liang, Yaobo and Wu, Chenfei and Song, Ting and Wu, Wenshan and Xia, Yan and Liu, Yu and Ou, Yang and Lu, Shuai and Ji, Lei and Mao, Shaoguang and others},
  journal={Intelligent Computing},
  volume={3},
  pages={0063},
  year={2024},
  publisher={AAAS}
}

@article{liang2023taskbench,
  title={Taskbench: Benchmarking large language models for task automation},
  author={Shen, Yongliang and Song, Kaitao and Tan, Xu and Zhang, Wenqi and Ren, Kan and Yuan, Siyu and Lu, Weiming and Li, Dongsheng and Zhuang, Yueting},
  journal={Advances in Neural Information Processing Systems},
  volume={37},
  pages={4540--4574},
  year={2024}
}

@article{yan2024bfcl,
  title={Berkeley Function-Calling Leaderboard},
  author={Yan, Fanjia and Mao, Huanzhi and Ji, Charlie Cheng-Jie and Zhang, Tianjun and Patil, Shishir G and Gonzalez, Joseph E and Stoica, Ion},
  journal={arXiv preprint arXiv:2402.02659},
  year={2024}
}

@article{liu2024toolace,
  title={Toolace: Winning the points of llm function calling},
  author={Liu, Weiwen and Huang, Xu and Zeng, Xingshan and Hao, Xinlong and Yu, Shuai and Li, Dexun and Wang, Shuai and Gan, Weinan and Liu, Zhengying and Yu, Yuanqing and others},
  journal={arXiv preprint arXiv:2409.00920},
  year={2024}
}

@article{prabhakar2025apigen,
  title={Apigen-mt: Agentic pipeline for multi-turn data generation via simulated agent-human interplay},
  author={Prabhakar, Akshara and Liu, Zuxin and Zhu, Ming and Zhang, Jianguo and Awalgaonkar, Tulika and Wang, Shiyu and Liu, Zhiwei and Chen, Haolin and Hoang, Thai and Niebles, Juan Carlos and others},
  journal={arXiv preprint arXiv:2504.03601},
  year={2025}
}

@article{garimella2026saag,
  title={SAAG: Structured Agent Assessment and Grounding},
  author={Garimella, Ritvik and Khandelwal, Vedant and Kohli, Anvi and Sheth, Amit},
  journal={arXiv preprint arXiv:2607.18245},
  year={2026}
}

\clearpage

\appendix
\newpage

\section*{Appendices}
\addcontentsline{toc}{section}{Appendices}

\apptocentry{app:discussion}{\textbf{A\quad Discussions}}
\apptocentry{app:discussion_evals}{\quad A.1\quad Broader Implications}
\apptocentry{app:discussion_limitation}{\quad A.2\quad Limitations}
\apptocentry{app:discussion_future}{\quad A.3\quad Future Directions}

\apptocentry{app:reproducibility}{\textbf{B\quad Reproducibility Statement}}
\apptocentry{app:ethics}{\textbf{C\quad Ethics Statement}}

\apptocentry{app:dataset}{\textbf{D\quad RFCD: Tools as a Registry}}
\apptocentry{app:dataflow}{\quad D.1\quad Data Creation pipeline}
\apptocentry{app:benchflow}{\quad D.2\quad Benchmark Design}
\apptocentry{app:registry_examples}{\quad D.3\quad Registry Snapshots}

\apptocentry{app:fixes}{\textbf{E\quad TAAG: Assessing grounding of tool-augmentation}}
\apptocentry{app:taag_pipeline}{\quad E.1\quad Cascaded evaluation pipeline}
\apptocentry{app:taag_metrics}{\quad E.2\quad Evaluation Metrics}
\apptocentry{app:taag_corrections}{\quad E.3\quad Actionability probes: Structured Correction prompts}

\apptocentry{app:models}{\textbf{F\quad Models and Inference Settings}}
\apptocentry{app:model_selection}{\quad F.1\quad Model Selection Rationale}
\apptocentry{app:inference_settings}{\quad F.2\quad Inference Settings}

\apptocentry{app:res}{\textbf{G\quad Complete Results and Additional Views}}
\apptocentry{app:rs_cp}{\quad G.1\quad Registry Scale and Context Pressure}
\apptocentry{app:rs_itbf}{\quad G.2\quad Iteration Budget and Feedback Depth}
\apptocentry{app:rs_mss}{\quad G.3\quad Model and Source Sensitivity}
\apptocentry{app:rs_sdf}{\quad G.4\quad Structured Diagnostic Feedback}
\apptocentry{app:rs_avs}{\quad G.5\quad Alternative Visual Summaries}



\newpage

\section{Discussion}
\label{app:discussion}

\

\subsection{Broader Implications.}
\label{app:discussion_evals}

\textbf{RFCD.} shows that the registry itself is a first-class variable in tool-calling
evaluation. The main finding, that pass rate drops sharply and consistently as the registry
grows regardless of source or model family, implies that published leaderboard results
obtained on small, closed tool sets may substantially overstate the reliability of deployed
systems, where a model must identify the correct capability from hundreds or thousands of
candidates. The source-level breakdown further shows that the same end-to-end failure rate
can reflect different underlying bottlenecks: BFCL instances are predominantly
grounding-limited, Tool-Call-Data instances are schema-limited, and Glaive is mixed, so a
single aggregate metric conceals which intervention would be most effective.

\textbf{TAAG.} shows that tool-calling evaluation can support different conclusions
depending on the level at which correctness is measured. In the reported experiments,
registry conformance is nearly saturated under binary feedback, yet lexical argument
grounding remains substantially weaker compared to heavier semantic based parsers (avoided
due to speed and token consumption). These patterns are difficult to interpret from a
single aggregate outcome, but they become legible when registry validity, schema
satisfaction, and value grounding are measured separately. The framework separates
diagnostic value from correction performance: a model can improve structural coverage
while degrading value grounding, and this kind of distinction is precisely what an
evaluation framework should expose before claims about tool-use reliability are made.

\subsection{Limitations.} \label{app:discussion_limitation}

\textbf{RFCD.} First, distractors are sampled uniformly at random from source-specific
registry pools rather than selected for semantic confusability; the resulting scale
perturbation may understate the difficulty of production settings where many tools share
similar names, overlapping schemas, or related domains. Second, the benchmark is restricted
to single-action function calling; multi-step tool plans, dynamic registries, real external
execution, and agent-chaining workflows introduce additional failure modes that RFCD does
not cover. Third, the three source datasets are narrow in terms of true enterprise
use-cases, and the degree to which findings transfer to organically collected,
domain-specific, or production tool registries is not established.

\textbf{TAAG.} The argument grounding stage relies on lexical overlap rather than semantic
equivalence, which can incorrectly penalize valid paraphrases or values legitimately
inferred from context but not present on the query surface. The correction loop can
overcorrect valid predictions, particularly around tool names, where high NCS but
non-matching predictions trigger $\psi_1$ even when the emitted name is functionally
acceptable; future variants should consider confidence gating or more conservative
correction thresholds.

\subsection{Future Directions.} \label{app:discussion_future}

The most direct extension of RFCD is to replace uniform distractor sampling with
semantically guided selection (e.g., embedding-based nearest neighbours or adversarially
chosen confusable tools) so that the benchmark stresses name disambiguation and schema
differentiation rather than only registry size. Multi-action and multi-step variants would
allow RFCD's registry-scale framing to be applied to sequential tool use, where the model
must select and chain several functions while maintaining registry awareness across turns.
Integration with retrieval-based tool selection methods, where a retriever first narrows
the registry before the model generates a call, would let RFCD serve as a joint benchmark
for retrieval and generation, measuring whether retrieval errors propagate into the TAAG
failure cascade. On the evaluation side, replacing or supplementing TAAG's lexical
grounding stage with an embedding-based or model-judged semantic equivalence check would
reduce false penalization of valid paraphrases and inferred values. Studying training-time
objectives informed by TAAG diagnostics, for example targeted fine-tuning on schema-limited
or grounding-limited failure modes identified by the stage cascade, would test whether the
benchmark's diagnostic signals translate into actionable training improvements.

\clearpage

\section{Reproducibility Statement}
\label{app:reproducibility}

All experiments follow the RFCD + TAAG evaluation protocol with deterministic inference. We evaluate six locally served sub-4B models via Ollama: \texttt{llama3.2:3b}, \texttt{qwen2.5:3b}, \texttt{gemma4:e4b}, \texttt{qwen3:4b-thinking}, \texttt{phi4-mini:3.8b}, and \texttt{granite4:3b} by using a fixed prompt format and temperature $\tau = 0$. The correction loop allows up to $D_{\text{max}} = 15$ iterations. All thresholds and feedback templates are fixed and deterministic, including $\gamma_\text{pds} = 0.8$ and the Stage~3 hard-drift boundary (PartialRatio $< 0.6$), with correction templates ($\psi_1$, $\psi_2$, $\psi_3$) producing identical outputs for identical failure conditions. Experiments use the RFCD dataset (Hugging Face), constructed via JSON-schema normalization, rule-based filtering to single-call valid instances, and registry scaling through uniform distractor sampling; the dataset contains over 1.19M rows across multiple scales. All runs are executed on a single NVIDIA A100 GPU, and prompts, preprocessing, and evaluation code are provided via an anonymized repository.


\section{Ethics Statement}
\label{app:ethics}

This work presents an evaluation benchmark dataset and a diagnostic evaluation framework for tool-calling systems under controlled registry scaling. The evaluation is limited to single-action function calling with fixed registries and synthetic or benchmark-derived queries, and does not assess real tool execution, multi-step planning, or deployment settings. As a result, findings should not be interpreted as evidence of real-world readiness. The RFCD dataset reflects benchmark-style distributions with clean schemas and explicit intents, which may not capture noisy or ambiguous real-world usage and can introduce bias toward predictable tool-call formats. Additionally, TAAG grounding metrics rely on lexical overlap rather than full semantic equivalence, which may penalize valid but lexically divergent outputs.


\newpage

\section{RFCD: Tools as a Registry}
\label{app:dataset}
RFCD treats the available tools as a registry rather than single shot invocation. For each source dataset, function schemas were extracted and aggregated into a source-specific registry. For a target registry size $s$, we construct an instance by sampling $s-1$ distractor tools and inserting the gold tool when the example has a valid function call. This guarantees that the correct tool is available for positive examples while increasing the number of irrelevant tools that the model must ignore. For no-call examples, the registry contains only non-answer tools, so the model must learn to abstain rather than force an incorrect function call.

\subsection{Data Creation Pipeline}
\label{app:dataflow}

We began with a pool of existing function-calling datasets and selected three sources (Berkeley Function Calling Leaderboard
(BFCL), Glaive Function Calling v2, and Tool Plannings v0.2) based on their broad task coverage, explicit function schemas, and reliable ground-truth labels. Rows from each dataset were normalized into a unified format consisting of index, a set of candidate functions, user query, and the ground-truth target. Function definitions were represented as JSON schema blocks including typed parameters, descriptions, and required fields whenever available.

\textbf{Data preprocessing. }Involved filtering out rows with malformed JSON, unusable schemas, ambiguous ground truth, or unsupported multi-call outputs. For datasets that included no-call behavior, we preserved null targets as explicit negative examples. This enables RFCD to evaluate not only correct tool selection but also appropriate abstention when no candidate function applies.

\textbf{Candidate registry construction. } We construct scale-based splits using registry sizes
\[
s \in \{5, 10, 15, 20, 30, 40, 50, 100, 200, 300, 400, 500, 1000, 2000, 3000\}.
\]
The maximum available scale depends on the source registry size: Glaive is released up to scale 3000, while BFCL and Tool Plannings are released up to scale 500 in the current dataset.

Let the source datasets be denoted by
\[
\mathcal{D} = \{\mathcal{D}^{\texttt{bfcl}}, \mathcal{D}^{\texttt{glaive}}, \mathcal{D}^{\texttt{tool\_plannings}}\}.
\]
Each raw example is represented as
\[
x_i = (p_i, q_i, y_i),
\]
where $p_i$ is the system prompt or function-schema context, $q_i$ is the user query, and $y_i$ is the ground-truth function call. For datasets that contain abstention examples, $y_i = \varnothing$ denotes that no tool should be called.

We extract the available tools from the system prompt using a parser
\[
\mathcal{E}(p_i) = \{t_{i,1}, t_{i,2}, \ldots, t_{i,k_i}\},
\]
where each tool $t$ is a JSON-schema function definition:
\[
t = (\text{name}, \text{description}, \text{parameters}, \text{required}).
\]
A tool is retained only if it satisfies schema validity:
\[
\operatorname{valid}(t)=1
\iff
\text{name}(t) \neq \varnothing
\land
\text{parameters}(t) \text{ is valid JSON schema}.
\]

For each source dataset $\mathcal{D}^{m}$, we aggregate all valid extracted tools into a source-specific registry pool:
\[
\mathcal{A}^{m}
=
\bigcup_{x_i \in \mathcal{D}^{m}}
\{t \in \mathcal{E}(p_i) : \operatorname{valid}(t)=1\}.
\]
After canonicalization and duplicate removal, this candidates registry pool is stored as the corresponding \texttt{agents\_\{n\}.jsonl} registry file. The benchmark examples are then created by sampling from this pool. Figures \ref{fig:cand-reg} of sub-section \ref{app:registry_examples} showcase the candidate registries with scale of $s=[5,10,15]$ generated for sample index $108$.

\textbf{Benchmark instance construction. } Let $g_i$ denote the gold tool associated with example $x_i$ when $y_i \neq \varnothing$. For a registry scale $s$, we construct the candidate registry for a positive example as
\[
\mathcal{R}_{i,s}
=
\{g_i\}
\cup
\operatorname{Sample}_{s-1}
\left(
\mathcal{A}^{m} \setminus \{g_i\}
\right),
\quad
|\mathcal{R}_{i,s}| = s,
\quad
g_i \in \mathcal{R}_{i,s}.
\]
For no-call examples, where $y_i = \varnothing$, the registry is sampled without inserting a gold tool:
\[
\mathcal{R}_{i,s}
=
\operatorname{Sample}_{s}
\left(
\mathcal{A}^{m}
\right),
\]
with expected output $\varnothing$. The resulting RFCD instance at scale $s$ is
\[
z_{i,s} = (q_i, \mathcal{R}_{i,s}, y_i),
\]
where the model must either select the correct function and arguments or abstain if no valid function applies.

We construct RFCD over the scale set
\[
\mathcal{S}
=
\{5,10,15,20,30,40,50,100,200,300,400,500,1000,2000,3000\},
\]
and define the dataset for source $m$ at scale $s$ as
\[
\mathcal{B}^{m}_{s}
=
\{z_{i,s} : x_i \in \mathcal{D}^{m},\ |\mathcal{R}_{i,s}| = s\}.
\]

As part of \textbf{Quality Control}, we apply rule-based filtering to ensure dataset consistency. An example is retained if and only if
\[
\operatorname{keep}(x_i)=1
\iff
\operatorname{valid}(\mathcal{E}(p_i))
\land
\operatorname{single}(y_i)
\land
\operatorname{unambiguous}(y_i),
\]
where $\operatorname{single}(y_i)$ enforces at most one function call and $\operatorname{unambiguous}(y_i)$ ensures a unique, reliably matchable ground-truth tool.

For positive examples, we enforce gold-tool containment:
\[
y_i \neq \varnothing \implies g_i \in \mathcal{R}_{i,s}.
\]
For all examples, we require fixed registry size and tool uniqueness:
\[
|\mathcal{R}_{i,s}| = s,
\qquad
|\{\text{name}(t): t \in \mathcal{R}_{i,s}\}| = s.
\]
If these constraints are violated, the distractor set is resampled.

This construction isolates the registry-selection problem: as $s$ increases, the query and gold target remain fixed while the number of irrelevant tools grows, so performance degradation reflects sensitivity to registry size rather than changes in task semantics.

\subsection{Benchmark Design and Quality Control}
\label{app:benchflow}

In the released huggingface benchmark, each $\mathcal{B}^{m}_{s}$ is instantiated as a scale-specific split, e.g., $\texttt{data/}m\texttt{/scale\_}s$, where $m$ indexes the source configuration and $s$ denotes the registry size. Each instance is evaluated in a fixed format consisting of a user query, a candidate function registry, and a target output specifying either a valid function call or abstention.

\paragraph{Benchmark Design.}
The benchmark is designed to isolate the effect of registry scale on function-calling performance. For a fixed example $x_i$, the query $q_i$ and ground-truth target $y_i$ remain invariant across all scales $s \in \mathcal{S}$. Only the size of the candidate registry $\mathcal{R}_{i,s}$ changes. This controlled scaling ensures that performance differences across $s$ are not attributable to distributional shifts in queries or targets, but instead reflect the model’s ability to operate over increasingly large candidate sets.

This setup instantiates a \emph{registry-as-database} evaluation regime: rather than exposing only the relevant function, the model must identify the correct tool from among $s$ candidates. As $s$ grows, the search space expands combinatorially, increasing the likelihood of name collisions, schema similarity, and argument confusion. Because distractors are not explicitly adversarially curated, the benchmark should be interpreted as a controlled scale perturbation rather than a confusability-optimized stress test.

The benchmark evaluates four core capabilities:
\begin{itemize}
    \item (i)~tool-use decision making (call vs.\ abstain),
    \item (ii)~function selection from the registry,
    \item (iii)~argument generation consistent with the schema, and
    \item (iv)~strict adherence to the required function-call output format.
\end{itemize}

\paragraph{Cross-Scale Consistency.}
A defining property of the benchmark is cross-scale consistency. For each source configuration $m$ and example index $i$, the family $\{z_{i,s}\}_{s \in \mathcal{S}}$ forms a matched set where only the registry size varies. This enables paired evaluation across scales, allowing direct measurement of performance degradation as a function of $s$ without confounding factors from query variation. Consequently, differences in accuracy, selection errors, or argument validity can be attributed to increased registry complexity.

\paragraph{Evaluation Scope.}
The benchmark is explicitly scoped to single-step function-calling. Each instance requires at most one function call or abstention, and evaluation focuses on correctness of tool selection, argument structure, and output formatting. The benchmark does not evaluate multi-step planning, tool chaining, or interactive reasoning, and therefore isolates the atomic function-calling capability under scale.

\paragraph{Quality Control.}
Quality control is enforced at the level of benchmark validity and evaluation integrity rather than data construction. All examples included in $\mathcal{B}^{m}_{s}$ satisfy structural invariants required for reliable evaluation: (i)~each instance has a well-defined target output, (ii)~function schemas are consistently formatted and machine-readable, and (iii)~the registry presented to the model is internally consistent.

For positive examples, correctness of evaluation requires that the gold function is present in the candidate registry; for abstention cases, no such constraint is imposed. Additionally, registry integrity is enforced by ensuring fixed size $|\mathcal{R}_{i,s}| = s$ and uniqueness of function identifiers within each registry. These guarantees ensure that evaluation failures arise from model behavior rather than inconsistencies in the benchmark.

\paragraph{Benchmark Packaging and Reproducibility.}
The benchmark is released in a scale-stratified format across multiple source configurations (e.g., \texttt{bfcl}, \texttt{glaive}, \texttt{tool\_plannings}), with identical split names across scales to support consistent evaluation pipelines. Data is stored as Zstandard-compressed Parquet shards to enable efficient loading and large-scale experimentation. 

To facilitate rapid inspection and debugging, a lightweight \texttt{sample} configuration is provided, containing a small, fixed subset of examples for each source and scale. This separation between full benchmark and sample subsets allows reproducible large-scale evaluation while supporting lightweight development workflows.

\paragraph{Intended Use and Limitations.}
The benchmark is intended for controlled evaluation of function-calling under varying registry sizes. It is not designed to reflect full production environments, multi-turn interactions, or real-world API behavior. Instead, it provides a standardized testbed for analyzing how model performance scales with the size of the candidate function space, particularly in terms of selection accuracy, argument correctness, and format compliance.

Unlike function-calling settings that expose only the relevant tool, our
construction forces the model to perform registry selection: the correct
function must be identified from among $S{-}1$ randomly sampled distractors.
Because distractors are not curated for semantic confusability, this setup is
a controlled scale perturbation rather than an adversarial registry benchmark.
As $S$ grows within the tested range, the opportunity for name confusability
and argument drift increases. This probes the failure modes captured by our
TAAG metrics.

Figures (~\ref{fig:full-2433-5}, ~\ref{fig:full-2433-10}, ~\ref{fig:full-2433-15})  of sub-section ~\ref{app:registry_examples} shows a representative system prompt at
$S{=}5$. At $S{=}10$ and $S{=}15$, five and ten further distractor
blocks are appended; the query and ground truth are unchanged. The only place ground truth is used is during the extrinsic metric calculation.

\clearpage

\subsection{Registry Snapshots}
\label{app:registry_examples}

\begin{figure}[h]
\centering

\begin{tcolorbox}[
    title=Candidate registry metadata at $S{=}5$ for query index 2433,
    colback=teal!5!white,
    colframe=teal!30!white,
    colbacktitle=teal!20!white,
    coltitle=black,
    fonttitle=\bfseries
]
\raggedright
\scriptsize
\texttt{\{} \\
\hspace*{1em}\textbf{"index"}\texttt{: 2433,} \\
\hspace*{1em}\textbf{"sampled\_agents"}\texttt{: ["calculate\_loan\_interest", "check\_availability",} \\
\hspace*{4em}\texttt{"create\_note", "generate\_random\_quote", "generate\_random\_password"],} \\
\hspace*{1em}\textbf{"sampled\_registry\_indices"}\texttt{: [82977, 400, 42888, 46683, 2886],} \\
\hspace*{1em}\textbf{"relevant\_agents"}\texttt{: ["generate\_random\_password"],} \\
\hspace*{1em}\textbf{"relevant\_registry\_indices"}\texttt{: [2886],} \\
\hspace*{1em}\textbf{"relevance\_source"}\texttt{: "function\_call",} \\
\hspace*{1em}\textbf{"injected\_relevant\_agent"}\texttt{: true,} \\
\hspace*{1em}\textbf{"contains\_relevant\_agent"}\texttt{: true} \\
\texttt{\}}
\end{tcolorbox}

\vspace{0.5em}

\begin{tcolorbox}[
    title=Candidate registry metadata at $S{=}10$ for query index 2433,
    colback=red!5!white,
    colframe=red!30!white,
    colbacktitle=red!20!white,
    coltitle=black,
    fonttitle=\bfseries
]
\raggedright
\scriptsize
\texttt{\{} \\
\hspace*{1em}\textbf{"index"}\texttt{: 2433,} \\
\hspace*{1em}\textbf{"sampled\_agents"}\texttt{: ["search\_books", "check\_email\_availability",} \\
\hspace*{4em}\texttt{"generate\_random\_password", "calculate\_area", "create\_calendar\_event",} \\
\hspace*{4em}\texttt{"play\_music", "generate\_password", "get\_translation",} \\
\hspace*{4em}\texttt{"generate\_qr\_code", "validate\_credit\_card"],} \\
\hspace*{1em}\textbf{"sampled\_registry\_indices"}\texttt{: [27568, 74124, 2886, 37892, 33957,} \\
\hspace*{4em}\texttt{5232, 24385, 73488, 17481, 42226],} \\
\hspace*{1em}\textbf{"relevant\_agents"}\texttt{: ["generate\_random\_password"],} \\
\hspace*{1em}\textbf{"relevant\_registry\_indices"}\texttt{: [2886],} \\
\hspace*{1em}\textbf{"relevance\_source"}\texttt{: "function\_call",} \\
\hspace*{1em}\textbf{"injected\_relevant\_agent"}\texttt{: true,} \\
\hspace*{1em}\textbf{"contains\_relevant\_agent"}\texttt{: true} \\
\texttt{\}}
\end{tcolorbox}

\vspace{0.5em}

\begin{tcolorbox}[
    title=Candidate registry metadata at $S{=}15$ for query index 2433,
    colback=orange!5!white,
    colframe=orange!30!white,
    colbacktitle=orange!20!white,
    coltitle=black,
    fonttitle=\bfseries
]
\raggedright
\scriptsize
\texttt{\{} \\
\hspace*{1em}\textbf{"index"}\texttt{: 2433,} \\
\hspace*{1em}\textbf{"sampled\_agents"}\texttt{: ["search\_movie\_theater", "calculate\_area",} \\
\hspace*{4em}\texttt{"send\_email", "generate\_random\_username", "generate\_qr\_code",} \\
\hspace*{4em}\texttt{"convert\_currency", "create\_invoice", "search\_restaurants",} \\
\hspace*{4em}\texttt{"generate\_random\_password", "create\_user\_profile", "calculate\_bmi",} \\
\hspace*{4em}\texttt{"schedule\_meeting", "create\_user", "get\_definition", "calculate\_tip"],} \\
\hspace*{1em}\textbf{"sampled\_registry\_indices"}\texttt{: [620, 7406, 61440, 73441, 73467,} \\
\hspace*{4em}\texttt{15859, 82906, 19271, 2886, 31752, 22620, 43821, 12426, 20555, 11343],} \\
\hspace*{1em}\textbf{"relevant\_agents"}\texttt{: ["generate\_random\_password"],} \\
\hspace*{1em}\textbf{"relevant\_registry\_indices"}\texttt{: [2886],} \\
\hspace*{1em}\textbf{"relevance\_source"}\texttt{: "function\_call",} \\
\hspace*{1em}\textbf{"injected\_relevant\_agent"}\texttt{: true,} \\
\hspace*{1em}\textbf{"contains\_relevant\_agent"}\texttt{: true} \\
\texttt{\}}
\end{tcolorbox}

\caption{Candidate registry metadata for query index 2433 across scales $s=[5,10,15]$.
  The relevant agent \texttt{generate\_random\_password} (registry index 2886) is
  injected into each sampled set. At $S{=}10$, a near-identical distractor
  \texttt{generate\_password} also appears, increasing ambiguity. The registry
  grows from 5 candidates at $S{=}5$ to $S{=}15$, with the query and
  ground truth unchanged across all scales.}
\label{fig:cand-reg}
\end{figure}

\begin{figure}[h]
\begin{tcolorbox}[
    title=Full registry entry at $S{=}5$ for query index 2433,
    colback=teal!5!white,
    colframe=teal!30!white,
    colbacktitle=teal!20!white,
    coltitle=black,
    fonttitle=\bfseries
]
\raggedright
\scriptsize
\texttt{\{} \\
\hspace*{1em}\textbf{"id"}\texttt{: 2433,} \\
\hspace*{1em}\textbf{"function"}\texttt{: [} \\
\hspace*{2em}\texttt{\{} \\
\hspace*{3em}\textbf{"name"}\texttt{: "calculate\_loan\_interest",} \\
\hspace*{3em}\textbf{"description"}\texttt{: "Calculate the interest on a loan",} \\
\hspace*{3em}\textbf{"parameters"}\texttt{: \{"type": "object", "properties": \{} \\
\hspace*{4em}\textbf{"loan\_amount"}\texttt{: \{"type": "number", "description": "The total loan amount"\},} \\
\hspace*{4em}\textbf{"interest\_rate"}\texttt{: \{"type": "number", "description": "The annual interest rate"\},} \\
\hspace*{4em}\textbf{"loan\_term"}\texttt{: \{"type": "integer", "description": "The term of the loan in years"\}\},} \\
\hspace*{3em}\texttt{"required": ["loan\_amount", "interest\_rate", "loan\_term"]\}\}} \\
\hspace*{2em}\texttt{\},} \\
\hspace*{2em}\texttt{\{} \\
\hspace*{3em}\textbf{"name"}\texttt{: "check\_availability",} \\
\hspace*{3em}\textbf{"description"}\texttt{: "Check the availability of a product or service",} \\
\hspace*{3em}\textbf{"parameters"}\texttt{: \{"type": "object", "properties": \{} \\
\hspace*{4em}\textbf{"product"}\texttt{: \{"type": "string", "description": "The name or ID of the product"\},} \\
\hspace*{4em}\textbf{"location"}\texttt{: \{"type": "string", "description": "The location to check availability"\}\},} \\
\hspace*{3em}\texttt{"required": ["product", "location"]\}\}} \\
\hspace*{2em}\texttt{\},} \\
\hspace*{2em}\texttt{\{} \\
\hspace*{3em}\textbf{"name"}\texttt{: "create\_note",} \\
\hspace*{3em}\textbf{"description"}\texttt{: "Create a new note",} \\
\hspace*{3em}\textbf{"parameters"}\texttt{: \{"type": "object", "properties": \{} \\
\hspace*{4em}\textbf{"title"}\texttt{: \{"type": "string", "description": "The title of the note"\},} \\
\hspace*{4em}\textbf{"content"}\texttt{: \{"type": "string", "description": "The content of the note"\}\},} \\
\hspace*{3em}\texttt{"required": ["title", "content"]\}\}} \\
\hspace*{2em}\texttt{\},} \\
\hspace*{2em}\texttt{\{} \\
\hspace*{3em}\textbf{"name"}\texttt{: "generate\_random\_quote",} \\
\hspace*{3em}\textbf{"description"}\texttt{: "Generate a random inspirational quote",} \\
\hspace*{3em}\textbf{"parameters"}\texttt{: \{\}} \\
\hspace*{2em}\texttt{\},} \\
\hspace*{2em}\texttt{\{} \\
\hspace*{3em}\textbf{"name"}\texttt{: "generate\_random\_password",} \\
\hspace*{3em}\textbf{"description"}\texttt{: "Generate a random password with given length",} \\
\hspace*{3em}\textbf{"parameters"}\texttt{: \{"type": "object", "properties": \{} \\
\hspace*{4em}\textbf{"length"}\texttt{: \{"type": "integer", "description": "The length of the password"\}\},} \\
\hspace*{3em}\texttt{"required": ["length"]\}\}} \\
\hspace*{2em}\texttt{\}],} \\
\hspace*{1em}\textbf{"question"}\texttt{: "Hi, I need a new password. Can you generate a random one for me?",} \\
\hspace*{1em}\textbf{"ground\_truth"}\texttt{: \{"name": "generate\_random\_password",} \\
\hspace*{4em}\texttt{"arguments": \{"length": 12\}\}} \\
\texttt{\}}
\end{tcolorbox}
\caption{Full registry entry at $S{=}5$ for query index 2433. Five candidate functions
  are presented, of which \texttt{generate\_random\_password} at position five is the
  sole ground-truth function. The remaining four are distractors with no overlap
  in functionality.}
\label{fig:full-2433-5}
\end{figure}

\begin{figure}[h]
\begin{tcolorbox}[
    title=Full registry entry at $S{=}10$ for query index 2433,
    colback=red!5!white,
    colframe=red!30!white,
    colbacktitle=red!20!white,
    coltitle=black,
    fonttitle=\bfseries
]
\raggedright
\scriptsize
\texttt{\{} \\
\hspace*{1em}\textbf{"id"}\texttt{: 2433,} \\
\hspace*{1em}\textbf{"function"}\texttt{: [} \\
\hspace*{2em}\texttt{\{} \\
\hspace*{3em}\textbf{"name"}\texttt{: "search\_books",} \\
\hspace*{3em}\textbf{"description"}\texttt{: "Search for books based on title or author",} \\
\hspace*{3em}\textbf{"parameters"}\texttt{: \{"type": "object", "properties": \{} \\
\hspace*{4em}\textbf{"query"}\texttt{: \{"type": "string", "description": "The search query for books"\}\},} \\
\hspace*{3em}\texttt{"required": ["query"]\}\}} \\
\hspace*{2em}\texttt{\},} \\
\hspace*{2em}\texttt{\{} \\
\hspace*{3em}\textbf{"name"}\texttt{: "check\_email\_availability",} \\
\hspace*{3em}\textbf{"description"}\texttt{: "Check the availability of an email address",} \\
\hspace*{3em}\textbf{"parameters"}\texttt{: \{"type": "object", "properties": \{} \\
\hspace*{4em}\textbf{"email"}\texttt{: \{"type": "string", "description": "The email address to check"\}\},} \\
\hspace*{3em}\texttt{"required": ["email"]\}\}} \\
\hspace*{2em}\texttt{\},} \\
\hspace*{2em}\texttt{\{} \\
\hspace*{3em}\textbf{"name"}\texttt{: "generate\_random\_password",} \\
\hspace*{3em}\textbf{"description"}\texttt{: "Generate a random password with given length",} \\
\hspace*{3em}\textbf{"parameters"}\texttt{: \{"type": "object", "properties": \{} \\
\hspace*{4em}\textbf{"length"}\texttt{: \{"type": "integer", "description": "The length of the password"\}\},} \\
\hspace*{3em}\texttt{"required": ["length"]\}\}} \\
\hspace*{2em}\texttt{\},} \\
\hspace*{2em}\texttt{\{} \\
\hspace*{3em}\textbf{"name"}\texttt{: "calculate\_area",} \\
\hspace*{3em}\textbf{"description"}\texttt{: "Calculate the area of a given shape",} \\
\hspace*{3em}\textbf{"parameters"}\texttt{: \{"type": "object", "properties": \{} \\
\hspace*{4em}\textbf{"shape"}\texttt{: \{"type": "string", "description": "The shape for which the area will be calculated"\},} \\
\hspace*{4em}\textbf{"dimensions"}\texttt{: \{"type": "object", "properties": \{} \\
\hspace*{5em}\textbf{"length"}\texttt{: \{"type": "number", "description": "The length of the shape (if applicable)"\},} \\
\hspace*{5em}\textbf{"width"}\texttt{: \{"type": "number", "description": "The width of the shape (if applicable)"\},} \\
\hspace*{5em}\textbf{"radius"}\texttt{: \{"type": "number", "description": "The radius of the shape (if applicable)"\}\}\}\},} \\
\hspace*{3em}\texttt{"required": ["shape"]\}\}} \\
\hspace*{2em}\texttt{\},} \\
\hspace*{2em}\texttt{\{} \\
\hspace*{3em}\textbf{"name"}\texttt{: "create\_calendar\_event",} \\
\hspace*{3em}\textbf{"description"}\texttt{: "Create a new calendar event",} \\
\hspace*{3em}\textbf{"parameters"}\texttt{: \{"type": "object", "properties": \{} \\
\hspace*{4em}\textbf{"title"}\texttt{: \{"type": "string", "description": "The title of the event"\},} \\
\hspace*{4em}\textbf{"start\_time"}\texttt{: \{"type": "string", "description": "The start time of the event"\},} \\
\hspace*{4em}\textbf{"end\_time"}\texttt{: \{"type": "string", "description": "The end time of the event"\}\},} \\
\hspace*{3em}\texttt{"required": ["title", "start\_time", "end\_time"]\}\}} \\
\hspace*{2em}\texttt{\},} \\
\hspace*{2em}\texttt{\{} \\
\hspace*{3em}\textbf{"name"}\texttt{: "play\_music",} \\
\hspace*{3em}\textbf{"description"}\texttt{: "Play music from a specified playlist or genre",} \\
\hspace*{3em}\textbf{"parameters"}\texttt{: \{"type": "object", "properties": \{} \\
\hspace*{4em}\textbf{"playlist"}\texttt{: \{"type": "string", "description": "The name of the playlist"\},} \\
\hspace*{4em}\textbf{"genre"}\texttt{: \{"type": "string", "description": "The genre of music to play"\}\}\}} \\
\hspace*{2em}\texttt{\},} \\
\hspace*{2em}\texttt{...}
\end{tcolorbox}
\caption{Full registry entry at $S{=}10$ for query index 2433, continued in
  Figure~\ref{fig:full-2433-10b}. The relevant function \texttt{generate\_random\_password}
  appears at position three. A near-identical distractor \texttt{generate\_password}
  appears later with empty parameters, increasing ambiguity.}
\label{fig:full-2433-10}
\end{figure}

\begin{figure}[h]
\addtocounter{figure}{-1}
\begin{tcolorbox}[
    title=Full registry entry at $S{=}10$ for query index 2433 \textnormal{(cont.)},
    colback=red!5!white,
    colframe=red!30!white,
    colbacktitle=red!20!white,
    coltitle=black,
    fonttitle=\bfseries
]
\raggedright
\scriptsize
\hspace*{2em}\texttt{...} \\
\hspace*{2em}\texttt{\{} \\
\hspace*{3em}\textbf{"name"}\texttt{: "generate\_password",} \\
\hspace*{3em}\textbf{"description"}\texttt{: "Generate a random password",} \\
\hspace*{3em}\textbf{"parameters"}\texttt{: \{\}} \\
\hspace*{2em}\texttt{\},} \\
\hspace*{2em}\texttt{\{} \\
\hspace*{3em}\textbf{"name"}\texttt{: "get\_translation",} \\
\hspace*{3em}\textbf{"description"}\texttt{: "Get the translation of a word or phrase",} \\
\hspace*{3em}\textbf{"parameters"}\texttt{: \{"type": "object", "properties": \{} \\
\hspace*{4em}\textbf{"text"}\texttt{: \{"type": "string", "description": "The word or phrase to translate"\},} \\
\hspace*{4em}\textbf{"source\_language"}\texttt{: \{"type": "string", "description": "The source language of the text"\},} \\
\hspace*{4em}\textbf{"target\_language"}\texttt{: \{"type": "string", "description": "The target language for the translation"\}\},} \\
\hspace*{3em}\texttt{"required": ["text", "source\_language", "target\_language"]\}\}} \\
\hspace*{2em}\texttt{\},} \\
\hspace*{2em}\texttt{\{} \\
\hspace*{3em}\textbf{"name"}\texttt{: "generate\_qr\_code",} \\
\hspace*{3em}\textbf{"description"}\texttt{: "Generate a QR code for a given text",} \\
\hspace*{3em}\textbf{"parameters"}\texttt{: \{"type": "object", "properties": \{} \\
\hspace*{4em}\textbf{"text"}\texttt{: \{"type": "string", "description": "The text to be encoded in the QR code"\}\},} \\
\hspace*{3em}\texttt{"required": ["text"]\}\}} \\
\hspace*{2em}\texttt{\},} \\
\hspace*{2em}\texttt{\{} \\
\hspace*{3em}\textbf{"name"}\texttt{: "validate\_credit\_card",} \\
\hspace*{3em}\textbf{"description"}\texttt{: "Validate a credit card number",} \\
\hspace*{3em}\textbf{"parameters"}\texttt{: \{"type": "object", "properties": \{} \\
\hspace*{4em}\textbf{"credit\_card\_number"}\texttt{: \{"type": "string", "description": "The credit card number to be validated"\}\},} \\
\hspace*{3em}\texttt{"required": ["credit\_card\_number"]\}\}} \\
\hspace*{2em}\texttt{\}],} \\
\hspace*{1em}\textbf{"question"}\texttt{: "Hi, I need a new password. Can you generate a random one for me?",} \\
\hspace*{1em}\textbf{"ground\_truth"}\texttt{: \{"name": "generate\_random\_password",} \\
\hspace*{4em}\texttt{"arguments": \{"length": 12\}\}} \\
\texttt{\}}
\end{tcolorbox}
\caption{Figure~\thefigure{} continued.}
\label{fig:full-2433-10b}
\end{figure}

\begin{figure}[h]
\begin{tcolorbox}[
    title=Full registry entry at $S{=}15$ for query index 2433,
    colback=orange!5!white,
    colframe=orange!30!white,
    colbacktitle=orange!20!white,
    coltitle=black,
    fonttitle=\bfseries
]
\raggedright
\scriptsize
\texttt{\{} \\
\hspace*{1em}\textbf{"id"}\texttt{: 2433,} \\
\hspace*{1em}\textbf{"function"}\texttt{: [} \\
\hspace*{2em}\texttt{\{} \\
\hspace*{3em}\textbf{"name"}\texttt{: "search\_movie\_theater",} \\
\hspace*{3em}\textbf{"description"}\texttt{: "Search for movie theaters in a specific location",} \\
\hspace*{3em}\textbf{"parameters"}\texttt{: \{"type": "object", "properties": \{} \\
\hspace*{4em}\textbf{"location"}\texttt{: \{"type": "string", "description": "The location to search for movie theaters"\},} \\
\hspace*{4em}\textbf{"movie\_genre"}\texttt{: \{"type": "string", "description": "The desired movie genre"\},} \\
\hspace*{4em}\textbf{"date"}\texttt{: \{"type": "string", "format": "date", "description": "The desired date for movie showtimes"\}\},} \\
\hspace*{3em}\texttt{"required": ["location"]\}\}} \\
\hspace*{2em}\texttt{\},} \\
\hspace*{2em}\texttt{\{} \\
\hspace*{3em}\textbf{"name"}\texttt{: "calculate\_area",} \\
\hspace*{3em}\textbf{"description"}\texttt{: "Calculate area of a square, rectangle, or triangle",} \\
\hspace*{3em}\textbf{"parameters"}\texttt{: \{"type": "object", "properties": \{} \\
\hspace*{4em}\textbf{"shape"}\texttt{: \{"type": "string", "enum": ["square", "rectangle", "triangle"],} \\
\hspace*{5em}\texttt{"description": "The shape to calculate area for"\},} \\
\hspace*{4em}\textbf{"dimensions"}\texttt{: \{"type": "object", "properties": \{} \\
\hspace*{5em}\textbf{"side"}\texttt{: \{"type": "number", "description": "The length of side (applicable to square)"\},} \\
\hspace*{5em}\textbf{"length"}\texttt{: \{"type": "number", "description": "The length of rectangle (applicable to rectangle)"\},} \\
\hspace*{5em}\textbf{"width"}\texttt{: \{"type": "number", "description": "The width of rectangle (applicable to rectangle)"\},} \\
\hspace*{5em}\textbf{"base"}\texttt{: \{"type": "number", "description": "The base of triangle (applicable to triangle)"\},} \\
\hspace*{5em}\textbf{"height"}\texttt{: \{"type": "number", "description": "The height of triangle (applicable to triangle)"\}\},} \\
\hspace*{4em}\texttt{"required": ["side", "length", "width", "base", "height"]\}\},} \\
\hspace*{3em}\texttt{"required": ["shape", "dimensions"]\}\}} \\
\hspace*{2em}\texttt{\},} \\
\hspace*{2em}\texttt{\{} \\
\hspace*{3em}\textbf{"name"}\texttt{: "send\_email",} \\
\hspace*{3em}\textbf{"description"}\texttt{: "Send an email",} \\
\hspace*{3em}\textbf{"parameters"}\texttt{: \{"type": "object", "properties": \{} \\
\hspace*{4em}\textbf{"to"}\texttt{: \{"type": "string", "description": "The recipient's email address"\},} \\
\hspace*{4em}\textbf{"subject"}\texttt{: \{"type": "string", "description": "The subject of the email"\},} \\
\hspace*{4em}\textbf{"body"}\texttt{: \{"type": "string", "description": "The body of the email"\}\},} \\
\hspace*{3em}\texttt{"required": ["to", "subject", "body"]\}\}} \\
\hspace*{2em}\texttt{\},} \\
\hspace*{2em}\texttt{\{} \\
\hspace*{3em}\textbf{"name"}\texttt{: "generate\_random\_username",} \\
\hspace*{3em}\textbf{"description"}\texttt{: "Generate a random username",}
\hspace*{1em}\textbf{"parameters"}\texttt{: \{\}} \\
\hspace*{2em}\texttt{\},} \\
\hspace*{2em}\texttt{\{} \\
\hspace*{3em}\textbf{"name"}\texttt{: "generate\_qr\_code",} \\
\hspace*{3em}\textbf{"description"}\texttt{: "Generate a QR code from a given text",} \\
\hspace*{3em}\textbf{"parameters"}\texttt{: \{"type": "object", "properties": \{} \\
\hspace*{4em}\textbf{"text"}\texttt{: \{"type": "string", "description": "The text to be encoded in the QR code"\}\},} \\
\hspace*{3em}\texttt{"required": ["text"]\}\}} \\
\hspace*{2em}\texttt{\},} \\
\hspace*{2em}\texttt{\{} \\
\hspace*{3em}\textbf{"name"}\texttt{: "convert\_currency",} \\
\hspace*{3em}\textbf{"description"}\texttt{: "Convert currency from one type to another",} \\
\hspace*{3em}\textbf{"parameters"}\texttt{: \{"type": "object", "properties": \{} \\
\hspace*{4em}\textbf{"from\_currency"}\texttt{: \{"type": "string", "description": "The currency to convert from"\},} \\
\hspace*{4em}\textbf{"to\_currency"}\texttt{: \{"type": "string", "description": "The currency to convert to"\},} \\
\hspace*{4em}\textbf{"amount"}\texttt{: \{"type": "number", "description": "The amount to convert"\}\},} \\
\hspace*{3em}\texttt{"required": ["from\_currency", "to\_currency", "amount"]\}\}} \\
\hspace*{2em}\texttt{\},} \\
\hspace*{2em}\texttt{\{} \\
\hspace*{3em}\textbf{"name"}\texttt{: "create\_invoice",} \\
\hspace*{3em}\textbf{"description"}\texttt{: "Create a new invoice",} \\
\hspace*{3em}\textbf{"parameters"}\texttt{: \{"type": "object", "properties": \{} \\
\hspace*{4em}\textbf{"client\_name"}\texttt{: \{"type": "string", "description": "The name of the client"\},} \\
\hspace*{4em}\textbf{"amount"}\texttt{: \{"type": "number", "description": "The amount of the invoice"\},} \\
\hspace*{4em}\textbf{"due\_date"}\texttt{: \{"type": "string", "description": "The due date of the invoice"\}\},} \\
\hspace*{3em}\texttt{"required": ["client\_name", "amount"]\}\}} \\
\hspace*{2em}\texttt{\},} \\
\hspace*{2em}\texttt{...}
\end{tcolorbox}
\caption{Full registry entry at $S{=}15$ for query index 2433, continued in
  Figure~\ref{fig:full-2433-15b}. Functions 1--7 shown. The relevant function
  \texttt{generate\_random\_password} appears at position nine.}
\label{fig:full-2433-15}
\end{figure}

\begin{figure}[h]
\addtocounter{figure}{-1}
\begin{tcolorbox}[
    title=Full registry entry at $S{=}15$ for query index 2433 \textnormal{(cont.)},
    colback=orange!5!white,
    colframe=orange!30!white,
    colbacktitle=orange!20!white,
    coltitle=black,
    fonttitle=\bfseries
]
\raggedright
\scriptsize
\hspace*{2em}\texttt{...} \\
\hspace*{2em}\texttt{\{} \\
\hspace*{3em}\textbf{"name"}\texttt{: "search\_restaurants",} \\
\hspace*{3em}\textbf{"description"}\texttt{: "Search for restaurants based on location and cuisine",} \\
\hspace*{3em}\textbf{"parameters"}\texttt{: \{"type": "object", "properties": \{} \\
\hspace*{4em}\textbf{"location"}\texttt{: \{"type": "string", "description": "The location to search for restaurants"\},} \\
\hspace*{4em}\textbf{"cuisine"}\texttt{: \{"type": "string", "description": "The type of cuisine to search for"\}\},} \\
\hspace*{3em}\texttt{"required": ["location", "cuisine"]\}\}} \\
\hspace*{2em}\texttt{\},} \\

\hspace*{2em}\texttt{\{} \\
\hspace*{3em}\textbf{"name"}\texttt{: "generate\_random\_password",} \\
\hspace*{3em}\textbf{"description"}\texttt{: "Generate a random password with given length",} \\
\hspace*{3em}\textbf{"parameters"}\texttt{: \{"type": "object", "properties": \{} \\
\hspace*{4em}\textbf{"length"}\texttt{: \{"type": "integer", "description": "The length of the password"\}\},} \\
\hspace*{3em}\texttt{"required": ["length"]\}\}} \\
\hspace*{2em}\texttt{\},} \\
\hspace*{2em}\texttt{\{} \\
\hspace*{3em}\textbf{"name"}\texttt{: "create\_user\_profile",} \\
\hspace*{3em}\textbf{"description"}\texttt{: "Create a new user profile",} \\
\hspace*{3em}\textbf{"parameters"}\texttt{: \{"type": "object", "properties": \{} \\
\hspace*{4em}\textbf{"name"}\texttt{: \{"type": "string", "description": "The user's name"\},} \\
\hspace*{4em}\textbf{"age"}\texttt{: \{"type": "integer", "description": "The user's age"\},} \\
\hspace*{4em}\textbf{"email"}\texttt{: \{"type": "string", "format": "email", "description": "The user's email address"\}\},}
\hspace*{1em}\texttt{"required": ["name", "age", "email"]\}\}} \\
\hspace*{2em}\texttt{\},} \\
\hspace*{2em}\texttt{\{} \\
\hspace*{3em}\textbf{"name"}\texttt{: "calculate\_bmi",} \\
\hspace*{3em}\textbf{"description"}\texttt{: "Calculate the Body Mass Index (BMI) of a person",} \\
\hspace*{3em}\textbf{"parameters"}\texttt{: \{"type": "object", "properties": \{} \\
\hspace*{4em}\textbf{"weight"}\texttt{: \{"type": "number", "description": "The weight of the person in kilograms"\},} \\
\hspace*{4em}\textbf{"height"}\texttt{: \{"type": "number", "description": "The height of the person in meters"\}\},} \\
\hspace*{3em}\texttt{"required": ["weight", "height"]\}\}} \\
\hspace*{2em}\texttt{\},} \\
\hspace*{2em}\texttt{\{} \\
\hspace*{3em}\textbf{"name"}\texttt{: "schedule\_meeting",} \\
\hspace*{3em}\textbf{"description"}\texttt{: "Schedule a meeting with participants",} \\
\hspace*{3em}\textbf{"parameters"}\texttt{: \{"type": "object", "properties": \{} \\
\hspace*{4em}\textbf{"participants"}\texttt{: \{"type": "array", "items": \{"type": "string"\},} \\
\hspace*{5em}\texttt{"description": "The participants of the meeting"\},} \\
\hspace*{4em}\textbf{"meeting\_date"}\texttt{: \{"type": "string", "format": "date-time",} \\
\hspace*{5em}\texttt{"description": "The date and time of the meeting"\},} \\
\hspace*{4em}\textbf{"location"}\texttt{: \{"type": "string", "description": "The location of the meeting"\}\},} \\
\hspace*{3em}\texttt{"required": ["participants", "meeting\_date"]\}\}} \\
\hspace*{2em}\texttt{\},} \\
\hspace*{2em}\texttt{\{} \\
\hspace*{3em}\textbf{"name"}\texttt{: "create\_user",} \\
\hspace*{3em}\textbf{"description"}\texttt{: "Create a new user",} \\
\hspace*{3em}\textbf{"parameters"}\texttt{: \{"type": "object", "properties": \{} \\
\hspace*{4em}\textbf{"username"}\texttt{: \{"type": "string", "description": "The username of the user"\},} \\
\hspace*{4em}\textbf{"email"}\texttt{: \{"type": "string", "description": "The email address of the user"\},} \\
\hspace*{4em}\textbf{"password"}\texttt{: \{"type": "string", "description": "The password for the user"\}\},} \\
\hspace*{3em}\texttt{"required": ["username", "email", "password"]\}\}} \\
\hspace*{2em}\texttt{\},} 
\hspace*{2em}\texttt{\{} \\
\hspace*{3em}\textbf{"name"}\texttt{: "get\_definition",} \\
\hspace*{3em}\textbf{"description"}\texttt{: "Get the definition of a word",} \\
\hspace*{3em}\textbf{"parameters"}\texttt{: \{"type": "object", "properties": \{} \\
\hspace*{4em}\textbf{"word"}\texttt{: \{"type": "string", "description": "The word to get the definition"\},} \\
\hspace*{4em}\textbf{"language"}\texttt{: \{"type": "string", "description": "The language of the word (optional)"\}\},} \\
\hspace*{3em}\texttt{"required": ["word"]\}\}} \\
\hspace*{2em}\texttt{\},} \\
\hspace*{2em}\texttt{\{} \\
\hspace*{3em}\textbf{"name"}\texttt{: "calculate\_tip",} \\
\hspace*{3em}\textbf{"description"}\texttt{: "Calculate the tip amount for a bill",} \\
\hspace*{3em}\textbf{"parameters"}\texttt{: \{"type": "object", "properties": \{} \\
\hspace*{4em}\textbf{"bill\_amount"}\texttt{: \{"type": "number", "description": "The total amount of the bill"\},} \\
\hspace*{4em}\textbf{"tip\_percentage"}\texttt{: \{"type": "number", "description": "The tip percentage to be applied"\}\},} \\
\hspace*{3em}\texttt{"required": ["bill\_amount", "tip\_percentage"]\}\}} \\
\hspace*{2em}\texttt{\}],} \\
\hspace*{1em}\textbf{"question"}\texttt{: "Hi, I need a new password. Can you generate a random one for me?",} \\
\hspace*{1em}\textbf{"ground\_truth"}\texttt{: \{"name": "generate\_random\_password",} \\
\hspace*{4em}\texttt{"arguments": \{"length": 12\}\}} \\
\texttt{\}}
\end{tcolorbox}
\caption{Figure~\thefigure{} continued. The \texttt{question} and \texttt{ground\_truth}
  fields are identical across all three scales. Notable distractors include
  \texttt{create\_user} and \texttt{create\_user\_profile}, which both handle
  passwords as fields, and \texttt{schedule\_meeting} with its array-typed
  \texttt{participants} parameter.}
\label{fig:full-2433-15b}
\end{figure}

\clearpage
\section{TAAG Details and Correction Messages}
\label{app:fixes}

This appendix gives the deterministic definitions behind Section~\ref{sec:methodology}. TAAG evaluates a prediction against the prompt registry, the selected tool schema, and the user request before any ground-truth comparison. The source ground truth is used only for extrinsic precision, recall, F1, and solve-rate reporting.

\subsection{Cascaded Evaluation Pipeline}
\label{app:taag_pipeline}

The cascade has three intrinsic stages. First, registry conformance checks whether the emitted function name belongs to the shown registry. Second, structural completeness checks whether the predicted argument keys satisfy the selected schema. Third, argument grounding checks whether predicted argument values are supported by the user request. Evaluation halts at the first failed stage, so all later-stage metrics are stage-conditional.

For feedback-enabled conditions, the first failed stage $k$ produces a deterministic fix $\psi_k$. The fix is appended to the conversation history,
\begin{equation}
    \hat{y}^{(t+1)} = \mathcal{M}\!\left(H^{(t)} \circ \psi_k\right),
\end{equation}
where $H^{(t)}$ is the conversation history at iteration $t$. The loop repeats up to $D_\text{max}=15$ iterations and terminates early on a full pass. The same failure mode and metric values always produce the same correction string.

\subsection{Evaluation Metrics}
\label{app:taag_metrics}

\paragraph{Registry conformance.}
Let $\mathcal{R}$ be the set of registered function names and $\hat{y}_\text{name}$ be the predicted name. Function Name Exact Match (FNEM) is
\begin{equation}
    FNEM (\hat{y}_\text{name},\, \mathcal{R})=
    \begin{cases}
        -1 & \text{if } \hat{y}_\text{name} =  \varnothing \\
        0 & \text{if } \hat{y}_\text{name} \notin \mathcal{R} \\
        1 & \text{if } \hat{y}_\text{name} \in \mathcal{R}.
    \end{cases}
\end{equation}
When FNEM $=0$, TAAG computes a Nearest Candidate Score (NCS):
\begin{equation}
\begin{aligned}
\text{NCS}(\hat{y}_\text{name}, \mathcal{R})
&= \max_{c \in \mathcal{C}} \Big[
0.6\,\text{WRatio}(\hat{y}_\text{name}, c) \\
&\quad + 0.4\,\text{Jaccard}(\hat{y}_\text{name}, c)
\Big].
\end{aligned}
\end{equation}
Here, $\mathcal{C} \subseteq \mathcal{R}$ contains candidates sharing at least one trigram with $\hat{y}_\text{name}$. WRatio is the weighted partial ratio from RapidFuzz~\citep{rapidfuzz}, normalized to $[0,1]$, and Jaccard is computed over whitespace-tokenized lowercase strings. A high NCS with FNEM $=0$ indicates recoverable name drift; a low NCS indicates fabrication.

\paragraph{Structural completeness.}
Given a conformant function name, let $P$ be the predicted argument set, $A_\text{req}$ the required parameters, and $A_\text{all}$ all admissible parameters. Required Parameter Recall is
\begin{equation}
    \text{RPR} = \frac{|A_\text{req} \cap P|}{|A_\text{req}|}.
\end{equation}
For each out-of-schema parameter $p \in P \setminus A_\text{all}$, the Parameter Drift Score is
\begin{equation}
    \text{PDS}(p,\, A_\text{all}) = \max_{a \in A_\text{all}}\; \text{WRatio}(p,\, a) \;/\; 100.
\end{equation}
A parameter is drifted if $\text{PDS}\geq0.8$ and its nearest match is not already present as an exact prediction; otherwise it is spurious. The Spurious Parameter Rate is
\begin{equation}
    \text{SPR} = \frac{\big|\{p \in P \setminus A_\text{all} : \text{PDS}(p, A_\text{all}) < 0.8\}\big|}{|P|}.
\end{equation}
Stage~2 passes if and only if RPR $=1$, SPR $=0$, and no drifted parameters require correction.

\paragraph{Argument value grounding.}
For user query $Q$ and predicted values $v_i$, Argument Value Exact Match is
\begin{equation}
    \text{AVEM} = \frac{\big|\{i : \text{ExactMatch}(v_i, Q) = 1\}\big|}{|P|}.
\end{equation}
The check is type-aware: strings are matched case-insensitively at word boundaries, numeric values must match a floating-point token in $Q$, and list values require every element to satisfy the condition individually.

For non-exact string values, let $\mathcal{S}=\{i : \text{type}(v_i)=\text{string} \wedge \text{ExactMatch}(v_i,Q)=0\}$. Query-Supported Lexical Overlap is
\begin{equation}
    \text{QSLO} = \frac{1}{|\mathcal{S}|} \sum_{i \in \mathcal{S}} \text{PartialRatio}(v_i,\, Q) \;/\; 100.
\end{equation}
QSLO is lexical rather than semantic; it measures whether values are extractable from the query surface form and can penalize valid paraphrases.

Value Hallucination Rate uses the per-argument function
\begin{equation}
h(v_i,Q)=
\begin{cases}
0, &  \mathrm{EM}(v_i,Q)=1,\\
1-\frac{\mathrm{PR}(v_i,Q)}{100}, & \mathrm{str}(v_i),\\
1, & \mathrm{num}(v_i)\wedge \mathrm{HN}(Q),\\
\text{excluded}, & \text{otherwise.}
\end{cases}
\end{equation}
Here, $\mathrm{EM}$ denotes ExactMatch, $\mathrm{PR}$ denotes PartialRatio, and $\mathrm{HN}$ denotes HasNumbers. The excluded case covers arguments for which no grounding context is present in the query, such as a ticker symbol inferred from a company name.

The pass criterion is defined over the contextual subset $\mathcal{G} = \{i : \text{type}(v_i) = \text{string}\} \cup \{i : \text{type}(v_i) = \text{numeric} \wedge \text{HasNumbers}(Q)\}$:
\begin{equation}
    \text{VHR}_\text{ctx} = \sum_{i \in \mathcal{G}} h(v_i, Q) = 0.
\end{equation}
A non-zero $\text{VHR}_\text{ctx}$ triggers $\psi_3$, which separates numeric mismatches, hard-drifted strings ($\text{PartialRatio}<0.6$), and soft-drifted strings.

\subsection{Actionability Probes: Structured Correction Prompts}
\label{app:taag_corrections}

Table~\ref{tab:fixes} shows representative fix messages $\psi_k$ for each failure mode for TAAG-S and TAAG-R. For Binary, the feedback is just "No". All fixes are constructed programmatically from the evaluation output. No fix reveals the ground-truth function name, parameter set, or argument values; the correction is grounded solely in the registry schema and the model's own prediction.

\begin{table*}[!htbp]
\centering
\small
\renewcommand{\arraystretch}{1.4}
\begin{tabular}{p{1.8cm} p{2.8cm} p{7.8cm}}
\toprule
\textbf{Stage} & \textbf{Failure Mode} & \textbf{TAAG-S (Representative Fix Message)} \\
\midrule
Q1 & No name found      & \textit{No function name found. Please return only the tool invocation in the exact JSON format specified in the registry.} \\
Q1 & Wrong name         & \textit{Function name ``\{predicted\}'' not found in registry. Please check the system prompt for the correct function names and respond with only the tool invocation in the exact JSON format specified.} \\
\midrule
Q2 & Missing required   & \textit{Key parameters of the tool from the registry are missing. Please ensure the invocation includes all parameters specified for that tool in the system prompt.} \\
Q2 & Spurious params    & \textit{Spurious parameters not present in the registry are included. Please include only the parameters specified for that tool.} \\
Q2 & Drifted params     & \textit{Some parameters appear misspelled and deviate slightly from the registry. Please ensure all parameter names exactly match the schema in the system prompt.} \\
\midrule
Q3 & Numeric mismatch   & \textit{The numeric values for \{params\} do not match what the user stated. Use the exact numbers from the user's request.} \\
Q3 & Hard value drift   & \textit{The values for \{params\} have no grounding in the user's request. Do not invent values. Use only what the user explicitly stated, or omit the argument.} \\
Q3 & Soft value drift   & \textit{The values for \{params\} are loosely related but not exact. Use the precise wording or value from the user's request.} \\
\bottomrule
\end{tabular}
\caption{Representative structured correction messages $\psi_k$ per failure mode. Bracketed tokens (\{predicted\}, \{params\}) are instantiated programmatically from the evaluation output.}
\label{tab:fixes}
\end{table*}

\begin{table*}[!htbp]
\centering
\small
\renewcommand{\arraystretch}{1.4}
\begin{tabular}{p{1.8cm} p{2.8cm} p{7.8cm}}
\toprule
\textbf{Stage} & \textbf{Failure Mode} & \textbf{TAAG-R (Representative Fix Message)} \\
\midrule
Q1 & No name found
& \textit{FACT: check = Q1\_NO\_NAME | FACT function\_name = missing | RULE valid\_name := function\_name is present in registry | ACTION: Return Tool Invocation. JSON ONLY.} \\

Q1 & Wrong name 
& \textit{FACT: check = Q1\_WRONG\_NAME | FACT predicted\_name = \{predicted\} | FACT registry\_names = \{registry\_names\} | RULE valid\_name := predicted\_name matches exactly one registry\_names item | ACTION: Replace with Registered Name | ACTION: Return JSON ONLY.} \\

\midrule
Q2 & Missing required 
& \textit{FACT: check = Q2\_MISSING\_REQUIRED | FACT missing\_parameters = \{params\} | FACT required\_parameters = \{required\_params\} | RULE valid\_arguments := all required\_parameters are present | ACTION: Add Missing Parameters | ACTION: Return JSON ONLY.} \\

Q2 & Spurious params 
& \textit{FACT: check = Q2\_SPURIOUS\_PARAMS | FACT unexpected\_parameters = \{params\} | FACT allowed\_parameters = \{allowed\_params\} | RULE valid\_arguments := no unexpected\_parameters present | ACTION: Remove Extra Parameters | ACTION: Return JSON ONLY.} \\

Q2 & Drifted params 
& \textit{FACT: check = Q2\_DRIFTED\_PARAMS | FACT observed\_parameters = \{params\} | FACT registry\_parameters = \{registry\_params\} | RULE valid\_arguments := parameter\_names exactly match registry | ACTION: Correct Parameter Names | ACTION: Return JSON ONLY.} \\

\midrule
Q3 & Numeric mismatch 
& \textit{FACT: check = Q3\_NUMERIC\_MISMATCH | FACT mismatched\_params = \{params\} | FACT user\_values = \{user\_values\} | RULE grounded\_value := numeric values must exactly match user input | ACTION: Replace with User Values | ACTION: Return JSON ONLY.} \\

Q3 & Hard value drift
& \textit{FACT: check = Q3\_HARD\_VALUE\_DRIFT | FACT unsupported\_params = \{params\} | FACT user\_evidence = none | RULE grounded\_value := values must originate from user input | ACTION: Remove or Replace Values | ACTION: Do Not Invent Values | ACTION: Return JSON ONLY.} \\

Q3 & Soft value drift
& \textit{FACT: check = Q3\_SOFT\_VALUE\_DRIFT | FACT loosely\_matched\_params = \{params\} | FACT user\_phrasing = \{user\_values\} | RULE grounded\_value := values must exactly match user phrasing | ACTION: Align with User Input | ACTION: Return JSON ONLY.} \\

\bottomrule
\end{tabular}
\caption{Representative structured correction messages $\psi_k$ per failure mode in TAAG-R. Each message follows a FACT–RULE–ACTION template. Bracketed tokens (\{predicted\}, \{params\}, \{registry\_names\}) are instantiated programmatically from the evaluation output.}
\label{tab:taag_s_fixes}
\end{table*}

\clearpage
\section{Models and Inference Settings}
\label{app:models}

Table~\ref{tab:model_suite} lists the exact locally served model identifiers used in the experimental grid. We use Ollama for all experiments, apply the same inference wrapper across models, and do not fine-tune any model on RFCD. The model grouping follows public model cards and documentation: compact instruction-following baselines, reasoning and thinking models, and models selected for tool or structured-output awareness under the same registry and feedback conditions.

\begin{table}[h]
\centering
\small
\renewcommand{\arraystretch}{1.2}
\begin{tabular}{lll}
\toprule
\textbf{Model tag} & \textbf{Display name} & \textbf{Analysis class} \\
\midrule
\texttt{llama3.2:3b} & Llama~3.2:3B & Compact instruction-following baseline \\
\texttt{qwen2.5:3b} & Qwen~2.5:3B & Instruction-following and JSON-capable baseline \\
\texttt{gemma4:e4b} & Gemma~4:e4B & Reasoning and agentic-workflow model \\
\texttt{qwen3:4b-thinking} & Qwen~3:4B-Thinking & Thinking-only reasoning model \\
\texttt{phi4-mini:3.8b} & Phi-4-mini:3.8B & Tool-enabled function-calling model \\
\texttt{granite4:3b} & Granite~4:3B & Tool-calling instruction model \\
\bottomrule \\
\end{tabular}
\caption{Exact model identifiers and analysis classes used in the experiments.}
\label{tab:model_suite}
\end{table}

\subsection{Model Selection Rationale}
\label{app:model_selection}

Llama~3.2:3B and Qwen~2.5:3B are compact instruction-following baselines. Llama~3.2:3B is a lightweight multilingual text model intended for assistant-like chat, retrieval, summarization, and prompt rewriting, while Qwen~2.5:3B is an instruction-tuned model with reported improvements in instruction following, long-text generation, structured-data understanding, and JSON output. This pair anchors how broadly trained compact instruction models behave before adding explicit thinking or tool-call-specific formatting.

Gemma~4:e4B and Qwen~3:4B-Thinking form the reasoning and thinking group. Gemma~4 is presented as a reasoning-oriented open model family for agentic workflows, with the E4B variant designed for efficient local deployment, while Qwen~3:4B-Thinking is a thinking-only model whose release emphasizes deeper reasoning and tool-use improvements. This group tests whether reasoning-oriented behavior changes registry selection and correction under scale.

Phi-4-mini:3.8B and Granite~4:3B form the tool and structured-output-aware group. Phi-4-mini provides a tool-enabled function-calling format, and Granite~4 emphasizes instruction following and tool-calling for agentic and enterprise workflows. This group is included because RFCD requires both selecting a valid function and emitting a schema-conformant call.

The grouping is analytic rather than a claim that the models differ along only one axis. Its purpose is to compare broad behavior classes under the same registry and feedback conditions while keeping model scale small enough for exhaustive sweeps across datasets, registry sizes, and feedback modes. Small models are a deliberate stress-test choice: they make failures observable, keep the benchmark reproducible without proprietary frontier systems, and prevent the experiment from becoming primarily a measurement of very large model capacity.

\subsection{Inference Settings}
\label{app:inference_settings}

All runs are served through Ollama with deterministic decoding and a fixed prompt format within each feedback mode. Single-pass runs use one attempt. Binary-feedback and structured-feedback runs share the same maximum correction budget, $D_\text{max}=15$, and terminate early once the emitted tool call passes the evaluation. For Qwen~3:4B-Thinking, the explicit thinking configuration is enabled. Gemma~4:e4B is grouped with reasoning models because its public release emphasizes built-in reasoning and agentic workflows. The remaining four models are evaluated as standard instruction-following models.

\clearpage
\section{Complete Results and Additional Views}
\label{app:res}

This appendix records result views that are not included in the main paper. The main paper uses the clearest line and table views for the core claims; the appendix keeps additional views for source, scale, model-family, and cost interpretation. All aggregates use completed non-error instances, and all table entries report mean $\pm$ standard deviation over configuration-level aggregates unless the caption states otherwise. We use SP for the single-pass baseline, BF for binary feedback, TAAG-S for stage-diagnosis feedback, and TAAG-R for requirement-diagnosis feedback.

\subsection{Registry Scale and Context Pressure}
\label{app:rs_cp}

\begin{figure*}[t]
    \centering
    \includegraphics[width=\textwidth]{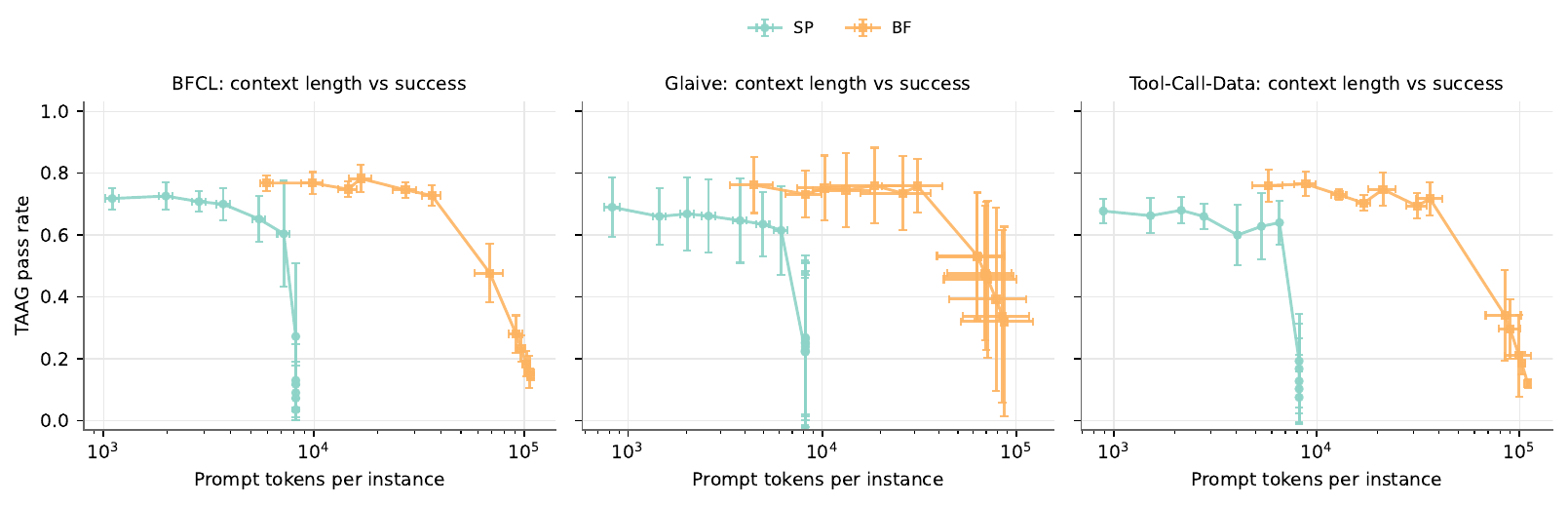}
    \caption{Long-context pressure from larger registries. Each panel is one source. The horizontal axis is prompt tokens per instance and the vertical axis is TAAG pass rate. Points aggregate configurations at the same registry size and feedback mode; horizontal and vertical error bars show standard deviations over configuration-level aggregates.}
    \label{fig:appendix_context_degradation}
\end{figure*}

Table~\ref{tab:appendix_scale_detail} reports the retained per-scale TAAG pass rates rather than only the small/large summary in the main paper. Table~\ref{tab:appendix_context_pressure} isolates the context-length component of the same effect by pairing pass-rate changes with measured prompt-token growth. Figure~\ref{fig:appendix_context_degradation} plots the same relationship directly: prompt length is on the horizontal axis and TAAG pass rate is on the vertical axis. Larger registries hurt because they add many executable candidates to the prompt, not because the evaluation quietly switches to a different task.

\begin{table*}[t]
\centering
\footnotesize
\setlength{\tabcolsep}{2.5pt}
\renewcommand{\arraystretch}{1.08}
\begin{adjustbox}{max width=\textwidth}
\begin{tabular}{rcccccc}
\toprule
\textbf{Registry size} & \textbf{BFCL single} & \textbf{BFCL binary} & \textbf{Glaive single} & \textbf{Glaive binary} & \textbf{Tool-Call single} & \textbf{Tool-Call binary} \\
\midrule
5 & \cellcolor[HTML]{E5F5E0}0.718 $\pm$ 0.035 & \cellcolor[HTML]{D9F0D3}0.768 $\pm$ 0.026 & \cellcolor[HTML]{E5F5E0}0.690 $\pm$ 0.095 & \cellcolor[HTML]{D9F0D3}0.762 $\pm$ 0.091 & \cellcolor[HTML]{E5F5E0}0.677 $\pm$ 0.040 & \cellcolor[HTML]{D9F0D3}0.760 $\pm$ 0.052 \\
10 & \cellcolor[HTML]{E5F5E0}0.726 $\pm$ 0.045 & \cellcolor[HTML]{D9F0D3}0.768 $\pm$ 0.036 & \cellcolor[HTML]{E5F5E0}0.660 $\pm$ 0.091 & \cellcolor[HTML]{E5F5E0}0.732 $\pm$ 0.076 & \cellcolor[HTML]{E5F5E0}0.662 $\pm$ 0.057 & \cellcolor[HTML]{D9F0D3}0.765 $\pm$ 0.039 \\
15 & \cellcolor[HTML]{E5F5E0}0.708 $\pm$ 0.033 & \cellcolor[HTML]{E5F5E0}0.748 $\pm$ 0.026 & \cellcolor[HTML]{E5F5E0}0.668 $\pm$ 0.118 & \cellcolor[HTML]{D9F0D3}0.752 $\pm$ 0.105 & \cellcolor[HTML]{E5F5E0}0.680 $\pm$ 0.042 & \cellcolor[HTML]{E5F5E0}0.730 $\pm$ 0.017 \\
20 & \cellcolor[HTML]{E5F5E0}0.700 $\pm$ 0.052 & \cellcolor[HTML]{D9F0D3}0.782 $\pm$ 0.045 & \cellcolor[HTML]{E5F5E0}0.662 $\pm$ 0.118 & \cellcolor[HTML]{E5F5E0}0.745 $\pm$ 0.121 & \cellcolor[HTML]{E5F5E0}0.660 $\pm$ 0.041 & \cellcolor[HTML]{E5F5E0}0.703 $\pm$ 0.025 \\
30 & \cellcolor[HTML]{E5F5E0}0.652 $\pm$ 0.074 & \cellcolor[HTML]{E5F5E0}0.746 $\pm$ 0.025 & \cellcolor[HTML]{E5F5E0}0.647 $\pm$ 0.136 & \cellcolor[HTML]{D9F0D3}0.760 $\pm$ 0.122 & \cellcolor[HTML]{E5F5E0}0.600 $\pm$ 0.098 & \cellcolor[HTML]{E5F5E0}0.747 $\pm$ 0.054 \\
40 & \cellcolor[HTML]{E5F5E0}0.604 $\pm$ 0.171 & \cellcolor[HTML]{E5F5E0}0.728 $\pm$ 0.034 & \cellcolor[HTML]{E5F5E0}0.635 $\pm$ 0.103 & \cellcolor[HTML]{E5F5E0}0.735 $\pm$ 0.120 & \cellcolor[HTML]{E5F5E0}0.628 $\pm$ 0.107 & \cellcolor[HTML]{E5F5E0}0.694 $\pm$ 0.041 \\
50 & \cellcolor[HTML]{F2FAEF}0.272 $\pm$ 0.236 & \cellcolor[HTML]{F2FAEF}0.477 $\pm$ 0.095 & \cellcolor[HTML]{E5F5E0}0.615 $\pm$ 0.143 & \cellcolor[HTML]{D9F0D3}0.758 $\pm$ 0.087 & \cellcolor[HTML]{E5F5E0}0.640 $\pm$ 0.070 & \cellcolor[HTML]{E5F5E0}0.718 $\pm$ 0.053 \\
100 & \cellcolor[HTML]{FFF7EC}0.130 $\pm$ 0.118 & \cellcolor[HTML]{F2FAEF}0.280 $\pm$ 0.060 & \cellcolor[HTML]{FFF7EC}0.238 $\pm$ 0.275 & \cellcolor[HTML]{F2FAEF}0.320 $\pm$ 0.307 & \cellcolor[HTML]{FFF7EC}0.193 $\pm$ 0.151 & \cellcolor[HTML]{F2FAEF}0.340 $\pm$ 0.147 \\
200 & \cellcolor[HTML]{FFF7EC}0.090 $\pm$ 0.088 & \cellcolor[HTML]{FFF7EC}0.233 $\pm$ 0.042 & \cellcolor[HTML]{F2FAEF}0.263 $\pm$ 0.244 & \cellcolor[HTML]{F2FAEF}0.392 $\pm$ 0.297 & \cellcolor[HTML]{FFF7EC}0.168 $\pm$ 0.145 & \cellcolor[HTML]{F2FAEF}0.297 $\pm$ 0.096 \\
300 & \cellcolor[HTML]{FFF7EC}0.035 $\pm$ 0.035 & \cellcolor[HTML]{FFF7EC}0.157 $\pm$ 0.051 & \cellcolor[HTML]{FFF7EC}0.222 $\pm$ 0.241 & \cellcolor[HTML]{F2FAEF}0.338 $\pm$ 0.279 & \cellcolor[HTML]{FFF7EC}0.128 $\pm$ 0.139 & \cellcolor[HTML]{FFF7EC}0.185 $\pm$ 0.035 \\
400 & \cellcolor[HTML]{FFF7EC}0.118 $\pm$ 0.074 & \cellcolor[HTML]{FFF7EC}0.183 $\pm$ 0.042 & \cellcolor[HTML]{F2FAEF}0.252 $\pm$ 0.266 & \cellcolor[HTML]{E5F5E0}0.530 $\pm$ 0.207 & \cellcolor[HTML]{FFF7EC}0.075 $\pm$ 0.083 & \cellcolor[HTML]{FFF7EC}0.120 $\pm$ 0.014 \\
500 & \cellcolor[HTML]{FFF7EC}0.072 $\pm$ 0.043 & \cellcolor[HTML]{FFF7EC}0.143 $\pm$ 0.006 & \cellcolor[HTML]{FFF7EC}0.224 $\pm$ 0.246 & \cellcolor[HTML]{F2FAEF}0.457 $\pm$ 0.255 & \cellcolor[HTML]{FFF7EC}0.103 $\pm$ 0.109 & \cellcolor[HTML]{FFF7EC}0.210 $\pm$ 0.132 \\
1000 & \cellcolor[HTML]{F2F2F2}n/a & \cellcolor[HTML]{F2F2F2}n/a & \cellcolor[HTML]{FFF7EC}0.226 $\pm$ 0.251 & \cellcolor[HTML]{F2FAEF}0.477 $\pm$ 0.218 & \cellcolor[HTML]{F2F2F2}n/a & \cellcolor[HTML]{F2F2F2}n/a \\
2000 & \cellcolor[HTML]{F2F2F2}n/a & \cellcolor[HTML]{F2F2F2}n/a & \cellcolor[HTML]{FFF7EC}0.248 $\pm$ 0.234 & \cellcolor[HTML]{E5F5E0}0.533 $\pm$ 0.205 & \cellcolor[HTML]{F2F2F2}n/a & \cellcolor[HTML]{F2F2F2}n/a \\
3000 & \cellcolor[HTML]{F2F2F2}n/a & \cellcolor[HTML]{F2F2F2}n/a & \cellcolor[HTML]{F2FAEF}0.268 $\pm$ 0.266 & \cellcolor[HTML]{F2FAEF}0.468 $\pm$ 0.239 & \cellcolor[HTML]{F2F2F2}n/a & \cellcolor[HTML]{F2F2F2}n/a \\
\bottomrule
\end{tabular}
\end{adjustbox}
\caption{TAAG pass rate at each registry size for single-pass and binary-feedback runs. Values are means with standard deviations over configuration-level aggregates. Cells with ``n/a'' mark source--scale combinations outside the retained aggregate analysis.}
\label{tab:appendix_scale_detail}
\end{table*}

\begin{table*}[t]
\centering
\footnotesize
\setlength{\tabcolsep}{2pt}
\renewcommand{\arraystretch}{1.1}
\begin{adjustbox}{max width=\textwidth}
\begin{tabular}{llccccc}
\toprule
\textbf{Source} & \textbf{Mode} & \textbf{Small prompt} & \textbf{Large prompt} & \textbf{Small pass} & \textbf{Large pass} & \textbf{Pass drop} \\
\midrule
BFCL & SP & \cellcolor[HTML]{D9EAF7}4384 $\pm$ 2498 & \cellcolor[HTML]{D9EAF7}8192 $\pm$ 0 & \cellcolor[HTML]{E5F5E0}0.626 $\pm$ 0.184 & \cellcolor[HTML]{FFF7EC}0.091 $\pm$ 0.080 & \cellcolor[HTML]{E5F5E0}0.535 \\
BFCL & BF & \cellcolor[HTML]{E8F2FB}22759 $\pm$ 18442 & \cellcolor[HTML]{FFF7EC}100456 $\pm$ 7358 & \cellcolor[HTML]{E5F5E0}0.730 $\pm$ 0.093 & \cellcolor[HTML]{FFF7EC}0.199 $\pm$ 0.064 & \cellcolor[HTML]{E5F5E0}0.530 \\
Glaive & SP & \cellcolor[HTML]{D9EAF7}3107 $\pm$ 1834 & \cellcolor[HTML]{D9EAF7}8184 $\pm$ 24 & \cellcolor[HTML]{E5F5E0}0.654 $\pm$ 0.110 & \cellcolor[HTML]{FFF7EC}0.244 $\pm$ 0.231 & \cellcolor[HTML]{F2FAEF}0.410 \\
Glaive & BF & \cellcolor[HTML]{E8F2FB}16156 $\pm$ 11039 & \cellcolor[HTML]{FFF7EC}74700 $\pm$ 27378 & \cellcolor[HTML]{E5F5E0}0.750 $\pm$ 0.098 & \cellcolor[HTML]{F2FAEF}0.427 $\pm$ 0.240 & \cellcolor[HTML]{F2FAEF}0.323 \\
Tool-Call-Data & SP & \cellcolor[HTML]{D9EAF7}3285 $\pm$ 1899 & \cellcolor[HTML]{D9EAF7}8192 $\pm$ 0 & \cellcolor[HTML]{E5F5E0}0.649 $\pm$ 0.069 & \cellcolor[HTML]{FFF7EC}0.133 $\pm$ 0.122 & \cellcolor[HTML]{E5F5E0}0.516 \\
Tool-Call-Data & BF & \cellcolor[HTML]{E8F2FB}18863 $\pm$ 11252 & \cellcolor[HTML]{FFF7EC}95846 $\pm$ 13699 & \cellcolor[HTML]{E5F5E0}0.734 $\pm$ 0.046 & \cellcolor[HTML]{FFF7EC}0.242 $\pm$ 0.121 & \cellcolor[HTML]{F2FAEF}0.491 \\
\bottomrule
\end{tabular}
\end{adjustbox}
\caption{Long-context pressure induced by registry scale. Prompt tokens are measured from the logs and reported per instance; pass-rate drops compare the retained small-registry and large-registry bands.}
\label{tab:appendix_context_pressure}
\end{table*}

\begin{figure*}[t]
    \centering
    \includegraphics[width=\textwidth]{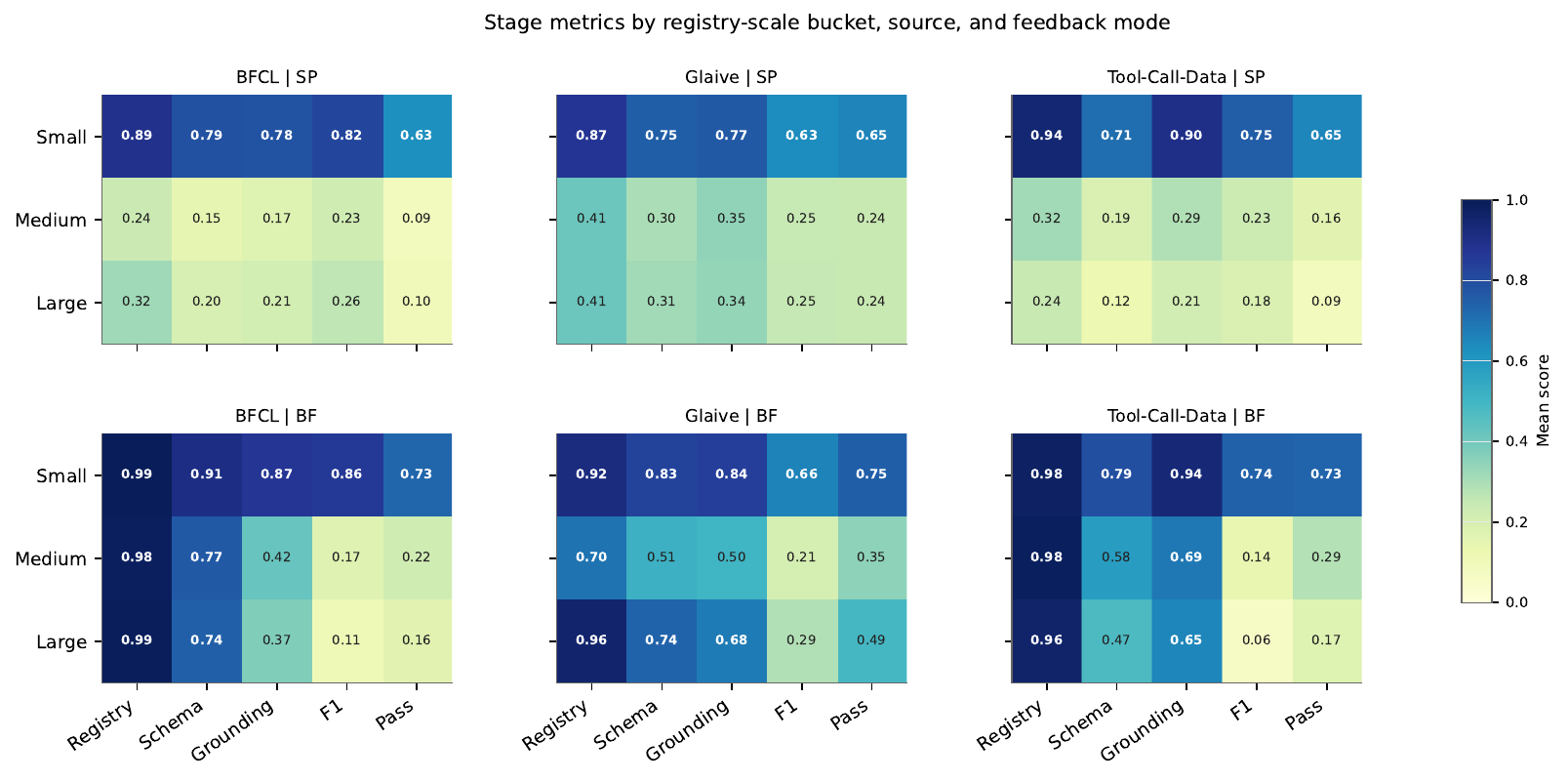}
    \caption{Stage metrics by registry-scale bucket, source, and feedback mode. Each cell is a mean score over configuration-level aggregates. Small means $S\leq50$, medium means $100\leq S\leq300$, and large means $S\geq400$ in the retained registry sizes. Color and printed values use the same 0--1 scale; grey n/a cells mark unavailable retained aggregates.}
    \label{fig:appendix_stage_scale_heatmap}
\end{figure*}

Table~\ref{tab:appendix_scale_bin_degradation} gives the compact small/large view used for the main narrative. Figure~\ref{fig:appendix_stage_scale_heatmap} then separates the scale effect across the TAAG stages: registry conformance, schema completeness, grounding, extrinsic F1, and final pass rate. The same final pass-rate drop can have different causes: some source--mode pairs mainly lose valid function selection, while others keep names valid but lose schema or grounding quality.

\begin{table*}[t]
\centering
\footnotesize
\setlength{\tabcolsep}{2.5pt}
\renewcommand{\arraystretch}{1.12}
\begin{tabular}{llcccc}
\toprule
\textbf{Source} & \textbf{Metric} & \textbf{SP} & \textbf{BF} & \textbf{TAAG-S} & \textbf{TAAG-R} \\
\midrule
BFCL & TAAG pass rate & \cellcolor[HTML]{F2FAEF}0.429 $\pm$ 0.302 & \cellcolor[HTML]{E5F5E0}0.557 $\pm$ 0.265 & \cellcolor[HTML]{E5F5E0}0.579 $\pm$ 0.404 & \cellcolor[HTML]{E5F5E0}0.684 $\pm$ 0.382 \\
 & Extrinsic F1 & \cellcolor[HTML]{E5F5E0}0.605 $\pm$ 0.353 & \cellcolor[HTML]{E5F5E0}0.628 $\pm$ 0.358 & \cellcolor[HTML]{D9F0D3}0.967 $\pm$ 0.060 & \cellcolor[HTML]{D9F0D3}0.949 $\pm$ 0.067 \\
 & Grounding score & \cellcolor[HTML]{E5F5E0}0.562 $\pm$ 0.348 & \cellcolor[HTML]{E5F5E0}0.717 $\pm$ 0.251 & \cellcolor[HTML]{F2FAEF}0.490 $\pm$ 0.071 & \cellcolor[HTML]{E5F5E0}0.555 $\pm$ 0.089 \\
 & Tokens & \cellcolor[HTML]{D9EAF7}5879 $\pm$ 2749 & \cellcolor[HTML]{F3F8FD}48550 $\pm$ 40107 & \cellcolor[HTML]{F3F8FD}48629 $\pm$ 49146 & \cellcolor[HTML]{F3F8FD}39421 $\pm$ 42726 \\
\midrule
Glaive & TAAG pass rate & \cellcolor[HTML]{F2FAEF}0.446 $\pm$ 0.274 & \cellcolor[HTML]{E5F5E0}0.616 $\pm$ 0.234 & \cellcolor[HTML]{E5F5E0}0.553 $\pm$ 0.390 & \cellcolor[HTML]{E5F5E0}0.573 $\pm$ 0.388 \\
 & Extrinsic F1 & \cellcolor[HTML]{F2FAEF}0.440 $\pm$ 0.253 & \cellcolor[HTML]{F2FAEF}0.489 $\pm$ 0.228 & \cellcolor[HTML]{E5F5E0}0.650 $\pm$ 0.281 & \cellcolor[HTML]{E5F5E0}0.644 $\pm$ 0.269 \\
 & Grounding score & \cellcolor[HTML]{E5F5E0}0.550 $\pm$ 0.311 & \cellcolor[HTML]{E5F5E0}0.740 $\pm$ 0.235 & \cellcolor[HTML]{E5F5E0}0.555 $\pm$ 0.071 & \cellcolor[HTML]{E5F5E0}0.620 $\pm$ 0.089 \\
 & Tokens & \cellcolor[HTML]{D9EAF7}5785 $\pm$ 2877 & \cellcolor[HTML]{F3F8FD}41225 $\pm$ 34903 & \cellcolor[HTML]{F3F8FD}42385 $\pm$ 39026 & \cellcolor[HTML]{F3F8FD}45635 $\pm$ 39938 \\
\midrule
Tool-Call-Data & TAAG pass rate & \cellcolor[HTML]{F2FAEF}0.434 $\pm$ 0.274 & \cellcolor[HTML]{E5F5E0}0.566 $\pm$ 0.249 & \cellcolor[HTML]{E5F5E0}0.537 $\pm$ 0.327 & \cellcolor[HTML]{E5F5E0}0.586 $\pm$ 0.299 \\
 & Extrinsic F1 & \cellcolor[HTML]{E5F5E0}0.527 $\pm$ 0.284 & \cellcolor[HTML]{E5F5E0}0.526 $\pm$ 0.307 & \cellcolor[HTML]{D9F0D3}0.819 $\pm$ 0.242 & \cellcolor[HTML]{D9F0D3}0.802 $\pm$ 0.208 \\
 & Grounding score & \cellcolor[HTML]{E5F5E0}0.630 $\pm$ 0.349 & \cellcolor[HTML]{D9F0D3}0.846 $\pm$ 0.138 & \cellcolor[HTML]{F2FAEF}0.425 $\pm$ 0.089 & \cellcolor[HTML]{F2FAEF}0.490 $\pm$ 0.107 \\
 & Tokens & \cellcolor[HTML]{D9EAF7}5423 $\pm$ 2873 & \cellcolor[HTML]{F3F8FD}45574 $\pm$ 39017 & \cellcolor[HTML]{F3F8FD}50403 $\pm$ 46326 & \cellcolor[HTML]{F3F8FD}50556 $\pm$ 41564 \\
\bottomrule
\end{tabular}
\caption{Source-level comparison across all retained feedback modes. SP is single pass, BF is binary feedback, TAAG-S is stage-only TAAG feedback, and TAAG-R is requirement-level TAAG feedback. Values are means with standard deviations over retained configurations. Green cells indicate higher-is-better score metrics, and blue cells indicate lower-is-better token cost.}
\label{tab:appendix_source_mode_summary}
\end{table*}

\begin{table*}[t]
\centering
\footnotesize
\setlength{\tabcolsep}{2.5pt}
\renewcommand{\arraystretch}{1.12}
\begin{tabular}{llcccc}
\toprule
\textbf{Source} & \textbf{Mode} & \textbf{Small pass} & \textbf{Large pass} & \textbf{Pass drop} & \textbf{Token ratio} \\
\midrule
BFCL & SP & \cellcolor[HTML]{E5F5E0}0.626 $\pm$ 0.184 & \cellcolor[HTML]{FFF7EC}0.091 $\pm$ 0.080 & \cellcolor[HTML]{E5F5E0}0.535 & \cellcolor[HTML]{D9EAF7}1.9$\times$ \\
BFCL & BF & \cellcolor[HTML]{E5F5E0}0.730 $\pm$ 0.093 & \cellcolor[HTML]{FFF7EC}0.199 $\pm$ 0.064 & \cellcolor[HTML]{E5F5E0}0.530 & \cellcolor[HTML]{D9EAF7}4.4$\times$ \\
BFCL & TAAG-S & \cellcolor[HTML]{D9F0D3}0.855 $\pm$ 0.258 & \cellcolor[HTML]{FFF7EC}0.200 $\pm$ 0.213 & \cellcolor[HTML]{E5F5E0}0.655 & \cellcolor[HTML]{D9EAF7}10.4$\times$ \\
BFCL & TAAG-R & \cellcolor[HTML]{D9F0D3}0.848 $\pm$ 0.289 & \cellcolor[HTML]{F2FAEF}0.469 $\pm$ 0.389 & \cellcolor[HTML]{F2FAEF}0.379 & \cellcolor[HTML]{D9EAF7}6.9$\times$ \\
Glaive & SP & \cellcolor[HTML]{E5F5E0}0.654 $\pm$ 0.110 & \cellcolor[HTML]{FFF7EC}0.244 $\pm$ 0.231 & \cellcolor[HTML]{F2FAEF}0.410 & \cellcolor[HTML]{D9EAF7}2.6$\times$ \\
Glaive & BF & \cellcolor[HTML]{E5F5E0}0.750 $\pm$ 0.098 & \cellcolor[HTML]{F2FAEF}0.427 $\pm$ 0.240 & \cellcolor[HTML]{F2FAEF}0.323 & \cellcolor[HTML]{D9EAF7}4.4$\times$ \\
Glaive & TAAG-S & \cellcolor[HTML]{E5F5E0}0.560 $\pm$ 0.414 & \cellcolor[HTML]{E5F5E0}0.546 $\pm$ 0.369 & \cellcolor[HTML]{FFF7EC}0.014 & \cellcolor[HTML]{D9EAF7}3.2$\times$ \\
Glaive & TAAG-R & \cellcolor[HTML]{E5F5E0}0.618 $\pm$ 0.391 & \cellcolor[HTML]{E5F5E0}0.532 $\pm$ 0.384 & \cellcolor[HTML]{FFF7EC}0.086 & \cellcolor[HTML]{D9EAF7}2.6$\times$ \\
Tool-Call-Data & SP & \cellcolor[HTML]{E5F5E0}0.649 $\pm$ 0.069 & \cellcolor[HTML]{FFF7EC}0.133 $\pm$ 0.122 & \cellcolor[HTML]{E5F5E0}0.516 & \cellcolor[HTML]{D9EAF7}2.5$\times$ \\
Tool-Call-Data & BF & \cellcolor[HTML]{E5F5E0}0.734 $\pm$ 0.046 & \cellcolor[HTML]{FFF7EC}0.242 $\pm$ 0.121 & \cellcolor[HTML]{F2FAEF}0.491 & \cellcolor[HTML]{D9EAF7}5.0$\times$ \\
Tool-Call-Data & TAAG-S & \cellcolor[HTML]{D9F0D3}0.779 $\pm$ 0.107 & \cellcolor[HTML]{FFF7EC}0.200 $\pm$ 0.205 & \cellcolor[HTML]{E5F5E0}0.579 & \cellcolor[HTML]{D9EAF7}7.1$\times$ \\
Tool-Call-Data & TAAG-R & \cellcolor[HTML]{D9F0D3}0.793 $\pm$ 0.098 & \cellcolor[HTML]{F2FAEF}0.323 $\pm$ 0.258 & \cellcolor[HTML]{F2FAEF}0.470 & \cellcolor[HTML]{D9EAF7}4.9$\times$ \\
\bottomrule
\end{tabular}
\caption{Registry-scale degradation by source and feedback mode. Small means registry sizes $S\leq50$ and large means $S\geq100$ in the retained aggregate. Pass drop is small-band pass rate minus large-band pass rate. Token ratio is large-band token use divided by small-band token use. Green cells indicate higher-is-better pass rates, and blue cells indicate lower-is-better token ratios.}
\label{tab:appendix_scale_bin_degradation}
\end{table*}

\begin{table}[t]
\centering
\small
\setlength{\tabcolsep}{3pt}
\begin{tabular}{lccc}
\toprule
\textbf{Mode} & \textbf{TAAG pass rate} & \textbf{Attempts} & \textbf{Tokens} \\
\midrule
SP & \cellcolor[HTML]{F2FAEF}0.438 $\pm$ 0.281 & \cellcolor[HTML]{D9EAF7}1.0 $\pm$ 0.0 & \cellcolor[HTML]{D9EAF7}5722 $\pm$ 2829 \\
BF, D<=5 & \cellcolor[HTML]{E5F5E0}0.617 $\pm$ 0.243 & \cellcolor[HTML]{E8F2FB}2.6 $\pm$ 1.0 & \cellcolor[HTML]{E8F2FB}15165 $\pm$ 12165 \\
BF, D<=10 & \cellcolor[HTML]{E5F5E0}0.626 $\pm$ 0.246 & \cellcolor[HTML]{E8F2FB}4.5 $\pm$ 2.2 & \cellcolor[HTML]{E8F2FB}27190 $\pm$ 24070 \\
BF, D<=15 & \cellcolor[HTML]{E5F5E0}0.629 $\pm$ 0.247 & \cellcolor[HTML]{F3F8FD}6.4 $\pm$ 3.4 & \cellcolor[HTML]{F3F8FD}39286 $\pm$ 35963 \\
TAAG-S & \cellcolor[HTML]{E5F5E0}0.554 $\pm$ 0.377 & \cellcolor[HTML]{F3F8FD}7.4 $\pm$ 5.2 & \cellcolor[HTML]{F3F8FD}45434 $\pm$ 42756 \\
TAAG-R & \cellcolor[HTML]{E5F5E0}0.596 $\pm$ 0.368 & \cellcolor[HTML]{F3F8FD}7.0 $\pm$ 5.1 & \cellcolor[HTML]{F3F8FD}45714 $\pm$ 40803 \\
\bottomrule \\
\end{tabular}
\caption{Feedback-cost summary across SP, BF depth budgets, and TAAG feedback variants. BF rows vary the maximum correction budget $D$; TAAG-S and TAAG-R use $D\leq15$ and report retained summary-log values. Tokens are per instance. Green cells indicate higher-is-better pass rate, and blue cells indicate lower-is-better attempts and token use.}
\label{tab:appendix_binary_depth}
\end{table}

\subsection{Iteration Budget and Feedback Depth}
\label{app:rs_itbf}

\begin{figure*}[t]
    \centering
    \includegraphics[width=0.88\textwidth]{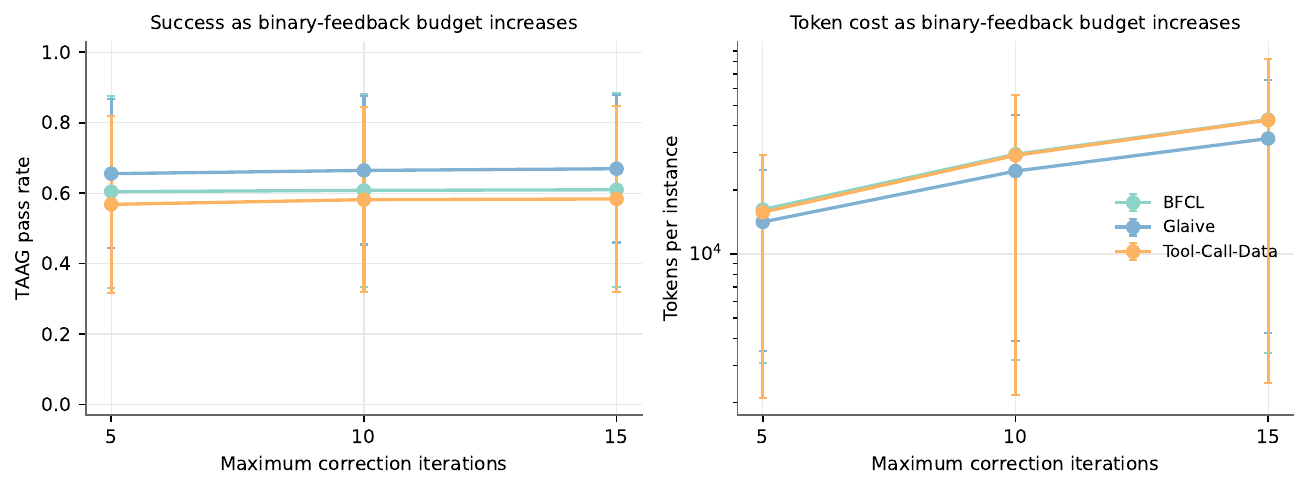}
    \caption{Feedback-depth tradeoff by source. The left panel reports TAAG pass rate, and the right panel reports tokens per instance. SP has one attempt, BF rows vary the maximum correction budget $D$, and TAAG-S/TAAG-R rows use the retained structured-feedback summaries. Error bars show standard deviations over configuration-level aggregates.}
    \label{fig:appendix_iteration_depth_by_source}
\end{figure*}

The main paper reports the overall binary-feedback depth tradeoff. Tables~\ref{tab:appendix_iteration_by_source} and~\ref{tab:appendix_iteration_by_family} break the same experiment down by source and model family. Figure~\ref{fig:appendix_iteration_depth_by_source} gives the corresponding visual summary: the left panel shows success and the right panel shows token cost as the maximum correction budget increases. Most of the observed binary-feedback gain appears early, while additional iterations mainly add cost.

\begin{table*}[t]
\centering
\small
\setlength{\tabcolsep}{3pt}
\renewcommand{\arraystretch}{1.1}
\begin{adjustbox}{max width=\textwidth}
\begin{tabular}{llccc}
\toprule
\textbf{Source} & \textbf{Mode} & \textbf{TAAG pass rate} & \textbf{Attempts} & \textbf{Tokens} \\
\midrule
BFCL & SP & \cellcolor[HTML]{F2FAEF}0.429 $\pm$ 0.302 & \cellcolor[HTML]{D9EAF7}1.0 $\pm$ 0.0 & \cellcolor[HTML]{D9EAF7}5879 $\pm$ 2749 \\
BFCL & BF, D<=5 & \cellcolor[HTML]{E5F5E0}0.604 $\pm$ 0.273 & \cellcolor[HTML]{E8F2FB}2.7 $\pm$ 1.1 & \cellcolor[HTML]{E8F2FB}16161 $\pm$ 13073 \\
BFCL & BF, D<=10 & \cellcolor[HTML]{E5F5E0}0.608 $\pm$ 0.275 & \cellcolor[HTML]{E8F2FB}4.6 $\pm$ 2.4 & \cellcolor[HTML]{E8F2FB}29416 $\pm$ 26241 \\
BFCL & BF, D<=15 & \cellcolor[HTML]{E5F5E0}0.610 $\pm$ 0.276 & \cellcolor[HTML]{F3F8FD}6.6 $\pm$ 3.8 & \cellcolor[HTML]{F3F8FD}42787 $\pm$ 39369 \\
BFCL & TAAG-S & \cellcolor[HTML]{E5F5E0}0.579 $\pm$ 0.404 & \cellcolor[HTML]{F3F8FD}7.0 $\pm$ 5.6 & \cellcolor[HTML]{F3F8FD}48629 $\pm$ 49146 \\
BFCL & TAAG-R & \cellcolor[HTML]{E5F5E0}0.684 $\pm$ 0.382 & \cellcolor[HTML]{F3F8FD}5.8 $\pm$ 5.3 & \cellcolor[HTML]{F3F8FD}39421 $\pm$ 42726 \\
\midrule
Glaive & SP & \cellcolor[HTML]{F2FAEF}0.446 $\pm$ 0.274 & \cellcolor[HTML]{D9EAF7}1.0 $\pm$ 0.0 & \cellcolor[HTML]{D9EAF7}5785 $\pm$ 2877 \\
Glaive & BF, D<=5 & \cellcolor[HTML]{E5F5E0}0.655 $\pm$ 0.211 & \cellcolor[HTML]{E8F2FB}2.5 $\pm$ 0.9 & \cellcolor[HTML]{E8F2FB}14142 $\pm$ 10646 \\
Glaive & BF, D<=10 & \cellcolor[HTML]{E5F5E0}0.665 $\pm$ 0.210 & \cellcolor[HTML]{E8F2FB}4.2 $\pm$ 1.9 & \cellcolor[HTML]{E8F2FB}24516 $\pm$ 20631 \\
Glaive & BF, D<=15 & \cellcolor[HTML]{E5F5E0}0.669 $\pm$ 0.209 & \cellcolor[HTML]{F3F8FD}5.9 $\pm$ 3.0 & \cellcolor[HTML]{F3F8FD}34873 $\pm$ 30628 \\
Glaive & TAAG-S & \cellcolor[HTML]{E5F5E0}0.553 $\pm$ 0.390 & \cellcolor[HTML]{F3F8FD}7.5 $\pm$ 5.3 & \cellcolor[HTML]{F3F8FD}42385 $\pm$ 39026 \\
Glaive & TAAG-R & \cellcolor[HTML]{E5F5E0}0.573 $\pm$ 0.388 & \cellcolor[HTML]{F3F8FD}7.3 $\pm$ 5.3 & \cellcolor[HTML]{F3F8FD}45635 $\pm$ 39938 \\
\midrule
Tool-Call-Data & SP & \cellcolor[HTML]{F2FAEF}0.434 $\pm$ 0.274 & \cellcolor[HTML]{D9EAF7}1.0 $\pm$ 0.0 & \cellcolor[HTML]{D9EAF7}5423 $\pm$ 2873 \\
Tool-Call-Data & BF, D<=5 & \cellcolor[HTML]{E5F5E0}0.568 $\pm$ 0.252 & \cellcolor[HTML]{E8F2FB}2.8 $\pm$ 1.0 & \cellcolor[HTML]{E8F2FB}15707 $\pm$ 13603 \\
Tool-Call-Data & BF, D<=10 & \cellcolor[HTML]{E5F5E0}0.582 $\pm$ 0.263 & \cellcolor[HTML]{E8F2FB}5.0 $\pm$ 2.3 & \cellcolor[HTML]{E8F2FB}29064 $\pm$ 26886 \\
Tool-Call-Data & BF, D<=15 & \cellcolor[HTML]{E5F5E0}0.584 $\pm$ 0.265 & \cellcolor[HTML]{F3F8FD}7.0 $\pm$ 3.6 & \cellcolor[HTML]{F3F8FD}42582 $\pm$ 40103 \\
Tool-Call-Data & TAAG-S & \cellcolor[HTML]{E5F5E0}0.537 $\pm$ 0.327 & \cellcolor[HTML]{F3F8FD}7.6 $\pm$ 4.5 & \cellcolor[HTML]{F3F8FD}50403 $\pm$ 46326 \\
Tool-Call-Data & TAAG-R & \cellcolor[HTML]{E5F5E0}0.586 $\pm$ 0.299 & \cellcolor[HTML]{F3F8FD}7.2 $\pm$ 4.1 & \cellcolor[HTML]{F3F8FD}50556 $\pm$ 41564 \\
\bottomrule
\end{tabular}
\end{adjustbox}
\caption{Source-level feedback-depth comparison. SP uses one attempt, BF rows vary the maximum correction budget, and TAAG-S/TAAG-R use structured TAAG feedback with $D\leq15$. TAAG-S/TAAG-R rows use retained summary-log aggregates so the values match the other structured-feedback source tables. Tokens are per instance. Green cells indicate higher-is-better pass rate; blue cells indicate lower-is-better attempts and token use.}
\label{tab:appendix_iteration_by_source}
\end{table*}

\begin{table*}[t]
\centering
\small
\setlength{\tabcolsep}{3pt}
\renewcommand{\arraystretch}{1.1}
\begin{tabular}{llccc}
\toprule
\textbf{Model family} & \textbf{Mode} & \textbf{TAAG pass rate} & \textbf{Attempts} & \textbf{Tokens} \\
\midrule
Instruction-following & SP & \cellcolor[HTML]{F2FAEF}0.435 $\pm$ 0.259 & \cellcolor[HTML]{D9EAF7}1.0 $\pm$ 0.0 & \cellcolor[HTML]{D9EAF7}5721 $\pm$ 2822 \\
Instruction-following & BF, D<=5 & \cellcolor[HTML]{E5F5E0}0.514 $\pm$ 0.249 & \cellcolor[HTML]{E8F2FB}3.0 $\pm$ 1.0 & \cellcolor[HTML]{E8F2FB}19759 $\pm$ 13398 \\
Instruction-following & BF, D<=10 & \cellcolor[HTML]{E5F5E0}0.519 $\pm$ 0.254 & \cellcolor[HTML]{F3F8FD}5.5 $\pm$ 2.2 & \cellcolor[HTML]{F3F8FD}36752 $\pm$ 26511 \\
Instruction-following & BF, D<=15 & \cellcolor[HTML]{E5F5E0}0.522 $\pm$ 0.256 & \cellcolor[HTML]{F3F8FD}7.9 $\pm$ 3.5 & \cellcolor[HTML]{F3F8FD}53863 $\pm$ 39537 \\
Instruction-following & TAAG-S & \cellcolor[HTML]{E5F5E0}0.617 $\pm$ 0.326 & \cellcolor[HTML]{F3F8FD}6.6 $\pm$ 4.4 & \cellcolor[HTML]{F3F8FD}38324 $\pm$ 33813 \\
Instruction-following & TAAG-R & \cellcolor[HTML]{E5F5E0}0.633 $\pm$ 0.295 & \cellcolor[HTML]{F3F8FD}6.5 $\pm$ 4.1 & \cellcolor[HTML]{F3F8FD}41677 $\pm$ 32564 \\
\midrule
Reasoning/thinking & SP & \cellcolor[HTML]{F2FAEF}0.371 $\pm$ 0.280 & \cellcolor[HTML]{D9EAF7}1.0 $\pm$ 0.0 & \cellcolor[HTML]{D9EAF7}5679 $\pm$ 2673 \\
Reasoning/thinking & BF, D<=5 & \cellcolor[HTML]{E5F5E0}0.739 $\pm$ 0.233 & \cellcolor[HTML]{E8F2FB}2.1 $\pm$ 1.0 & \cellcolor[HTML]{D9EAF7}8541 $\pm$ 5290 \\
Reasoning/thinking & BF, D<=10 & \cellcolor[HTML]{D9F0D3}0.755 $\pm$ 0.215 & \cellcolor[HTML]{E8F2FB}3.4 $\pm$ 2.1 & \cellcolor[HTML]{E8F2FB}12888 $\pm$ 9987 \\
Reasoning/thinking & BF, D<=15 & \cellcolor[HTML]{D9F0D3}0.766 $\pm$ 0.204 & \cellcolor[HTML]{E8F2FB}4.6 $\pm$ 3.1 & \cellcolor[HTML]{E8F2FB}17027 $\pm$ 14570 \\
Reasoning/thinking & TAAG-S & \cellcolor[HTML]{E5F5E0}0.509 $\pm$ 0.415 & \cellcolor[HTML]{F3F8FD}8.0 $\pm$ 5.8 & \cellcolor[HTML]{F3F8FD}51835 $\pm$ 48240 \\
Reasoning/thinking & TAAG-R & \cellcolor[HTML]{E5F5E0}0.582 $\pm$ 0.424 & \cellcolor[HTML]{F3F8FD}7.2 $\pm$ 5.8 & \cellcolor[HTML]{F3F8FD}48449 $\pm$ 46882 \\
\midrule
Tool/structured-output & SP & \cellcolor[HTML]{F2FAEF}0.476 $\pm$ 0.301 & \cellcolor[HTML]{D9EAF7}1.0 $\pm$ 0.0 & \cellcolor[HTML]{D9EAF7}5745 $\pm$ 2948 \\
Tool/structured-output & BF, D<=5 & \cellcolor[HTML]{D9F0D3}0.751 $\pm$ 0.118 & \cellcolor[HTML]{E8F2FB}2.1 $\pm$ 0.5 & \cellcolor[HTML]{D9EAF7}9729 $\pm$ 7683 \\
Tool/structured-output & BF, D<=10 & \cellcolor[HTML]{D9F0D3}0.762 $\pm$ 0.126 & \cellcolor[HTML]{E8F2FB}3.3 $\pm$ 1.1 & \cellcolor[HTML]{E8F2FB}16092 $\pm$ 14050 \\
Tool/structured-output & BF, D<=15 & \cellcolor[HTML]{D9F0D3}0.763 $\pm$ 0.126 & \cellcolor[HTML]{E8F2FB}4.5 $\pm$ 1.7 & \cellcolor[HTML]{E8F2FB}22551 $\pm$ 20454 \\
Tool/structured-output & TAAG-S & \cellcolor[HTML]{F2FAEF}0.436 $\pm$ 0.361 & \cellcolor[HTML]{FFF7EC}9.0 $\pm$ 5.1 & \cellcolor[HTML]{F3F8FD}46127 $\pm$ 50544 \\
Tool/structured-output & TAAG-R & \cellcolor[HTML]{F2FAEF}0.409 $\pm$ 0.348 & \cellcolor[HTML]{FFF7EC}9.3 $\pm$ 4.9 & \cellcolor[HTML]{F3F8FD}55340 $\pm$ 44853 \\
\bottomrule
\end{tabular}
\caption{Model-family feedback-depth comparison. SP uses one attempt, BF rows vary the maximum correction budget, and TAAG-S/TAAG-R use structured TAAG feedback with $D\leq15$. TAAG-S/TAAG-R rows use retained summary-log aggregates so the values match the other structured-feedback family tables. Tokens are per instance. Green cells indicate higher-is-better pass rate; blue cells indicate lower-is-better attempts and token use.}
\label{tab:appendix_iteration_by_family}
\end{table*}

\subsection{Model and Source Sensitivity}
\label{app:rs_mss}

\begin{figure*}[t]
    \centering
    \includegraphics[width=0.74\textwidth]{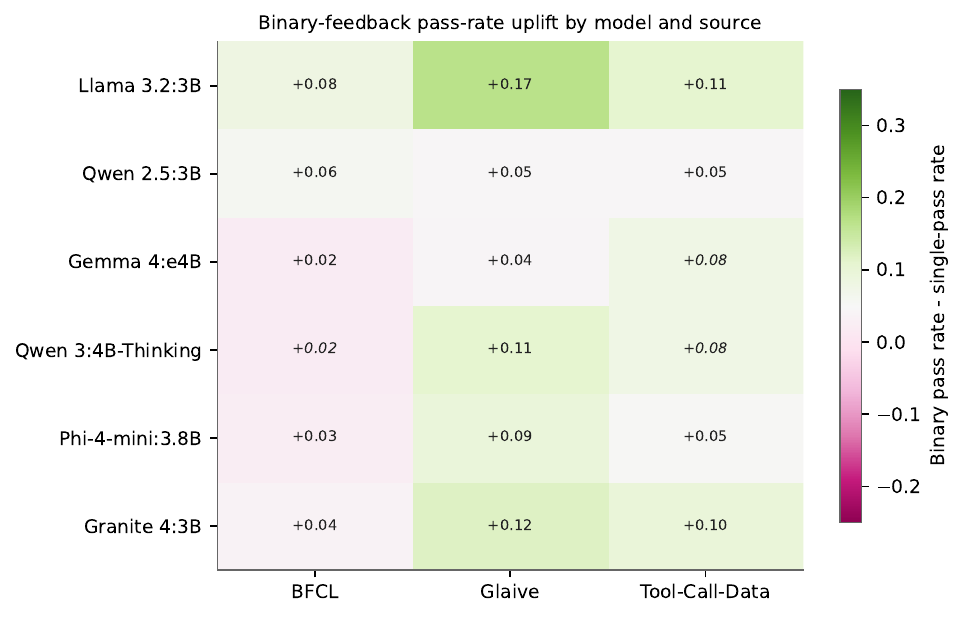}
    \caption{Binary-feedback uplift by model and source. Each cell is the mean TAAG pass-rate difference BF minus SP, averaged over matched registry sizes for that model--source pair when available. Positive values favor BF. Italic cells use the retained family--source aggregate when the model--source matched aggregate is unavailable.}
    \label{fig:appendix_model_source_uplift}
\end{figure*}

Table~\ref{tab:appendix_family_scale_sensitivity} asks whether model families fail in the same way as registries grow. It reports both final pass rate and argument grounding, because a model can keep selecting registered tools while still losing grounded argument values. Figure~\ref{fig:appendix_model_source_uplift} shows a different slice of the same data: the pass-rate change from binary feedback relative to single pass for each model and source. Binary feedback helps unevenly; it is a useful probe of recoverability, not a uniform ranking of models.

\begin{table*}[t]
\centering
\footnotesize
\setlength{\tabcolsep}{2pt}
\renewcommand{\arraystretch}{1.1}
\begin{adjustbox}{max width=\textwidth}
\begin{tabular}{llccccc}
\toprule
\textbf{Model family} & \textbf{Mode} & \textbf{Small pass} & \textbf{Large pass} & \textbf{Pass drop} & \textbf{Small grounding} & \textbf{Large grounding} \\
\midrule
Instruction-following & SP & \cellcolor[HTML]{E5F5E0}0.651 $\pm$ 0.076 & \cellcolor[HTML]{FFF7EC}0.183 $\pm$ 0.138 & \cellcolor[HTML]{F2FAEF}0.469 & \cellcolor[HTML]{D9F0D3}0.853 $\pm$ 0.084 & \cellcolor[HTML]{F2FAEF}0.333 $\pm$ 0.161 \\
Instruction-following & BF & \cellcolor[HTML]{E5F5E0}0.737 $\pm$ 0.076 & \cellcolor[HTML]{F2FAEF}0.271 $\pm$ 0.132 & \cellcolor[HTML]{F2FAEF}0.466 & \cellcolor[HTML]{D9F0D3}0.896 $\pm$ 0.078 & \cellcolor[HTML]{E5F5E0}0.607 $\pm$ 0.197 \\
\midrule
Reasoning/thinking & SP & \cellcolor[HTML]{E5F5E0}0.542 $\pm$ 0.174 & \cellcolor[HTML]{FFF7EC}0.077 $\pm$ 0.154 & \cellcolor[HTML]{F2FAEF}0.465 & \cellcolor[HTML]{E5F5E0}0.622 $\pm$ 0.223 & \cellcolor[HTML]{FFF7EC}0.092 $\pm$ 0.182 \\
Reasoning/thinking & BF & \cellcolor[HTML]{E5F5E0}0.663 $\pm$ 0.088 & \cellcolor[HTML]{FFF7EC}0.098 $\pm$ 0.195 & \cellcolor[HTML]{E5F5E0}0.565 & \cellcolor[HTML]{D9F0D3}0.769 $\pm$ 0.144 & \cellcolor[HTML]{FFF7EC}0.111 $\pm$ 0.223 \\
\midrule
Tool/structured-output & SP & \cellcolor[HTML]{E5F5E0}0.696 $\pm$ 0.116 & \cellcolor[HTML]{FFF7EC}0.218 $\pm$ 0.239 & \cellcolor[HTML]{F2FAEF}0.478 & \cellcolor[HTML]{D9F0D3}0.863 $\pm$ 0.141 & \cellcolor[HTML]{F2FAEF}0.308 $\pm$ 0.293 \\
Tool/structured-output & BF & \cellcolor[HTML]{D9F0D3}0.779 $\pm$ 0.069 & \cellcolor[HTML]{F2FAEF}0.493 $\pm$ 0.245 & \cellcolor[HTML]{F2FAEF}0.286 & \cellcolor[HTML]{D9F0D3}0.901 $\pm$ 0.062 & \cellcolor[HTML]{E5F5E0}0.577 $\pm$ 0.237 \\
\bottomrule
\end{tabular}
\end{adjustbox}
\caption{Model-family sensitivity to registry scale. The table separates end-to-end pass rate from argument-grounding score to show whether scale primarily changes final success, grounding, or both.}
\label{tab:appendix_family_scale_sensitivity}
\end{table*}

\subsection{Structured Diagnostic Feedback}
\label{app:rs_sdf}

\begin{figure*}[t]
    \centering
    \includegraphics[width=\textwidth]{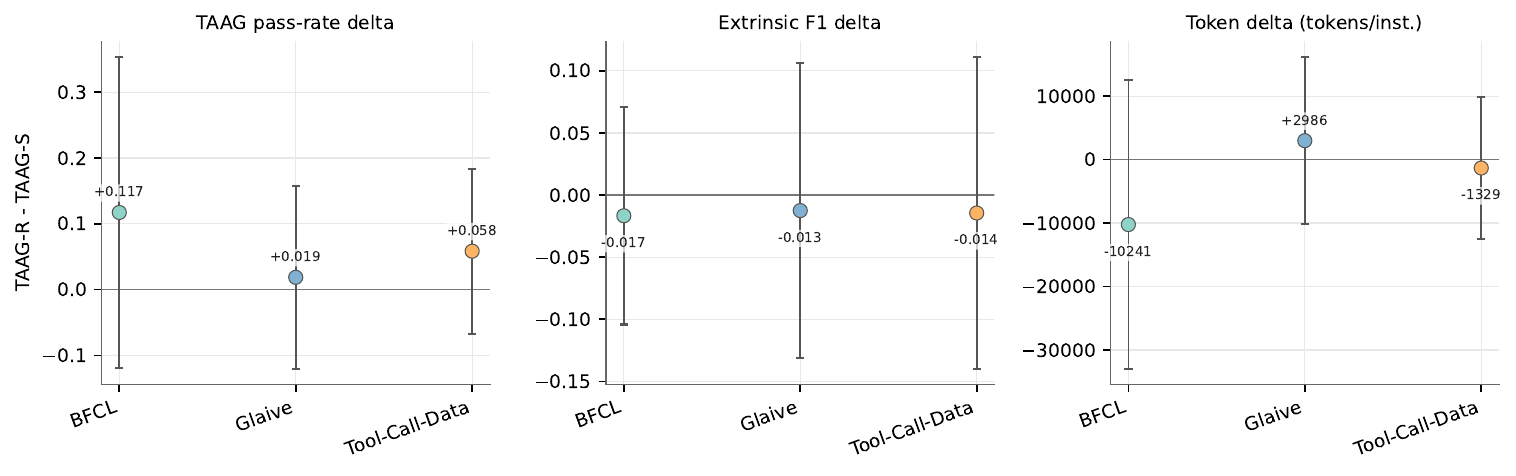}
    \caption{Paired effect of TAAG-R relative to TAAG-S. Each point is the mean paired delta across matched source--model--scale configurations. Positive pass-rate and F1 deltas favor TAAG-R, while positive token delta means TAAG-R used more tokens per instance. Error bars are standard deviations across matched configurations.}
    \label{fig:appendix_structured_delta}
\end{figure*}

The structured-feedback summary table in the main paper reports comparable stage-diagnosis and requirement-diagnosis summary-log results. For the configurations whose row-level structured outputs are present, Table~\ref{tab:rci2_structured_family_summary} reports the corresponding TAAG and extrinsic metrics by model family. Table~\ref{tab:appendix_structured_delta} and Figure~\ref{fig:appendix_structured_delta} use paired source--model--scale comparisons to show how much the requirement-level signal changes pass rate, F1, attempts, and token use relative to stage-only feedback. TAAG-R can improve pass rate over TAAG-S, but the gain is small compared with registry-scale degradation and does not guarantee lower token use.

\begin{table*}[t]
\centering
\small
\setlength{\tabcolsep}{3pt}
\renewcommand{\arraystretch}{1.15}
\begin{tabular}{llcc}
\toprule
\textbf{Model Family} & \textbf{Metric} & \textbf{TAAG-S} & \textbf{TAAG-R} \\
\midrule
Instruction-following & TAAG pass rate & \cellcolor[HTML]{D9F0D3}0.760 $\pm$ 0.336 & \cellcolor[HTML]{D9F0D3}0.860 $\pm$ 0.313 \\
 & Extrinsic F1 & \cellcolor[HTML]{D9F0D3}0.882 $\pm$ 0.176 & \cellcolor[HTML]{D9F0D3}0.965 $\pm$ 0.063 \\
 & Grounding score & \cellcolor[HTML]{D9F0D3}0.827 $\pm$ 0.255 & \cellcolor[HTML]{D9F0D3}0.900 $\pm$ 0.224 \\
 & Tokens & \cellcolor[HTML]{F3F8FD}35482 $\pm$ 46448 & \cellcolor[HTML]{E8F2FB}19573 $\pm$ 39169 \\
\midrule
Reasoning/thinking & TAAG pass rate & \cellcolor[HTML]{F2FAEF}0.425 $\pm$ 0.107 & \cellcolor[HTML]{F2FAEF}0.490 $\pm$ 0.035 \\
 & Extrinsic F1 & \cellcolor[HTML]{F2FAEF}0.425 $\pm$ 0.107 & \cellcolor[HTML]{F2FAEF}0.490 $\pm$ 0.035 \\
 & Grounding score & \cellcolor[HTML]{F2FAEF}0.425 $\pm$ 0.053 & \cellcolor[HTML]{F2FAEF}0.490 $\pm$ 0.071 \\
 & Tokens & \cellcolor[HTML]{FFF7EC}124784 $\pm$ 1800 & \cellcolor[HTML]{FFF7EC}123496 $\pm$ 2525 \\
\midrule
Tool/structured-output & TAAG pass rate & \cellcolor[HTML]{E5F5E0}0.700 $\pm$ 0.071 & \cellcolor[HTML]{E5F5E0}0.550 $\pm$ 0.071 \\
 & Extrinsic F1 & \cellcolor[HTML]{E5F5E0}0.600 $\pm$ 0.071 & \cellcolor[HTML]{E5F5E0}0.537 $\pm$ 0.053 \\
 & Grounding score & \cellcolor[HTML]{D9F0D3}0.900 $\pm$ 0.107 & \cellcolor[HTML]{D9F0D3}0.887 $\pm$ 0.018 \\
 & Tokens & \cellcolor[HTML]{E8F2FB}19834 $\pm$ 2525 & \cellcolor[HTML]{F3F8FD}30349 $\pm$ 1226 \\
\bottomrule
\end{tabular}
\caption{Structured-feedback row-level results, summarized by model family and feedback variant. Values are means with standard deviations across configuration-level aggregates computed from non-error rows.}
\label{tab:rci2_structured_family_summary}
\end{table*}

\begin{table*}[t]
\centering
\small
\setlength{\tabcolsep}{3pt}
\renewcommand{\arraystretch}{1.1}
\begin{tabular}{llc}
\toprule
\textbf{Source} & \textbf{Metric} & \textbf{TAAG-R $-$ TAAG-S} \\
\midrule
BFCL & TAAG pass rate & \cellcolor[HTML]{FFF7EC}0.117 $\pm$ 0.236 \\
 & Extrinsic F1 & \cellcolor[HTML]{FFF7EC}-0.017 $\pm$ 0.088 \\
 & Attempts & \cellcolor[HTML]{D9EAF7}-1.3 $\pm$ 2.7 \\
 & Tokens & \cellcolor[HTML]{D9EAF7}-10241 $\pm$ 22729 \\
\midrule
Glaive & TAAG pass rate & \cellcolor[HTML]{FFF7EC}0.019 $\pm$ 0.139 \\
 & Extrinsic F1 & \cellcolor[HTML]{FFF7EC}-0.013 $\pm$ 0.119 \\
 & Attempts & \cellcolor[HTML]{D9EAF7}-0.2 $\pm$ 1.7 \\
 & Tokens & \cellcolor[HTML]{D9EAF7}2986 $\pm$ 13136 \\
\midrule
Tool-Call-Data & TAAG pass rate & \cellcolor[HTML]{FFF7EC}0.058 $\pm$ 0.125 \\
 & Extrinsic F1 & \cellcolor[HTML]{FFF7EC}-0.014 $\pm$ 0.125 \\
 & Attempts & \cellcolor[HTML]{D9EAF7}-0.6 $\pm$ 1.2 \\
 & Tokens & \cellcolor[HTML]{D9EAF7}-1329 $\pm$ 11232 \\
\bottomrule
\end{tabular}
\caption{Paired structured-feedback deltas. TAAG-S is stage-diagnosis feedback and TAAG-R is requirement-diagnosis feedback. Positive pass-rate and F1 values favor TAAG-R; positive attempts and token values mean TAAG-R used more attempts or more tokens per instance for the same source, model, and registry size.}
\label{tab:appendix_structured_delta}
\end{table*}

\subsection{Alternative Visual Summaries}
\label{app:rs_avs}

\begin{figure*}[t]
    \centering
    \includegraphics[width=\textwidth]{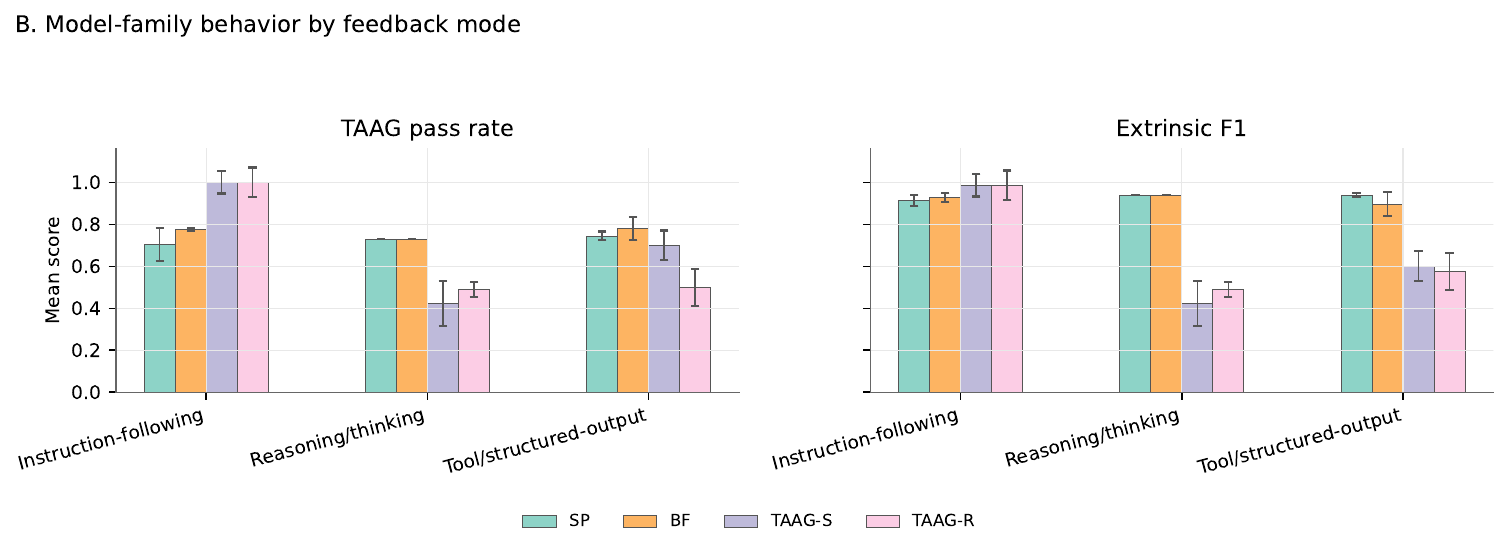}
    \caption{Grouped bar summary by model family and feedback mode. Bars show mean TAAG pass rate and extrinsic F1 for SP, BF, TAAG-S, and TAAG-R where retained row-level aggregates are available. Error bars show standard deviations across configuration-level aggregates.}
    \label{fig:appendix_family_bar}
\end{figure*}

\begin{figure*}[t]
    \centering
    \includegraphics[width=\textwidth]{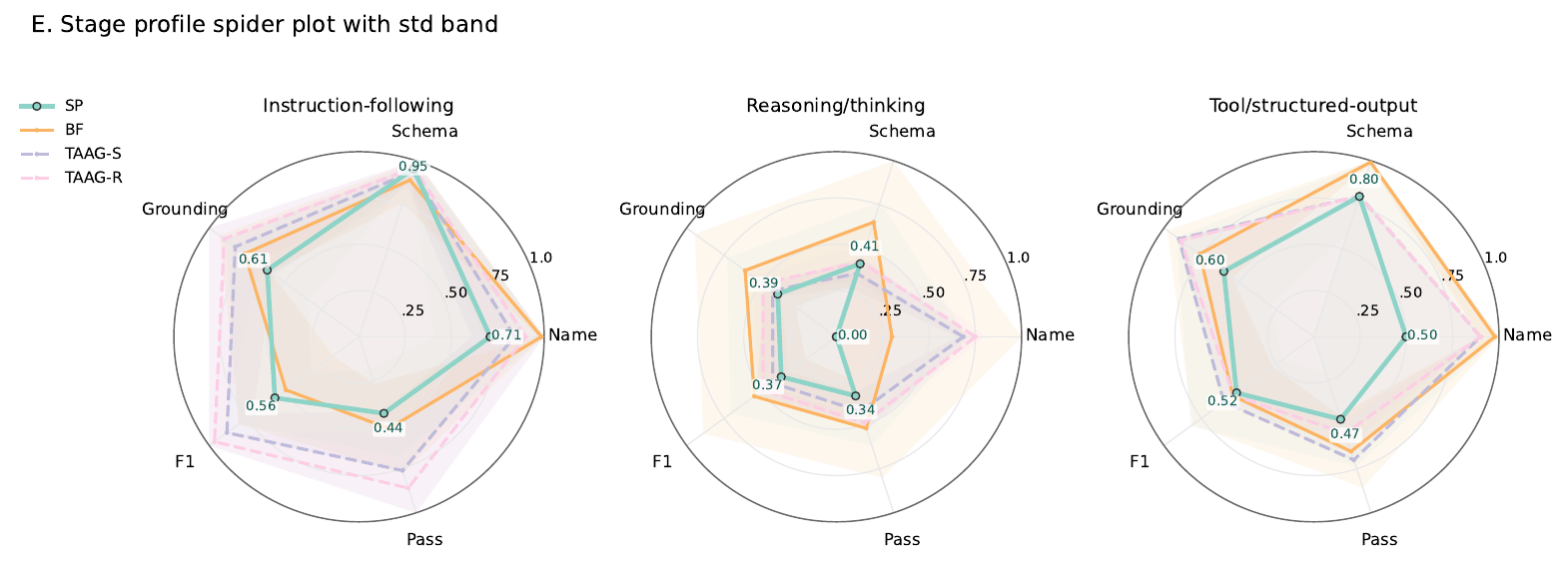}
    \caption{Spider plot of stage profiles by model family. Axes report registry conformance, schema completeness, grounding, extrinsic F1, and final TAAG pass rate on a common 0--1 scale. Lines show model-family means for SP, BF, TAAG-S, and TAAG-R; shaded bands show standard deviations across configuration-level aggregates.}
    \label{fig:appendix_stage_radar}
\end{figure*}

Figures~\ref{fig:appendix_family_bar} and~\ref{fig:appendix_stage_radar} show two compact model-family views. The grouped bar chart is easiest to read for headline pass rate and F1, while the spider plot makes the stage profile visible. The bar chart shows that model families differ in level but not in the existence of the feedback and scale tradeoffs. The spider plot shows that final pass rate, grounding, and extrinsic F1 do not always move together, which is why the paper reports the stages separately.

\begin{figure*}[t]
    \centering
    \includegraphics[width=0.78\textwidth]{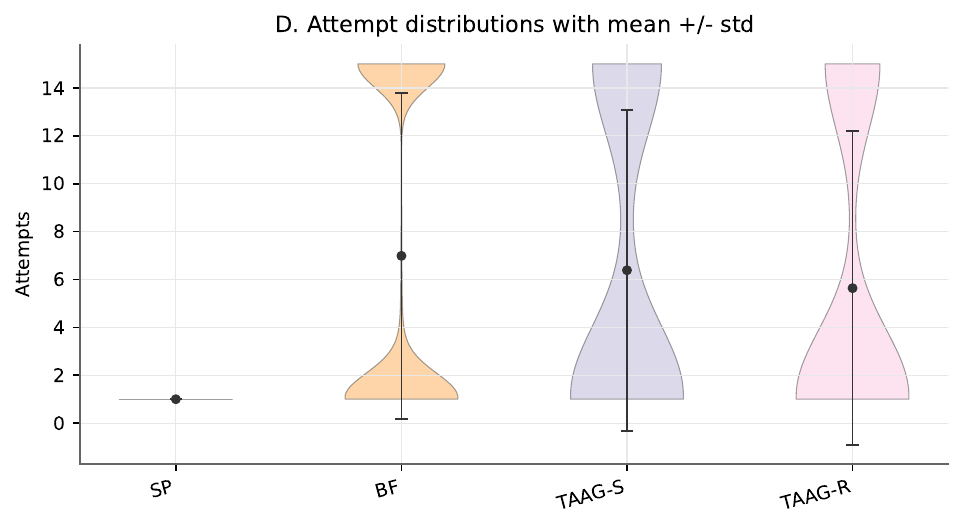}
    \caption{Attempt distributions for the considered feedback modes. The horizontal axis is feedback mode and the vertical axis is attempts per instance. Violin shapes show the distribution across retained configurations; black points mark means and vertical bars mark standard deviations.}
    \label{fig:appendix_attempts_violin}
\end{figure*}

\begin{figure*}[t]
    \centering
    \includegraphics[width=0.78\textwidth]{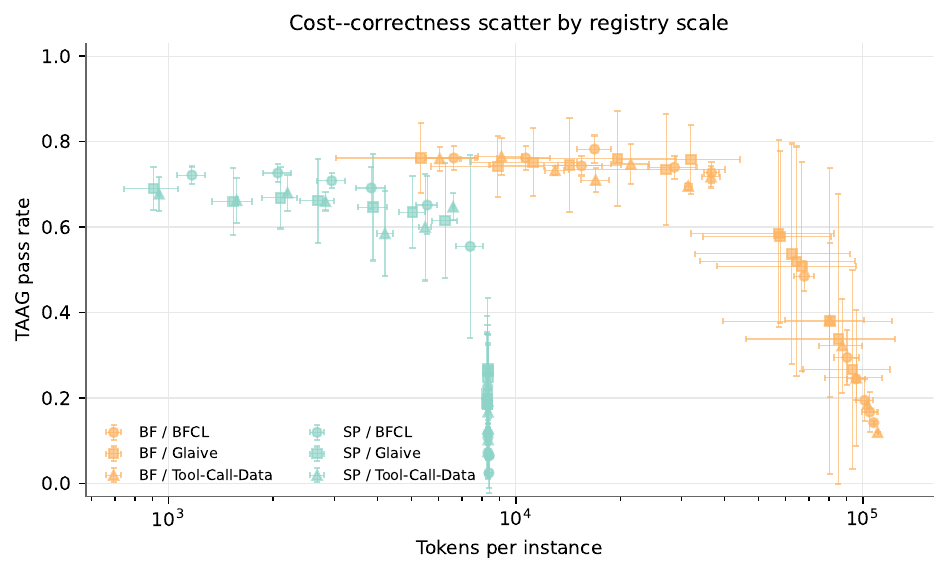}
    \caption{Cost--correctness scatter by registry scale. The horizontal axis is tokens per instance and the vertical axis is TAAG pass rate. Points show source--mode--scale aggregates with marker shape indicating source and color indicating feedback mode; error bars show standard deviations across configuration-level aggregates.}
    \label{fig:appendix_cost_scatter}
\end{figure*}

Figures~\ref{fig:appendix_attempts_violin} and~\ref{fig:appendix_cost_scatter} give distributional and cost views. The violin plot shows the attempt distribution for each considered feedback mode, with mean and standard deviation overlaid. The scatter plot shows the cost--correctness relationship by registry scale, with pastel colors for feedback mode and marker shape for source. Correction loops make the model spend many more attempts and tokens, and the most expensive configurations are not automatically the most accurate ones.

\end{document}